\documentclass[journal]{IEEEtran}
\usepackage[noadjust]{cite}
\ifCLASSINFOpdf
  \usepackage{graphicx}
  \usepackage{epstopdf}
  \graphicspath{{./graphics/}}
\else
\fi
\usepackage{amsmath,mathtools}
\usepackage{array}

\ifCLASSOPTIONcompsoc
  \usepackage[caption=false,font=normalsize,labelfont=sf,textfont=sf]{subfig}
\else
  \usepackage[caption=false,font=footnotesize]{subfig}
\fi
\begin{document}
%
\title{Segmentation of 3D High-frequency Ultrasound Images of Human Lymph Nodes Using Graph
Cut with Energy Functional Adapted to Local Intensity Distribution}
%
%
%

\author{Jen-wei~Kuo, Jonathan~Mamou, Yao~Wang, Emi~Saegusa-Beecroft,
             Junji~Machi, and Ernest~J.~Feleppa
        
\thanks{J.-w. Kuo and Y. Wang was with the Department of Electrical and Computer Engineering, NYU Tandon School of Engineering, Brooklyn, NY} 
\thanks{J. Mamou and E. J. Feleppa was with F. L. Lizzi Center for Biomedical Engineering, Riverside Research, New York, NY} 
\thanks{E. Saegusa-Beecroft and J. Machi was with University of Hawaii and Kuakini Medical Center, Honolulu, HI}} 

%
%

\markboth{Manuscript resubmitted, May~2017}%
{Shell \MakeLowercase{\textit{et al.}}: Bare Demo of IEEEtran.cls for IEEE Journals}
%



\maketitle

\begin{abstract}
Previous studies by our group have shown that three-dimensional high-frequency quantitative ultrasound methods have the potential to differentiate metastatic lymph nodes from cancer-free lymph nodes dissected from human cancer patients. To successfully perform these methods inside the lymph node parenchyma, an automatic segmentation method is highly desired to exclude the surrounding thin layer of fat from quantitative ultrasound processing and accurately correct for ultrasound attenuation. In high-frequency ultrasound images of lymph nodes, the intensity distribution of lymph node parenchyma and fat varies spatially because of acoustic attenuation and focusing effects. Thus, the intensity contrast between two object regions (e.g., lymph node parenchyma and fat) is also spatially varying. In our previous work, nested graph cut demonstrated its ability to simultaneously segment lymph node parenchyma, fat, and the outer phosphate-buffered saline bath even when some boundaries are lost because of acoustic attenuation and focusing effects. This paper describes a novel approach called graph cut with locally adaptive energy to further deal with spatially varying distributions of lymph node parenchyma and fat caused by inhomogeneous acoustic attenuation. The proposed method achieved Dice similarity coefficients of 0.937$\pm$0.035 when compared to expert manual segmentation on a representative dataset consisting of 115 three-dimensional lymph node images obtained from colorectal cancer patients.
\end{abstract}


%
\IEEEpeerreviewmaketitle

\section{Introduction}
%
%
%
%
\IEEEPARstart{H}{igh}-frequency (i.e., $>$20 MHz) ultrasound (HFU) provides fine-resolution images, on the order of 100 $\mu$m, because of its short wavelengths (i.e., $\sim$75 $\mu$m at 20 MHz) and small focal-zone beam diameters. Our previous work has shown that high-frequency, three-dimensional (3D), quantitative ultrasound (QUS) has the potential to differentiate fully metastatic from cancer-free dissected human lymph nodes (LNs) by quantifying tissue microstructure inside the LN parenchyma (LNP) \cite{saegusa2013three,mamou2011three,mamou2010three,coron2012quantitative}. A key limitation of our previous studies was our use of semi-automatic segmentation methods, which still required a certain amount of human supervision \cite{coron2008three}.

Accurate knowledge of whether LNs associated with a primary tumor are metastatic or non-metastatic is crucial for optimal treatment planning and patient management \cite{resch2013lymph}. For colorectal cancer patients undergoing colectomy, at least 12 LNs are harvested from each specimen by pathologists. In current clinical practice, only one central cross section of each dissected LN is stained and evaluated under a microscope by a pathologist due to labor and costs constraints. This severe spatial undersampling results in unsatisfactory rates of false-negative determinations in cases of small, off-center foci of metastatic cancer \cite{saegusa2013three}. The goal of our QUS studies is to guide the pathologists towards suspicious regions within LNs and markedly reduce the false-negative rate, particularly for micrometastases, i.e., foci with sizes between 0.2 and 2.0 mm.

\begin{figure}[t]
	\centering
	\begin{minipage}[t]{.32\linewidth}
		\includegraphics[width=\linewidth, height=\linewidth]{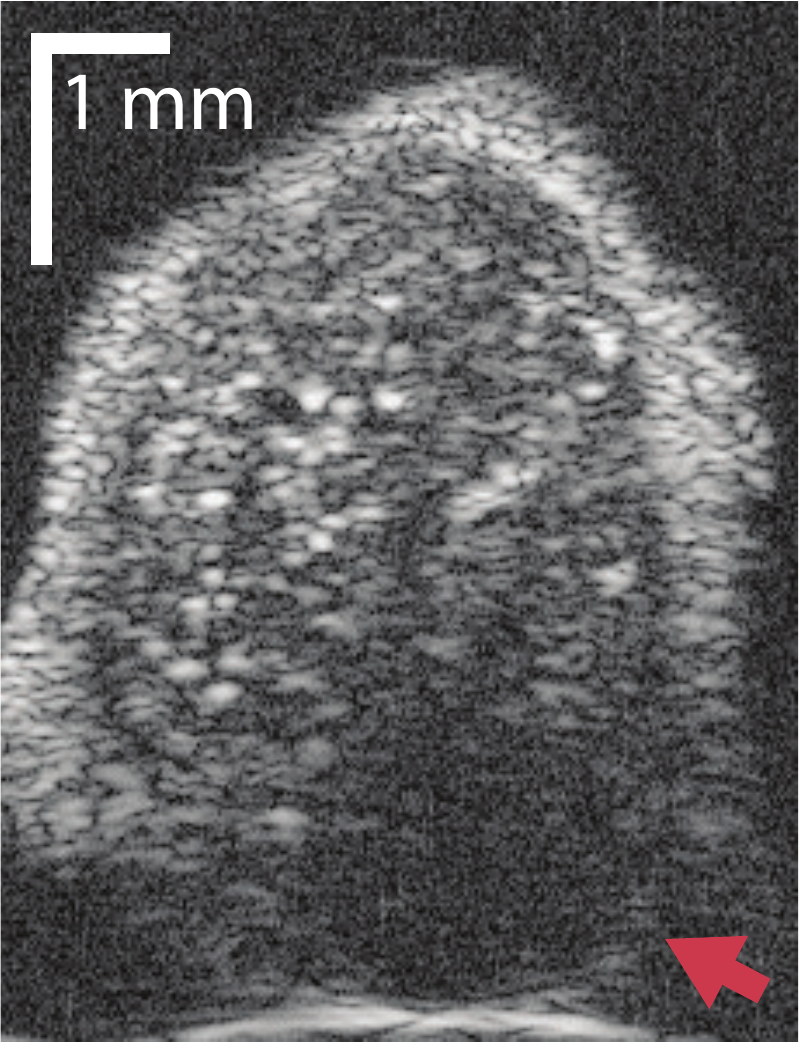}
                 \centerline{(a)}
	\end{minipage} 
	\begin{minipage}[t]{.32\linewidth}
		\includegraphics[width=\linewidth, height=\linewidth]{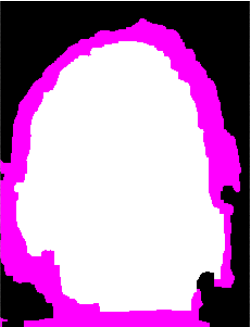}
                 \centerline{(b)}
	\end{minipage} 
	\begin{minipage}[t]{.32\linewidth}
		\includegraphics[width=\linewidth, height=\linewidth]{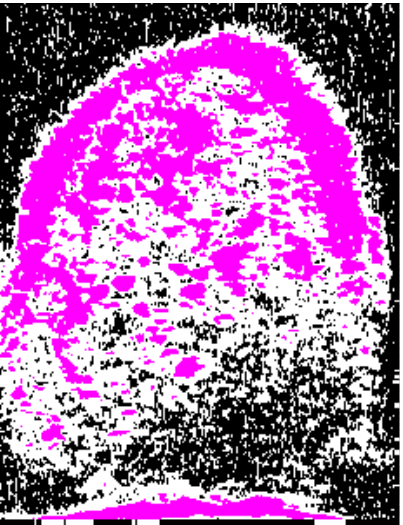}
                 \centerline{(c)}
	\end{minipage} 
	\caption{(a) Target HFU image. (b) Manual-segmentation result by an expert. LNP is white and fat region is pink. The vertical axis is depth. (c) The segmentation result obtained by K-means.}
\label{fig:intro_1}
\end{figure}

Figure \ref{fig:intro_1}(a) shows a representative HFU image of a human LN immersed in a phosphate-buffered saline (PBS) bath. Two different LN tissue types are visible: LNP (inside) and fat (surrounding). In order to perform QUS processing successfully, a 3D segmentation method is required to restrict QUS processing to LNP and to correct accurately for ultrasound attenuation \cite{mamou2011three}. Furthermore, accurate segmentation can provide detailed LNP shape information, which can be used to produce additional bio-markers of cancer.

Automatic LN segmentation in HFU images faces several challenges, such as speckle noise, low contrast between LNP and fat, implicit boundary and spatially varying intensity distributions caused by acoustic attenuation and focusing effects. Therefore, obtaining satisfactory results with intensity information alone is not possible. To illustrate this point, Figure \ref{fig:intro_1}(c) shows the result of a K-means analysis, which clusters voxels based on their intensities. Compared to manual segmentation result, some LNP voxels with high intensity are mislabeled as fat, and some LNP voxels with low intensity are mislabeled as PBS. Due to speckle noise and spatially varying intensity distributions in HFU images, other intensity-based segmentation methods such as region growing \cite{Gonzalez:2006:DIP:1076432} would also yield unsatisfactory results.

In addition to intensity information, we take advantage of structural information. As shown in Figure \ref{fig:intro_1}(b), LNP, fat, and PBS are in a nested configuration. Some approaches based on graph cuts (GCs) can segment multiple objects at the same time and preserve some structural relationship between these objects in the segmentation result. Ishikawa described representing different objects by separate layers of a graph, and showed that a multi-object segmentation problem can be solved as a binary GC problem if the energy term satisfies certain convex conditions \cite{ishikawa2003exact}. Although objects in the segmentation result under his structure have a nested relationship, he did not consider how to take advantage of it. To exploit the known generic relationships between objects (e.g., containment, attraction, and exclusion), Delong and Boykov described a multi-region GC \cite{delong2009globally}. However, Delong and Boykov's model requires manually selected seeds to define the missing boundary between objects with similar intensity distributions. In order to deal with the missing boundary in a nested structure automatically, Kuo et al. described nested graph cut (NGC) \cite{kuo2016nested}. In NGC, the missing boundary of an inner object can be defined by the convex hull of its outer object automatically. Kuo et al. applied NGC to segmenting LNs and overcame the missing boundary between LNP and PBS \cite{kuo2015novel}. Level-set-based approaches also can be used to preserve the nested relationships in segmentation processing. Nosrati et al. described a level-set version \cite{6814274} of Delong and Boykov's model. In addition, multiphase level-set approach also works well to preserve nested relationships by deforming multiple, non-crossing, nested, level-set functions. Bui et al. described a 3-phase, level-set method to segment LNs \cite{bui2015level}. Compared to GC-based methods, level-set-based methods can avoid grid bias and can permit faster calculations, but they require proper initialization or initial seeds to avoid converging to a local minimum. In a more-recent work, Bui et al. \cite{bui2015random} used random-forest classification to pre-segment images and initiate a level-set approach.

\begin{figure}[t]
	\centering
	\begin{minipage}[t]{.32\linewidth}
		\includegraphics[width=\linewidth, height=\linewidth]{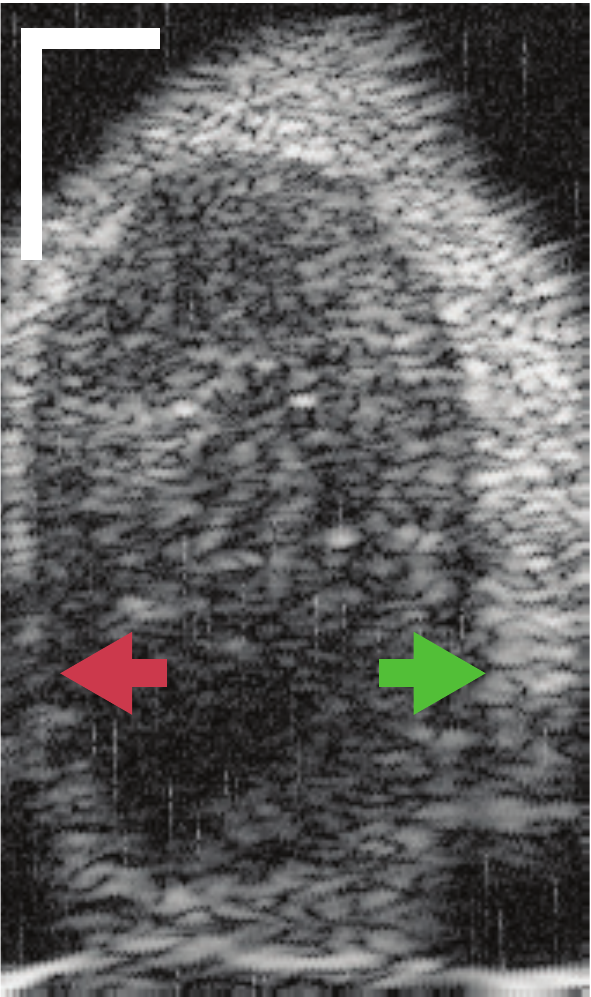}
                 \centerline{(a)}
	\end{minipage} 
	\begin{minipage}[t]{.32\linewidth}
		\includegraphics[width=\linewidth, height=\linewidth]{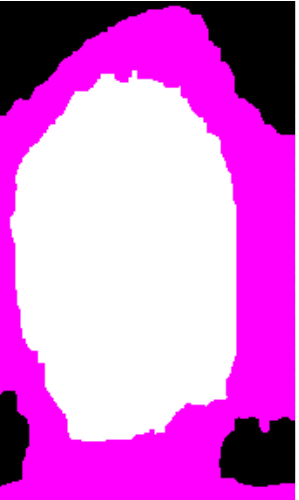}
                 \centerline{(b)}
	\end{minipage} 
	\begin{minipage}[t]{.32\linewidth}
		\includegraphics[width=\linewidth, height=\linewidth]{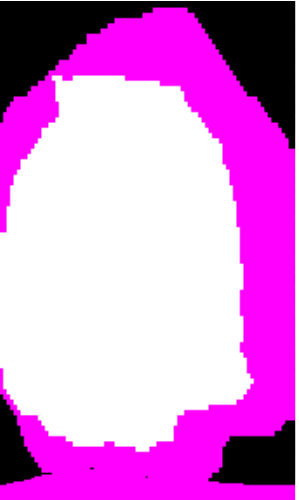}
                 \centerline{(c)}
	\end{minipage} 
	\caption{(a) Target HFU image. (b) Manual-segmentation result. (c) The segmentation result obtained by NGC with depth-dependent profile.}
\label{fig:intro_2}
\end{figure}

Intensity-distribution estimation is another challenge in LN segmentation. To obtain satisfactory segmentation, accurate intensity distributions for the three components of isolated LNs are required. Some clustering algorithms, such as K-means or expectation maximization (EM), may be able to categorize the intensity values of all voxels into three distributions to represent target objects. However, these approaches often yield unsatisfactory segmentation results for larger LNs (Figure \ref{fig:intro_1}(c)), where the intensity distribution fluctuates significantly with depth because of acoustic attenuation and focusing effects. In addition, attenuation effects may become so extreme that fat boundaries become invisible. (See the red arrow in Figure \ref{fig:intro_1}(a).) To mitigate the attenuation and missing-boundary problems, we used depth-dependent intensity profiles to model the depth-dependent variation of intensity mean and standard deviation in our prior work \cite{kuo2015novel}. Using an expectation-maximization (EM) -based iterative framework, depth-dependent profiles were estimated by a spline-based fitting process from the previous segmentation result, and the segmentation result was obtained by NGC using the estimated depth-dependent profiles. Instead of using an EM-based framework to update profiles, level-set-based approaches can also update profiles during deformation. Bui et al. developed a method, which applied depth-dependent profiles, called statistical transverse slice level-set (STS-LS) \cite{bui2016local}. Nevertheless, depth-dependent profiles are unable to deal with intensity inconsistency within the same depth caused by inhomogeneous acoustic attenuation as shown in Figure \ref{fig:intro_2}. Figure \ref{fig:intro_2}(c) shows the segmentation result obtained using a depth-dependent profile \cite{kuo2015novel}. Compared to the manual segmentation result in Figure \ref{fig:intro_2}(b), some fat regions (red arrow) are mislabeled as LNP because the fat on the right (green arrow) at the same depth is much brighter than the fat on the left. In this case, depth-dependent profiles are unable to model the intensity variations accurately. Using local distribution to differentiate LNP and the fat can solve the error caused by intensity inconsistency within the same depth. Bui et al. has used a local-region-based, gamma distribution to segment LNs \cite{bui2015level}. Level-set-based methods is good in applying local intensity distributions to differentiate regions. Since level-set-based methods deform the contour gradually, the local distributions can be updated by the boundary accordingly. However, updating distributions in local regions is computationally expensive. Furthermore, using local-region-based distributions with level-set-based methods is likely to result in convergence to a local minimum when the image is noisy. In contrast, GC-based approaches generally set the similarity cost by predefined intensity distributions. Because the boundary is unknown, the local region is hard to be determined in advance. 

The graph-based segmentation algorithm is a kind of region merging algorithm that can do segmentation without predefined distributions. Felzenszwalb and Huttenlocher \cite{felzenszwalb2004efficient} proposed a method called efficient graph-based algorithm (EGB), which maps pixels in an image to connected vertices in a graph and merges vertices into sub-graphs. EGB treats all pixel as different sub-graphs initially, and clusters similar sub-graphs gradually. It applies Pairwise Region Comparison Predicate (PRCP) to determine the minimum internal difference between pairs of sub-graphs, and merge two sub-graphs when the difference of the mean intensity of these two sub-graphs is lower than the minimum internal difference. To segment ultrasound images, Huang et al. \cite{huang2012robust} proposed robust graph-based (RGB) segmentation algorithm, which applied a novel PRCP that takes into account local statistics and signal-to-noise ratio (SNR). Furthermore, Zheng and Huang \cite{zheng2012graph} extended the RGB method to segment 3D ultrasound images. To deal with large scale voxel mergence, Chang et al. \cite{chang2015graph} proposed to apply hypothesis testing to measure the similarity when the sub-graphs are larger than certain threshold. However, the graph-based algorithm cannot cluster all sub-graphs in a target object with noise and high spatially varying intensity distributions and find the boundary between properly. 

\begin{figure}[t]
	\centering
	\begin{minipage}[t]{.4\linewidth}
		\includegraphics[width=\linewidth, height=\linewidth]{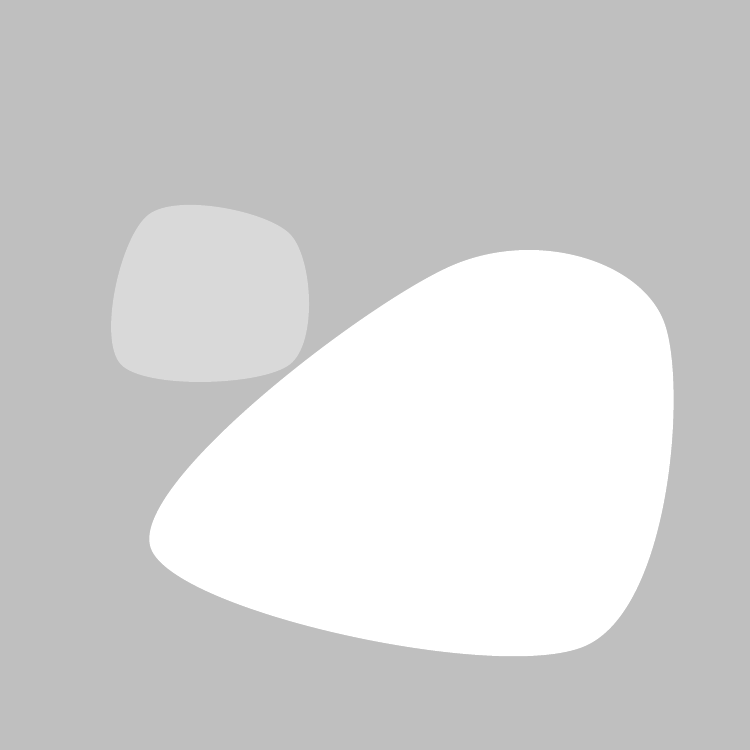}
                 \centerline{(a)}
	\end{minipage} 
	\begin{minipage}[t]{.4\linewidth}
		\includegraphics[width=\linewidth, height=\linewidth]{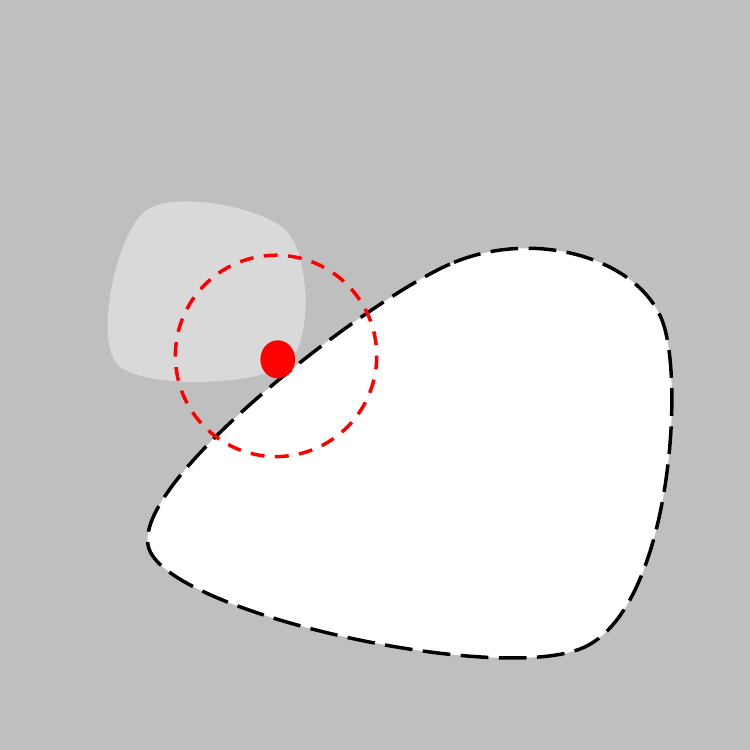}
                 \centerline{(b)}
	\end{minipage} 
	\begin{minipage}[t]{.4\linewidth}
		\includegraphics[width=\linewidth, height=\linewidth]{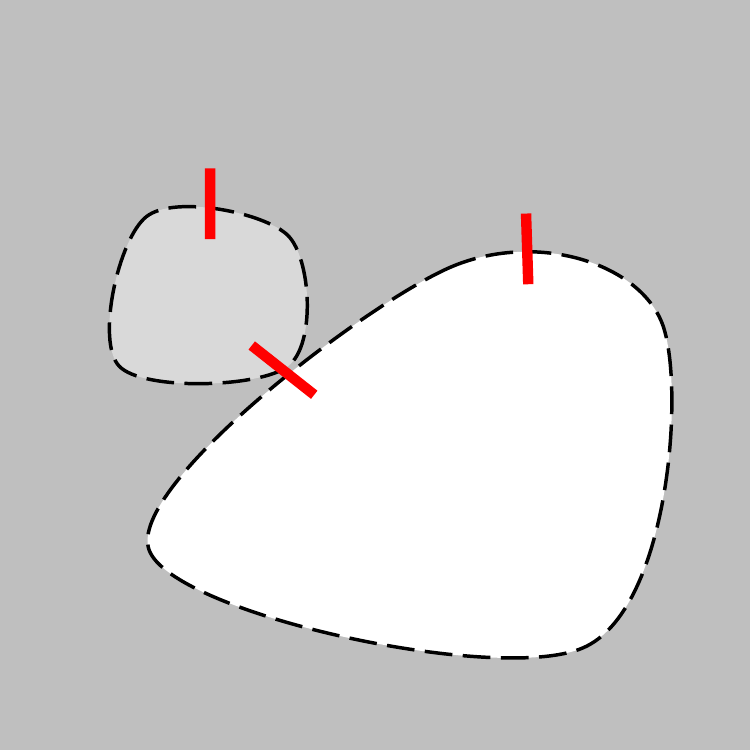}
                 \centerline{(c)}
	\end{minipage} 
	\begin{minipage}[t]{.4\linewidth}
		\includegraphics[width=\linewidth, height=\linewidth]{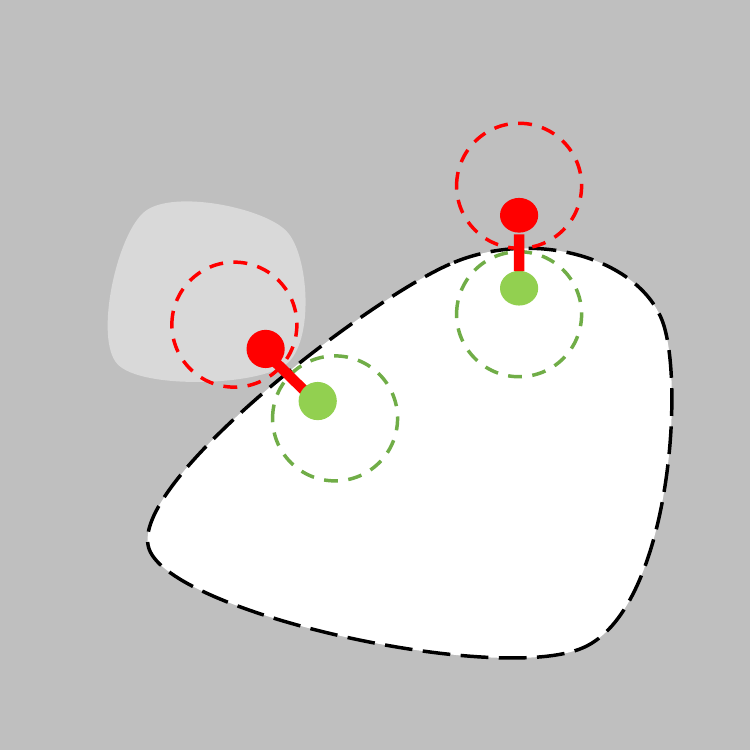}
                 \centerline{(d)}
	\end{minipage} 
	\caption{(a) Target toy image. (b) An illustration of level-set-based methods. The force on the red pixel is defined by the intensity distributions within the red dash line. (c) An illustration of graph-based segmentation methods. Red lines indicate the connections between sub-graphs. (d) An illustration of GC-LAE. The similarity between two connecting pixels (red and green) of an edge is defined by the similarity between these pixels and their corresponding local regions (dash lines in the opposite color).}
\label{fig:Toy}
\end{figure}

In order to deal with the spatially varying intensity distributions of LNP and fat and intensity inconsistency within the same depth, we propose a novel way to segment LNP and fat based on the locally observed intensity statistics. We call this method GC with locally adaptive energy (GC-LAE), which can differentiate LNP and fat voxels according to the similarity of distributions in local regions. Figure \ref{fig:Toy} illustrates the difference between level-set-based methods, graph-based segmentation methods, and the proposed method. Figure \ref{fig:Toy}(b) shows the way level-set-based method takes advantage of local distributions. In Figure \ref{fig:Toy}(b), the black dash line indicates the boundary of foreground found by a level-set based method. The similarity between the red pixel and both foreground and background can be defined by the distributions within the local region inside the red dash line. The intensity distribution of the foreground is obtained by the pixels inside the red and black dash lines, and the intensity distribution of the background is obtained by the pixels inside the red dash line and outside the black dash line. These intensity distributions are updated with the deformation. Different from level-set-based methods, graph-based segmentation methods defines local intensity distributions by sub-graphs. Figure \ref{fig:Toy}(c) illustrates a possible segmentation result of graph-based segmentation methods. There are three sub-graphs (white, light gray, and dark gray) in Figure \ref{fig:Toy}(c), and these sub-graphs are connected by edges in red. Graph-based methods obtain one intensity distribution for each region and compares the similarity of intensity distributions between connecting sub-graphs. If the difference of the mean intensity of two connecting sub-graphs is smaller than some threshold, they will be merged together. However, graph-based segmentation methods start from treating each pixel as a sub-graph and therefore is sensitive to noise and may not be able to segment an object in its entirety when it has large intensity variation. The proposed method, called GC-LAE, can separate target image into a foreground and a background by pre-defined local intensity distributions. Generally, GC-based approaches cannot define local regions like level-set-based methods because we cannot know the boundary (black dash line in Figure \ref{fig:Toy}(b)) in advance. If we define the local region of a pixel by the region centered on this pixel, the local regions of two connecting pixels will have some overlap and reduce the dissimilarity. To solve this problem, GC-LAE defines the local region according to the direction of edges. As illustrated in Figure \ref{fig:Toy}(d), the local regions of two connecting pixels (red and green) can be defined in the same direction of the edge between them and be non-overlapping. When the local region is small and the boundary of the foreground is smooth, the intensity dissimilarity of the local region of pixels at the cut boundary (black dash line) can be well preserved. With the proposed energy function, GC-LAE can make the cut forced by the local intensity dissimilarity, which is insensitive to the noise present in ultrasound images. To further deal with the blurry boundary in ultrasound images, GC-LAE can give pixels having larger distance to the edge higher weights to obtain local intensity distributions. In contrast to level-set-based approaches, GC-LAE can always achieve the global minimum of the defined energy function. 

The GC-LAE alone cannot be applied to the entire image volume to successfully separate all three regions. We introduce a robust framework combining NGC and GC-LAE to segment LNs. It first applies NGC to the entire image volume to obtain an initial segmentation into three nested regions: the PBS, fat, and LNP.  The resulting LNP and fat regions are then combined to form the LN-mask. Then it applies GC-LAE to the LN-mask to refine the segmentation of LNP and fat. The resulting segmentation is used to update the intensity distributions of the three regions. Finally the NGC is applied again using these updated distributions to obtain the final segmentation of three nested regions.This framework is fully automatic and does not require any manual initialization or predefined intensity distributions. 

This paper is organized as follows: Section \ref{sec:GCLAE} describes the graph structure of GC-LAE and defines local adaptive energy term and global energy term. Section \ref{sec:the overall algorithm} introduces the whole framework combining GC-LAE and NGC. Section \ref{sec:result} illustrates the segmentation results and compares the performance of NGC with depth-dependent profiles \cite{kuo2015novel}, RFC+STS-LS \cite{bui2015random}, and the proposed method. Section \ref{sec:discussion and conclusion} discusses the clinical significance and the future work and concludes this paper.

\section{Graph Cut With Locally Adapted Data and Edge Costs}
\label{sec:GCLAE}

Graph cut \cite{boykov2001fast} (GC) is a segmentation method that can segment all voxels in a 3D volume into two regions by minimizing an energy functional. Generally, the energy consists of a data term and a smoothness term. The data term is the sum of the cost at each voxel that is typically inversely proportional to the likelihood of a voxel belonging to the assigned region, determined from the voxel's intensity and the assumed intensity distribution of the assigned region. In GC, all voxels pairs that are directly adjacent to each other in 3D by 6-, 18- or 26-connectivity are considered in the neighborhood set. The smoothness term is the sum of the cost between every two voxels in a defined neighborhood set, which typically represents the cost if these two voxels are assigned into two different regions. Based on the defined energy functional, one can set up a corresponding graph, which includes all voxels as nodes as well as two additional nodes called source (region 1) and sink (region 2). The link cost between a voxel and source (resp. sink) is equal to the data cost corresponding to assign the voxel to region 1 (resp. region 2). The edge cost between every two neighboring voxels defines the link cost between the two corresponding nodes. The segmentation problem is defined as finding a minimal cut, which separates voxels that are connected to the source (hence assigned to region 1) and those connected to the sink (hence assigned to region 2) with a minimal total sum of link cost along the cut.

\begin{figure}[t]
	\centering
	\begin{minipage}[t]{.45\linewidth}
		\includegraphics[width=\linewidth, height=\linewidth]{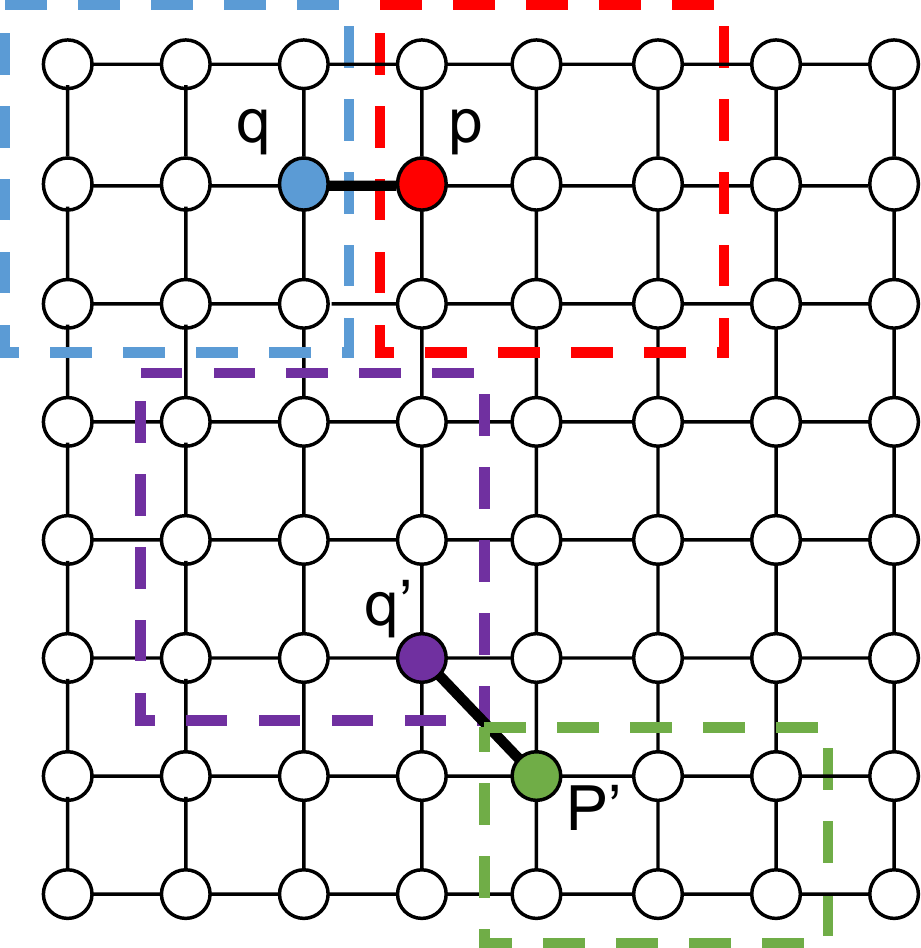}
                 \centerline{(a)}
	\end{minipage} 
	\begin{minipage}[t]{.4\linewidth}
		\includegraphics[width=1.2\linewidth, height=\linewidth]{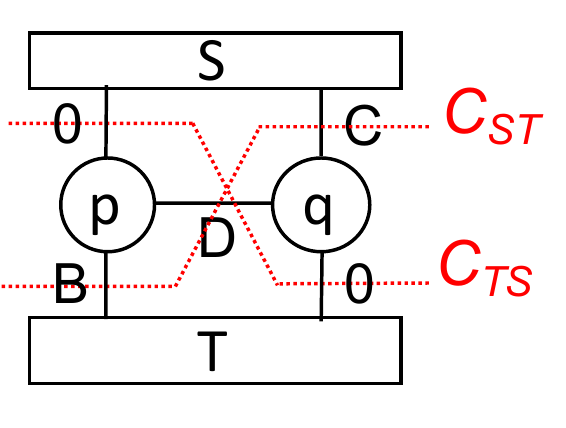}
                 \centerline{(b)}
	\end{minipage} 
	\caption{(a) The colored dash lines indicate the corresponding local regions of pixels in different neighborhood sets. (b) The energy cost between $p$ and $q$ can be made by three nonzero edges depending on the region assignment of $p$ and $q$. The red line associated with $C_{ST}$ indicates the energy cost when voxel $p$ is labeled as source $S$ and voxel $q$ is labeled as sink $T$.}
\label{fig:build_graph}
\end{figure}

\subsection{Energy Definition}

\subsubsection{Local adaptive energy}

In GC-LAE,  the local adaptive energy (LAE) of each edge is defined by the similarity of their corresponding local regions. As shown in Figure \ref{fig:Toy}(d), the LAE of an edge between two connecting voxels (red and green points) is defined by the similarity of its local regions (red and green dash lines). Because we know the location of the two voxels and the direction of the edge connecting them, we can define the local regions for the red and green voxels, respectively, to make them non-overlapping (as shown in Fig. 3(d)). Figure \ref{fig:build_graph}(a) illustrates how we define the local region for different pairs of voxels. The intensity distribution of local regions can be modeled by preferred model. To avoid the effect of blurry boundary, it is possible to give pixels in local regions having larger distances to the edge higher weights to obtain the intensity distributions. In this experiment, the intensity distributions of local regions are modeled by Gaussian distribution with unitary weight. We define the LAE of a pair of neighboring voxels $p$ and $q$ based on the product of the probability of voxel $p$ belonging to the local distribution of voxel $q$, and the probability that $q$ belonging to the local distribution of $p$. Maximizing the product of these probabilities is equivalent to minimizing the sum of the negative log likelihood of these probabilities. Let $x_p$ denote the assigned label of voxel $p$, and $I_p$ the image intensity at this voxel. We call the region with higher intensity the foreground (Region 1), which is connected to the source ($S$), and the region with lower intensity the background (Region 2), which is connected to the sink ($T$). Assuming the local distribution over each neighborhood is Gaussian, when the assigned labels for the two voxels are the same, the LAE cost $C_{p,q}$ between voxels $p$ and $q$ is defined as follows:
\[
	C_{p,q}(x_p=x_q)=\frac{(I_p-\mu_q)^2}{{2\sigma_q^2}}+\frac{(I_q-\mu_p)^2}{{2\sigma_p^2}}
\]
where $\mu_p$ and $\sigma_p$ denote the mean and the standard deviation (STD) of the local distribution of voxel $p$, respectively. When the assigned labels for the two voxels are different, we assign the following LAE costs:
\[ \begin{split}
	C_{p,q}(x_p=S,&x_q=T)=\\
	&\frac{(I_p-(\mu_q+k\sigma_q))^2}{{2\sigma_q^2}}+\frac{(I_q-(\mu_p-k\sigma_p))^2}{{2\sigma_p^2}}
\end{split} \]
\begin{equation}
\begin{split}
	C_{p,q}(x_p=T,&x_q=S)=\\
	&\frac{(I_p-(\mu_q-k\sigma_q))^2}{{2\sigma_q^2}}+\frac{(I_q-(\mu_p+k\sigma_p))^2}{{2\sigma_p^2}}
\end{split} 
\label{eq:LAE}
\end{equation}

With the above definition, when the difference between voxel $p$ (resp. $q$) and $\mu_q$ (resp. $\mu_p)$ is larger than $\frac{k\sigma_q}{2}$ (resp. $\frac{k\sigma_p}{2}$), the cost of having the same label is larger than the cost of having different labels, therefore such energy definition encourages a cut between the two pixels. The introduction of the $\frac{k\sigma}{2}$ term in the definition of (\ref{eq:LAE}) has a similar effect as the minimum internal difference used in graph-based segmentation methods, but GC-LAE does not treat it as a hard constraint. We can change the parameter $k$ based on the expected difference of local distributions between two regions. 

The minimization of the total energy cost between all voxels belonging to the neighborhood set can be formulated as a graph cut problem, with the graph structure shown in Figure \ref{fig:build_graph}(b). We connect node $p$ and $q$ to the source $S$ and sink $T$ by two edges with zero cost and three edges with link costs denoted by $B$, $C$, and $D$ respectively. These link costs can be derived from previously defined energy costs: $C_{SS}$, $C_{ST}$, $C_{TS}$, and $C_{TT}$ by solving the following equation, following the original idea of graph cut \cite{kolmogorov2004energy}:
\[
\left[
\begin{matrix}
 C_{SS} & C_{ST} \\
 C_{TS} & C_{TT} \\
\end{matrix}
\right]=A\left[
\begin{matrix}
 1 & 1 \\
 1 & 1 \\
\end{matrix}
\right]+B\left[
\begin{matrix}
 1 & 1 \\
 0 & 0 \\
\end{matrix}
\right]+C\left[
\begin{matrix}
 0 & 1 \\
 0 & 1 \\
\end{matrix}
\right]+D\left[
\begin{matrix}
 0 & 1 \\
 1 & 0 \\
\end{matrix}
\right]
\]

$A$ is a constant that can be ignored. Figure \ref{fig:build_graph}(b) illustrates the cuts corresponding to $C_{ST}$ and $C_{TS}$. When we label voxel $p$ as $S$ and voxel $q$ as $T$, the cost will be $C_{ST}$. When we label voxel $p$ as $T$ and voxel $q$ as $S$, the cost will be $C_{TS}$. 

\subsubsection{Global energy}
\label{sec:global energy}

In addition to local adaptive energy term, we can also add a global energy term based on pre-defined intensity distributions. Considering that the intensity distribution varies with the depth in ultrasound images due to depth-dependent acoustic attenuation, we estimate the mean intensity and STD of each region (LNP or fat) at each depth, and assign the same mean and variance to all voxels at the same depth. In this work, we connect all nodes shown in Figure \ref{fig:build_graph}(b) to both source (for FAT) and sink (for LNP) nodes by additional edges, and define the cost of these edges using following sigmoidal functions: 
\begin{equation}
	W_p^{i-1,i}(I_p) =
  	\begin{dcases}
    		\alpha\frac{1}{1+\exp(\frac{I_p-\mu_{LNP,p}}{\sigma})}  & \text{node } p \text{ to } T\\
    		\alpha\frac{{\exp(\frac{I_p-\mu_{FAT,p}}{\sigma})}}{1+\exp(\frac{I_p-\mu_{FAT,p}}{\sigma})},    & \text{node } p \text{ to } S
	\end{dcases}  \\
	\label{eq:threshold}
\end{equation}
where $\mu_{LNP,p}$ and $\mu_{FAT,p}$ stand for the mean of LNP and fat at voxel $p$, respectively. The slope $\sigma$ is defined as $(\mu_{FAT,p}-\mu_{LNP,p})/8$ in this experiment. $\alpha$ is the weight of this global energy term, with respect to the local adaptive term. In this experiment, we defined it as 5 times the number of pairs in a neighborhood set. In Sec. \ref{sec:result}, we show results with and without using the global energy term.

\section{The Overall Algorithm for Lymph Node Segmentation}
\label{sec:the overall algorithm}

\begin{figure}[t]
	\centering
	\begin{minipage}[t]{.9\linewidth}
		\includegraphics[width=\linewidth, height=\linewidth]{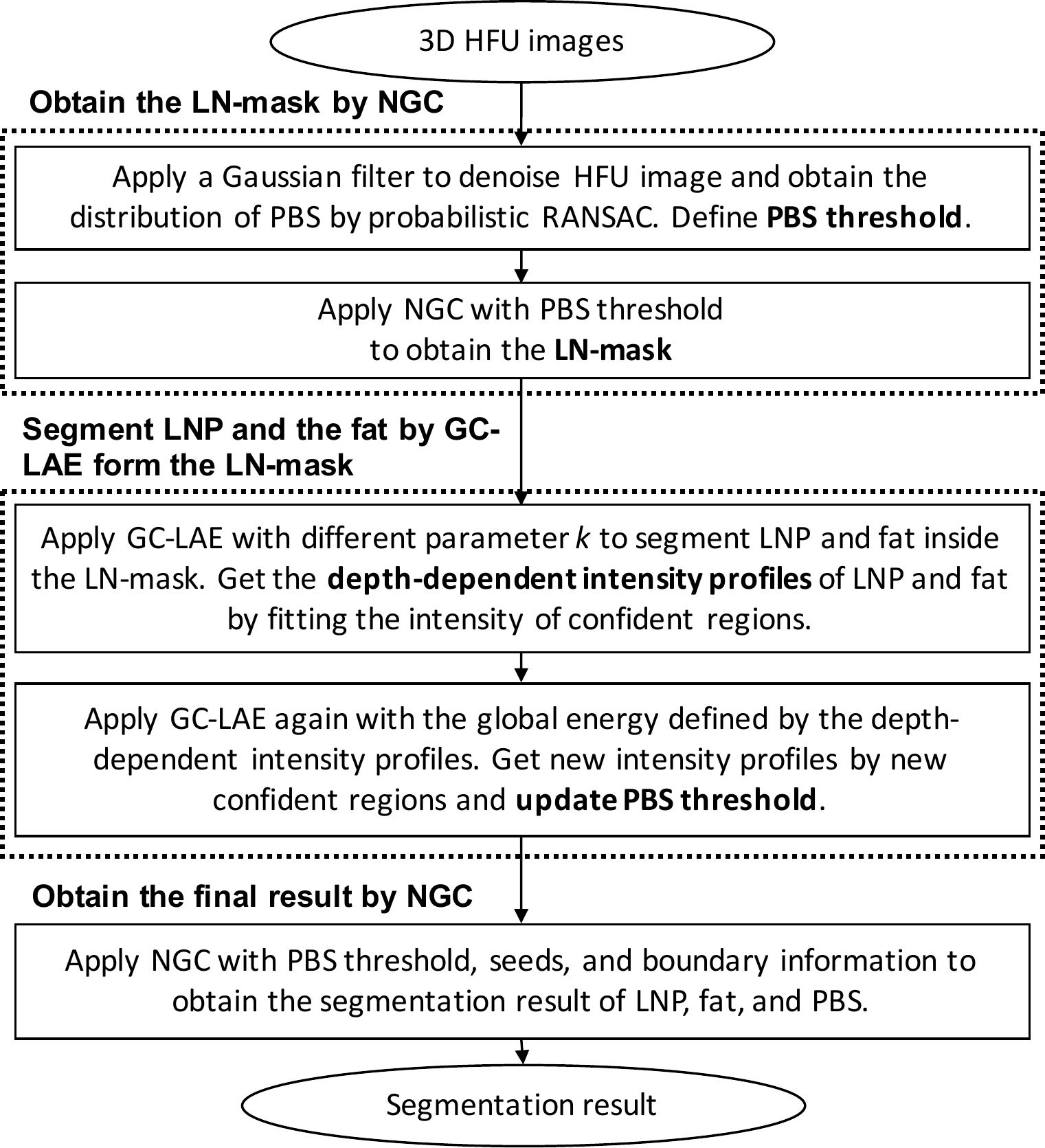}
	\end{minipage} 
	\caption{The flow of the proposed framework.}
\label{fig:flow}
\end{figure}

Figure \ref{fig:flow} shows the overall procedure for the proposed segmentation method, and the corresponding images are illustrated in Figure \ref{fig:result_flow}. The algorithm basically contains three parts. First, Nested Graph Cut (NGC) \cite{kuo2016nested} is used to segment the LN-mask, which contains LNP and the fat, from PBS. Second, after obtaining the LN-mask, GC-LAE is applied to separate LNP and fat voxels inside the LN-mask. Third, NGC is applied again to segment all three objects simultaneously to further refine the segmentation result by explicitly making use of the nested relationship between these three regions.

\subsection{3D HFU Data Acquisition}

LN dissection, histological preparation and HFU data acquisition protocols were previously described in detail \cite{mamou2011three}. Briefly, LNs were dissected from patients with histologically-proven cancers at the Kuakini Medical Center (KMC) in Honolulu, HI. Individual, freshly dissected LNs were placed in a PBS solution at room temperature and scanned. Radiofrequency (RF) signals were acquired using a single-element transducer with a 12.2 mm focal length, an f-number of 2, and a center frequency of 25.6 MHz. The RF echo signals were digitized at 400 MS/s using an 8-bit A/D board. 3D HFU data were acquired from each LN by scanning adjacent planes with a uniform plane and A-line spacing of 25 $\mu$m. The present segmentation study was carried out using a representative database of 3D HFU envelope data from 115 LNs (19 metastatic and 96 cancer-free LNs) from 67 colorectal-cancer patients. The Institutional Review Boards (IRBs) of the University of Hawaii and the KMC approved the participation of human subjects in the study. All participants were recruited at KMC and gave written informed consent as required by the IRBs.

\subsection{Initialize LN-mask}

We initialize the LN-mask, which contains only LNP and fat (Figure \ref{fig:result_flow}(c)), using the NGC method. In our previous work \cite{kuo2015novel}, NGC was used to separate three nested regions (PBS, fat, and LN) simultaneously based on the depth-dependent intensity distributions of each region. We first estimated the mean and STD of PBS by probabilistic random sample consensus (RANSAC) described in Sec. \ref{sec:RANSAC}. By assigning probabilities to candidate PBS voxels based on their intensities, RANSAC \cite{fischler1981random} obtained good estimations in a few tries (i.e., $\sim$5). Figure \ref{fig:result_flow}(b) shows the resulting probability map where bright voxels have a higher probability of belonging to PBS than dark voxels. After obtaining the distribution of PBS, we define PBS threshold, which is the intensity threshold between PBS and the LN-mask ($I^T$ in (\ref{eq:threshold2})), to differentiate PBS and LN-mask. With the data term defined by the PBS threshold (\ref{eq:threshold2}), NGC is able to obtain an accurate LN-mask (Figure \ref{fig:result_flow}(c)). Note that NGC can separate LN and PBS even when the lower part of LN is almost as dark as PBS due to acoustic attenuation.

\subsubsection{RANSAC-based PBS distribution estimation}
\label{sec:RANSAC}

Because the PBS region is generally very dark and can be approximated as a constant, we only need to estimate its mean intensity. To accomplish this, we used a probabilistic RANSAC approach. We assigned a probability of being PBS to each voxel, and randomly selected candidate PBS voxels from all voxels according to these probabilities. In each sampling, the mean of all selected candidate voxels was calculated, and the voxels that had intensities within a distance threshold to the mean intensity were identified and considered to be inliers for the PBS. The mean value determined from the sampling that led to the largest inlier set was the final estimated intensity mean for the PBS. We further computed the STD of the inlier voxels, and used a Gaussian function with the mean and STD as the probability density function for the PBS. This probability density function was employed by the NGC algorithm to separate PBS from LN-mask region in the next step. 

\begin{figure}[t]
	\centering
	\begin{minipage}[t]{.32\linewidth}
		\includegraphics[width=\linewidth, height=\linewidth]{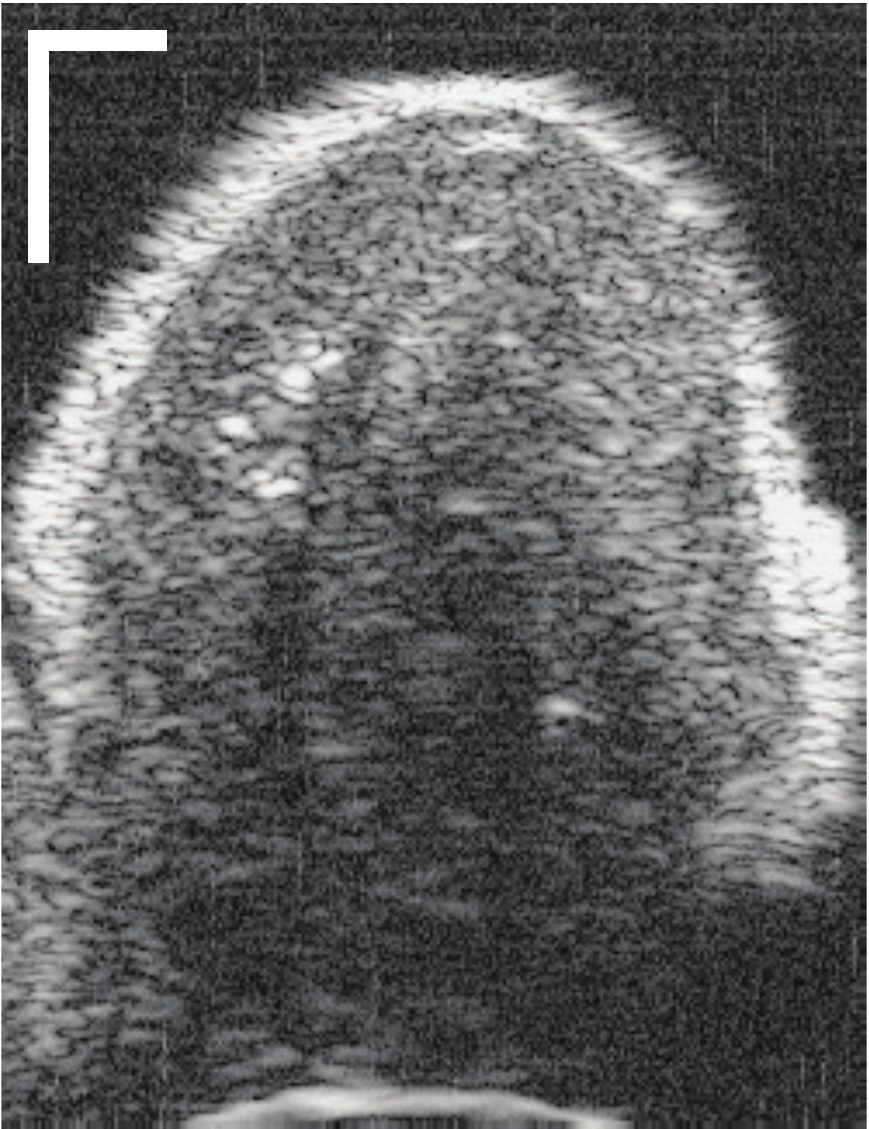}
                 \centerline{(a)}
	\end{minipage} 
	\begin{minipage}[t]{.32\linewidth}
		\includegraphics[width=\linewidth, height=\linewidth]{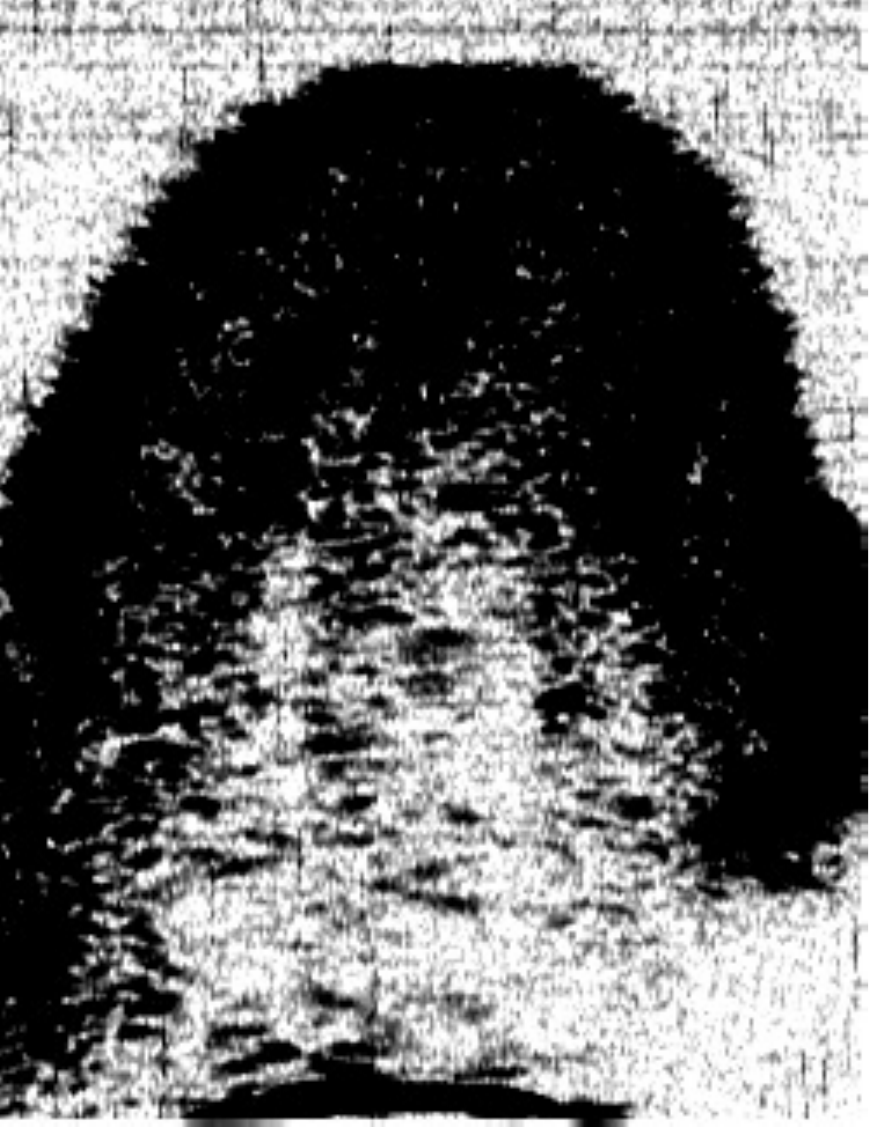}
                 \centerline{(b)}
	\end{minipage} 
	\begin{minipage}[t]{.32\linewidth}
		\includegraphics[width=\linewidth, height=\linewidth]{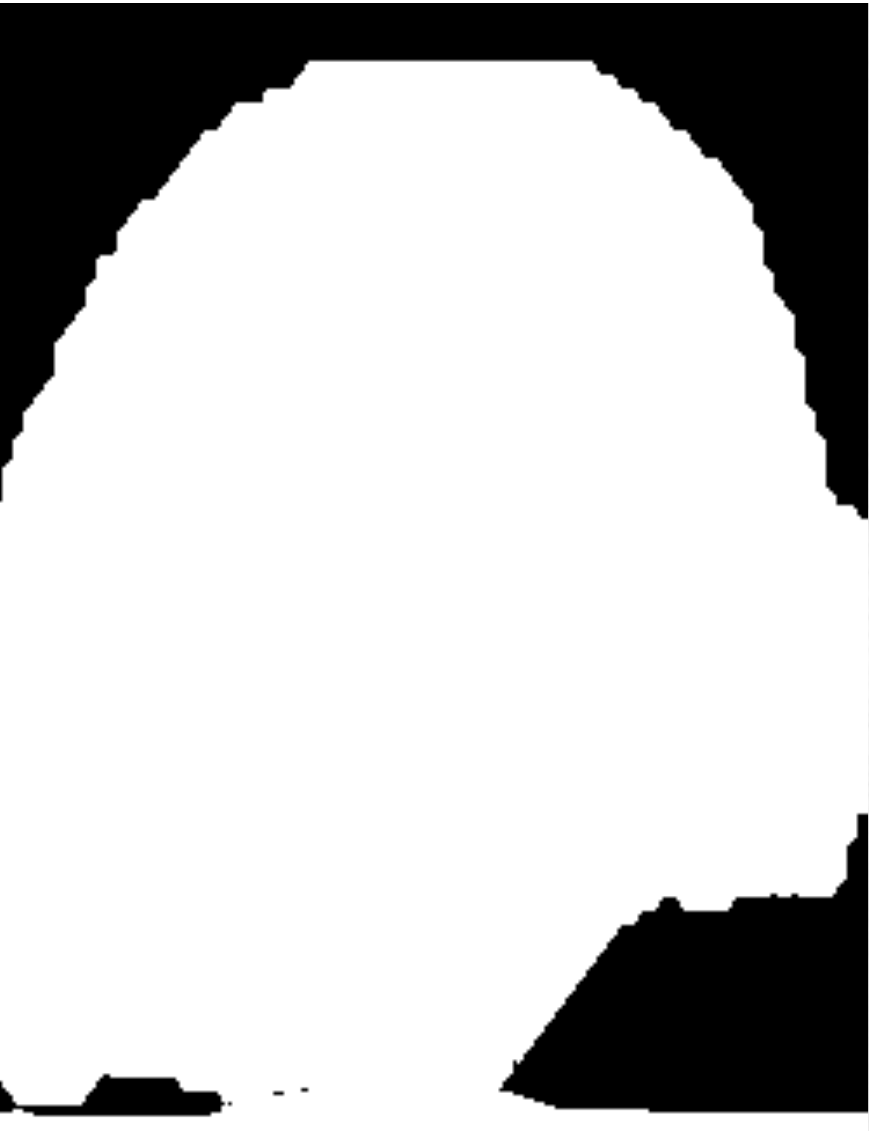}
                 \centerline{(c)}
	\end{minipage} 
	\begin{minipage}[t]{.32\linewidth}
		\includegraphics[width=\linewidth, height=\linewidth]{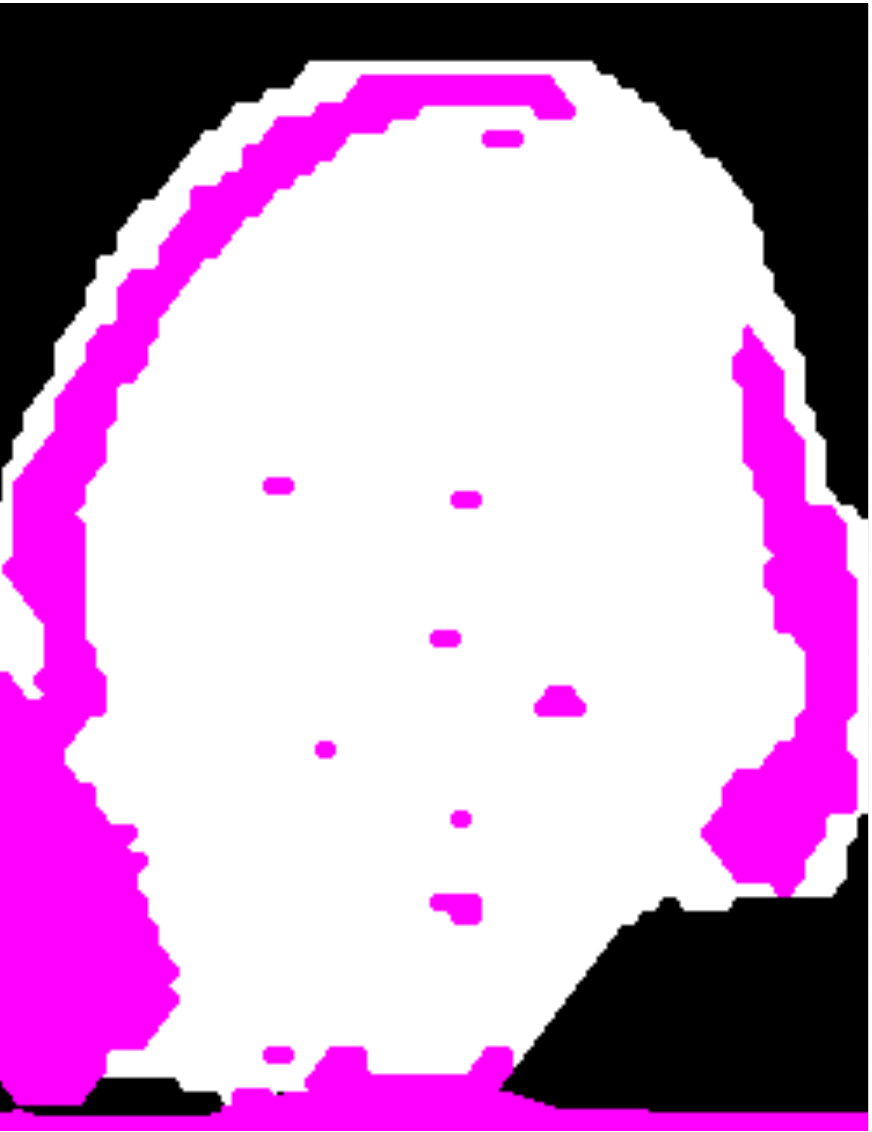}
                 \centerline{(d)}
	\end{minipage} 
	\begin{minipage}[t]{.32\linewidth}
		\includegraphics[width=\linewidth, height=\linewidth]{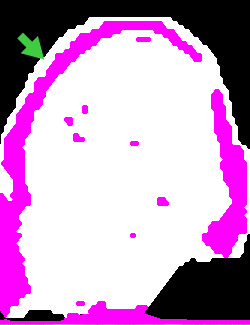}
                 \centerline{(e)}
	\end{minipage} 
	\begin{minipage}[t]{.32\linewidth}
		\includegraphics[width=\linewidth, height=\linewidth]{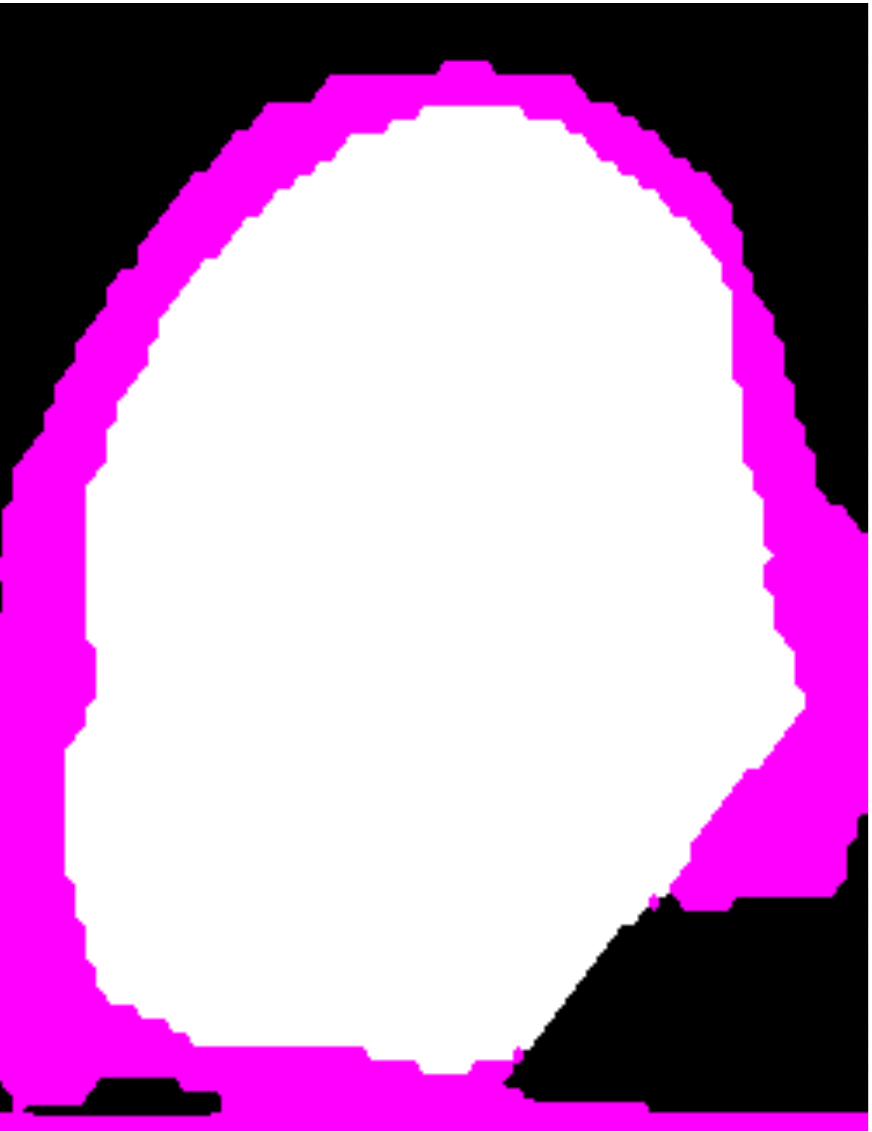}
                 \centerline{(f)}
	\end{minipage} 
	\caption{The images corresponding to steps listed in Figure \ref{fig:flow}. (a) Target HFU image. (b) The probability of belonging to PBS for all voxels. (c) The LN-mask obtained by NGC with the estimated distribution of PBS. (d)The segmentation result of GC-LAE in the first round when $k=3$. (e) The segmentation result of GC-LAE with global energy when $k=3$. (f) The final result obtained by using NGC to refine (e).}
\label{fig:result_flow}
\end{figure}

To determine the probability that a voxel belongs to the PBS, we first calculated the mean over all voxels. Because the PBS voxels are typically darker than LNP and fat voxels, a true PBS voxel is likely to have an intensity much lower than the overall mean value. Therefore, we set the probability to zero for all voxels whose intensities are above the overall mean. For a voxel whose intensity is lower than the overall mean, we set the probability $P(I_p)$ to be proportional to the difference $d(I_p)$ between the mean and the intensity of voxel $p$, using 

\[
	P(I_p)=d(I_p)/\max_{j\in\mathcal{U}}d(I_j)
\]
where $\mathcal{U}$ is the set of voxels $j$ with $d(I_j)>0$. Figure \ref{fig:result_flow}(b) shows the probability defined by the intensity difference. In each sampling, we chose 20$\%$ of voxels in $\mathcal{U}$. In our experiment, we conducted 5 random samplings and chose the largest inlier set from these 5 samplings to estimate the mean intensity and STD of PBS voxels. 

\subsubsection{Segment PBS by NGC}
\label{sec:segment PBS by NGC}

In Figure \ref{fig:result_flow}(a), LNP (dark and light) is inside the fat (light), and these two objects are inside the PBS (dark). In this example, NGC limited the region of LNP inside the convex hull of the fat and defined the missing boundary between LNP and PBS as a straight line (illustrated in Figure \ref{fig:result_flow}(c)). Using the conventional GC approach would have falsely labeled the lower part of LNP as PBS. We briefly describe the NGC algorithm below. A more-detailed description can be found in \cite{kuo2016nested}.

NGC is designed to simultaneously segment multiple objects with a nested relationship, i.e., an inner object is contained in an outer object, and NGC only needs to have the intensity distribution of each object. (which can be either global distributions or locally adaptive  location-dependent distributions, e.g. depth dependent). NGC applies duplicated layers to represent different objects. Each node in layer $i$ corresponds to a voxel in the image and has an associated binary variable $x_p^i$, with $x_p^i=1$ indicating voxel $p$ is assigned to region $i$, $x_p^i=0$ otherwise. To determine $x_p^i$, we minimize the following energy function: 

\[
\begin{split}
	E(x_p^i, i\in\mathcal{L}, p\in\mathcal{P}) = \sum_{i\in\mathcal{L}}\big(&{\sum_{p\in\mathcal{P}}{D_p^i(x_p^{i-1},x_p^i)}} \\
	&+{\alpha_i\sum_{p,q\in\mathcal{N}}{V_{p,q}^i(x_p^{i},x_q^i)}} \big)
\end{split}
\]

\begin{equation*}
	D_p^i(x_p^{i-1}, x_p^i) =
	\begin{dcases}
		0  &  x_p^i = x_p^{i-1} \\
		W_p^{i-1,i}  &  x_p^i \neq x_p^{i-1}
	\end{dcases}  \\
	\label{eq:dataterm}
\end{equation*}

\begin{equation}
	W_p^{i-1,i}(I_p) =
  	\begin{dcases}
    		\frac{1}{1+\exp(\frac{I_p-I^T}{\sigma})}  & i \text{ is PBS} \\
    		\frac{\exp(\frac{I_p-I^T}{\sigma})}{1+\exp(\frac{I_p-I^T}{\sigma})}, & i \text{ is fat} \\
		\frac{\min\big(1,\exp(\frac{I_p-I^T}{\sigma})\big)}{1+\exp(\frac{I_p-I^T}{\sigma})}, & i \text{ is LNP}
	\end{dcases} 
	\label{eq:threshold2}
\end{equation}

\begin{equation}
	V_{p,q}^i(x_p^i, x_q^i) =
	\begin{dcases}
		0  &  x_p^i = x_q^i \\
		1 &  x_p^i \neq x_q^i
	\end{dcases}  \\ 
	\label{eq:regterm}
\end{equation}\\ 
where $\mathcal{L}$ denotes the set of layer indices, $\mathcal{P}$ is the set of voxels, and $\mathcal{N}$ defines neighborhood set (we used 18-connectivity to define the neighborhood structure in 3D). $V^i$ and $D^i$ stand for the regularization term and data term, respectively. We define $V_{p,q}^i$ as in (\ref{eq:regterm}) so that the total regularization cost for layer $i$ is the total boundary length of region $i$. Minimizing such a length will lead to smoother boundaries. Data term $D_p^i$ is defined as the log likelihood that pixel intensity $I_p$ belongs to object $i$. Because the goal here is to separate PBS from target HFU image, we only consider the case where each object has either high or low intensity. 

As illustrated in Figure \ref{fig:result_flow}(a), the intensity of PBS is ``low'', the intensity of the fat is ``high,'' and the lower part of LNP is ``low'' and the higher part of LNP is ``high.'' Therefore, we define the energy terms by a sigmoid function as in Eq. (\ref{eq:threshold2}). The parameters $I^T$ and $\sigma$ are determined based on the observed intensity distribution of PBS in the previous step. We use Gaussian distribution to model PBS. In (\ref{eq:threshold2}), PBS threshold $I^T$ is defined as $\mu_{PBS}+3\sigma_{PBS}$, and $\sigma$ is defined as $\sigma_{PBS}/2$. Figure \ref{fig:result_flow}(c) shows the segmentation result obtained using NGC. The white region contains LNP and fat, and the PBS is labeled in black. The details of NGC can be found in \cite{kuo2016nested}.

\subsection{LNP Segmentation from the LN-Mask}

In LN images, the intensity distribution of a given object may not be uniform even at a constant depth. To deal with the spatially varying intensity distributions, we apply the GC-LAE method described in Sec. \ref{sec:GCLAE} to the LN-mask. The parameter $k$ should be set based on the degree of intensity dissimilarity between LNP and fat. Our experience revealed that using a fixed $k$ value led to difficulties in obtaining satisfactory segmentation results for images with non-uniform contrast variations. Therefore, we performed GC-LAE multiple times using several $k$ values, and fuse these results to obtain the confident regions of LNP and fat. Finally, we applied NGC to obtain the final segmentation result. The method to automatically select $k$ values is described in Sec. \ref{subsec:determine k}.

\begin{figure}[t]
	\centering
	\begin{minipage}[t]{.24\linewidth}
		\includegraphics[width=\linewidth, height=\linewidth]{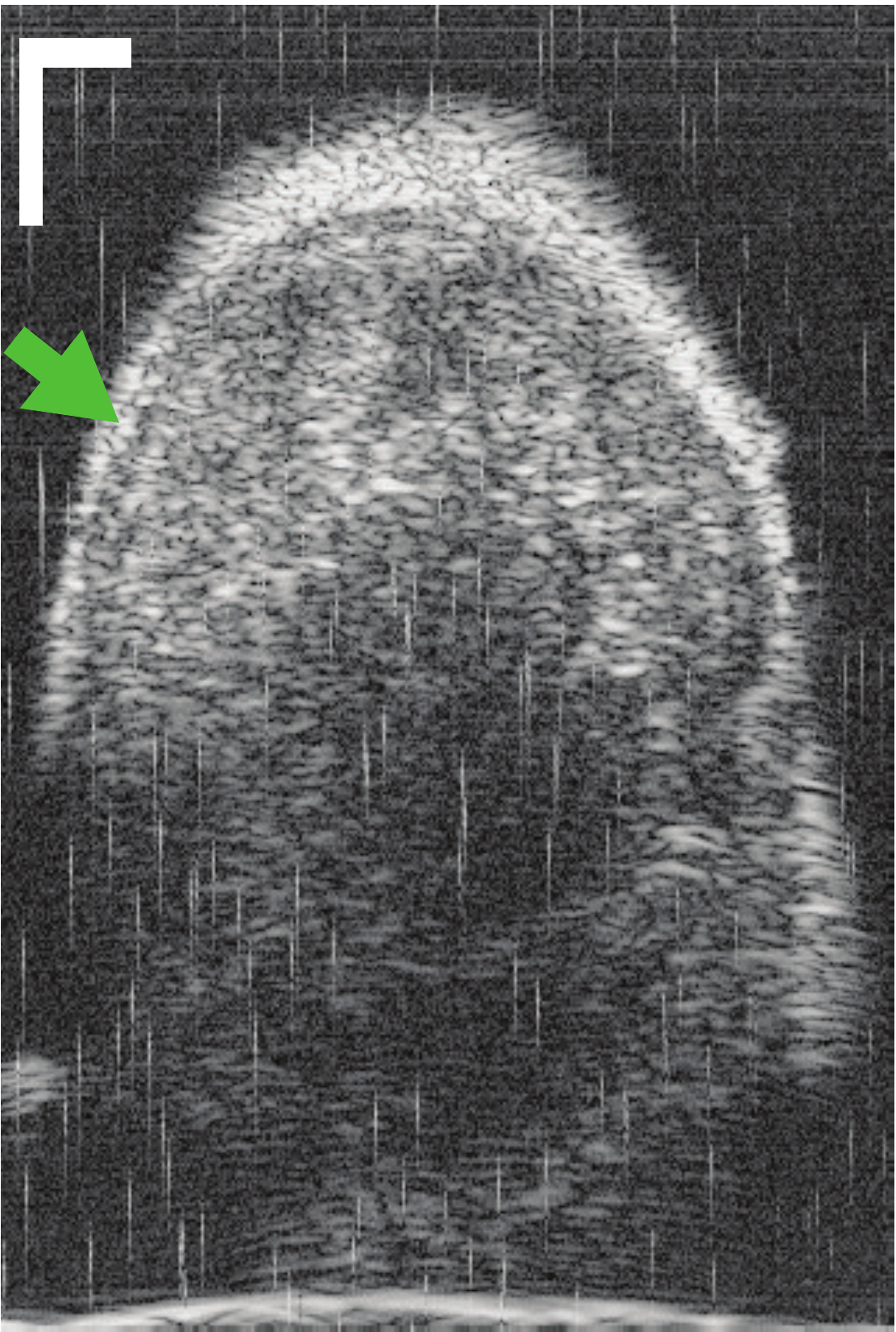}
	\end{minipage} 
	\begin{minipage}[t]{.24\linewidth}
		\includegraphics[width=\linewidth, height=\linewidth]{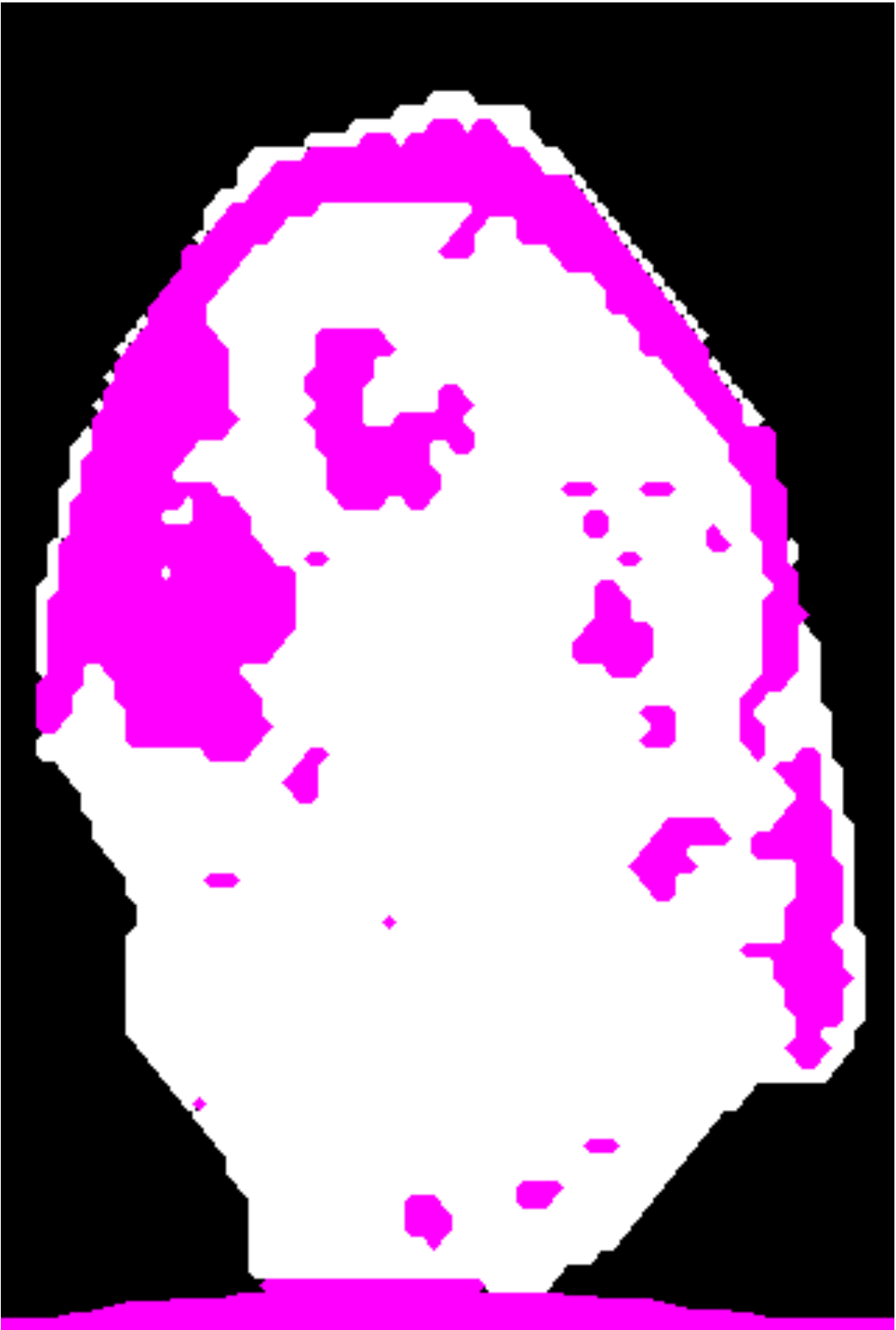}
	\end{minipage} 
	\begin{minipage}[t]{.24\linewidth}
		\includegraphics[width=\linewidth, height=\linewidth]{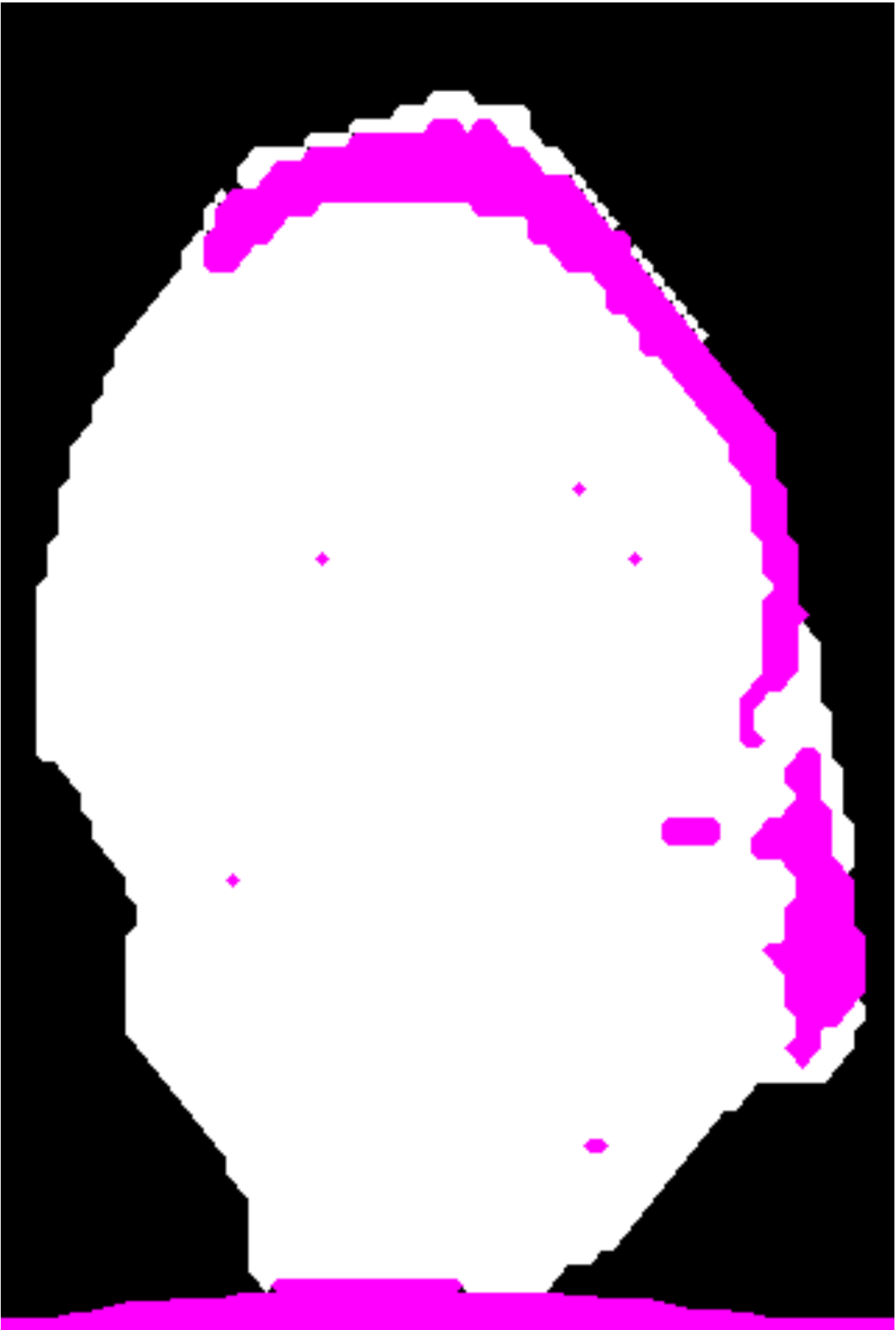}
	\end{minipage} 
	\begin{minipage}[t]{.24\linewidth}
		\includegraphics[width=\linewidth, height=\linewidth]{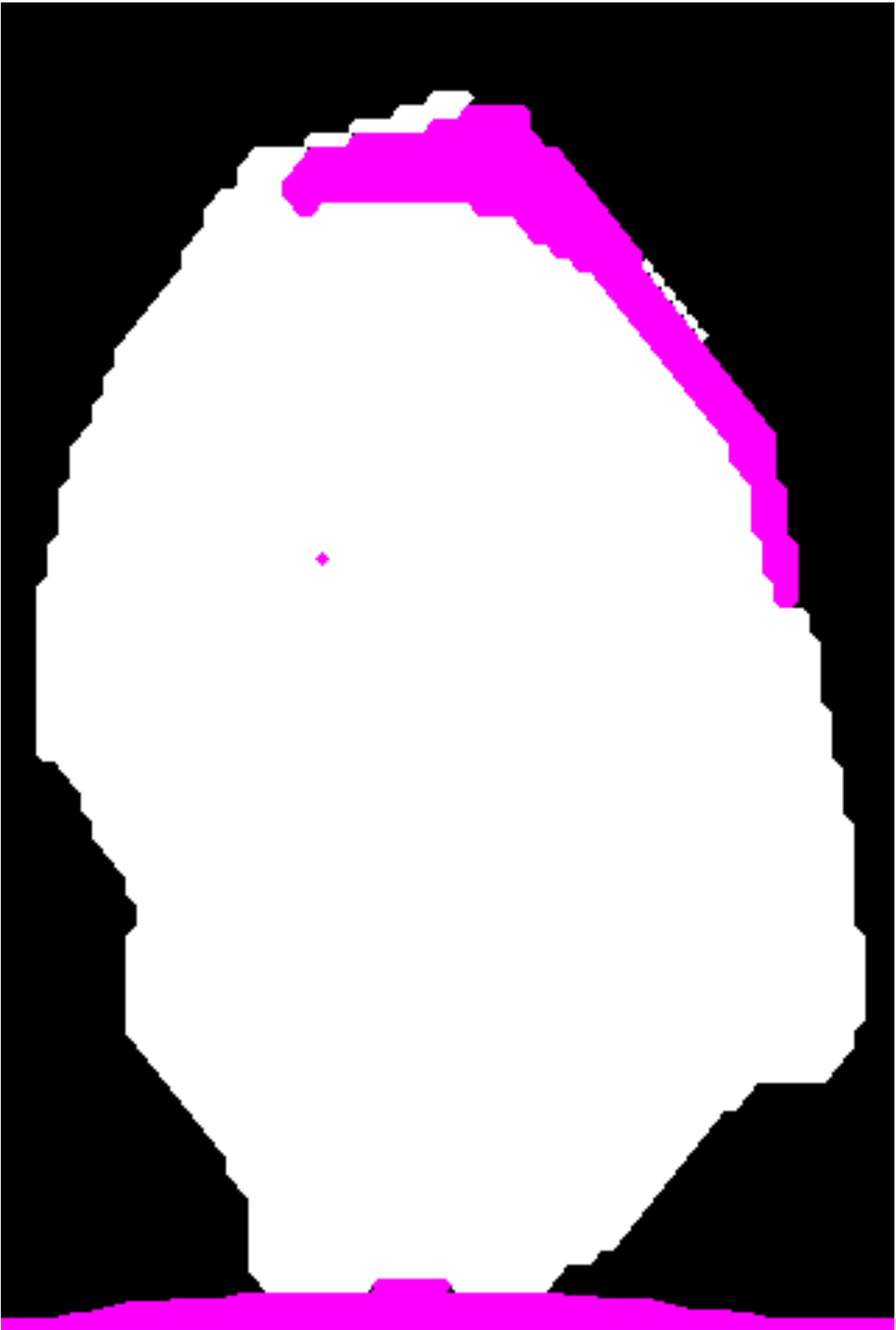}
	\end{minipage} 
	\begin{minipage}[t]{.24\linewidth}
		\includegraphics[width=\linewidth, height=\linewidth]{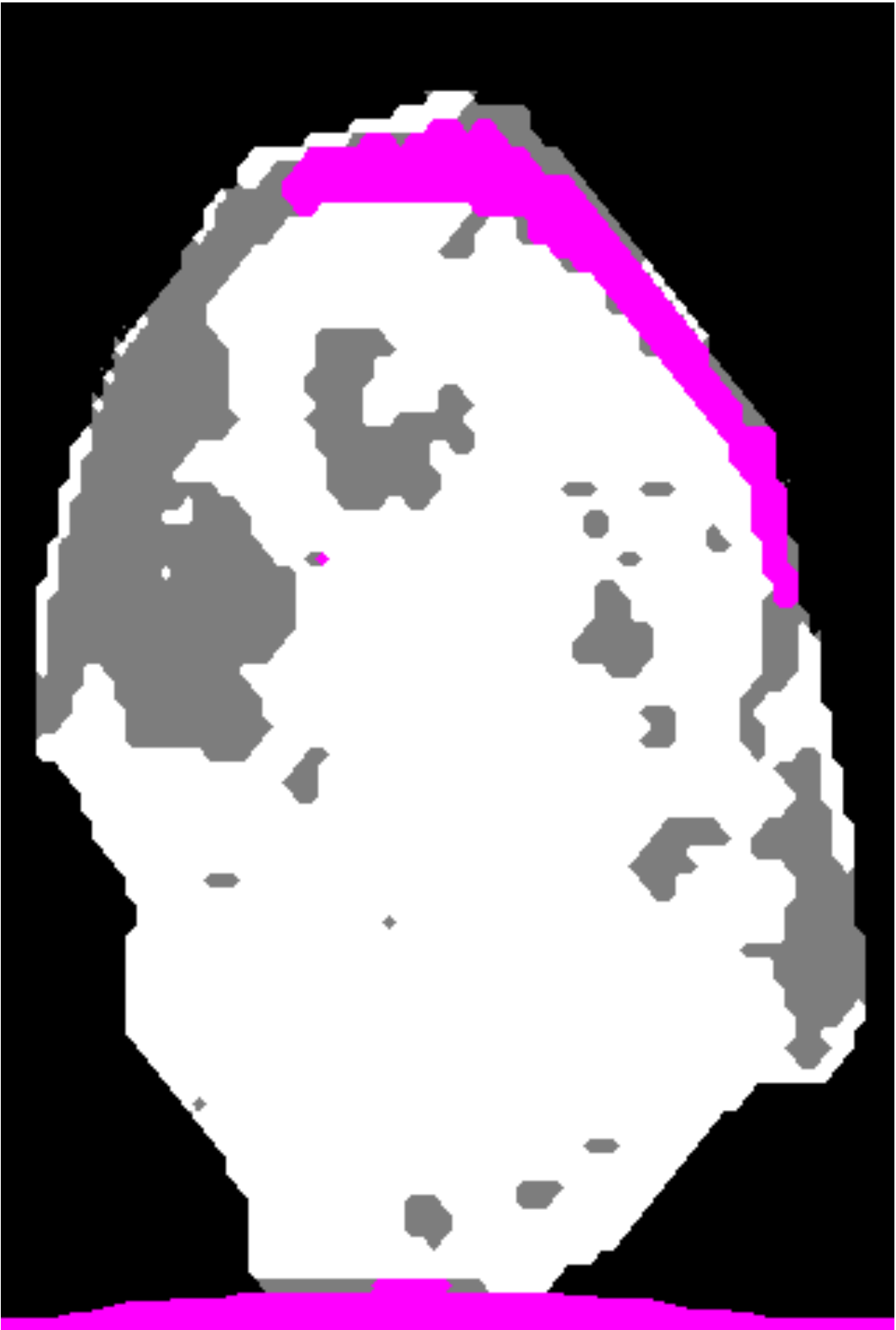}
		\centerline{(a)}
	\end{minipage} 
	\begin{minipage}[t]{.24\linewidth}
		\includegraphics[width=\linewidth, height=\linewidth]{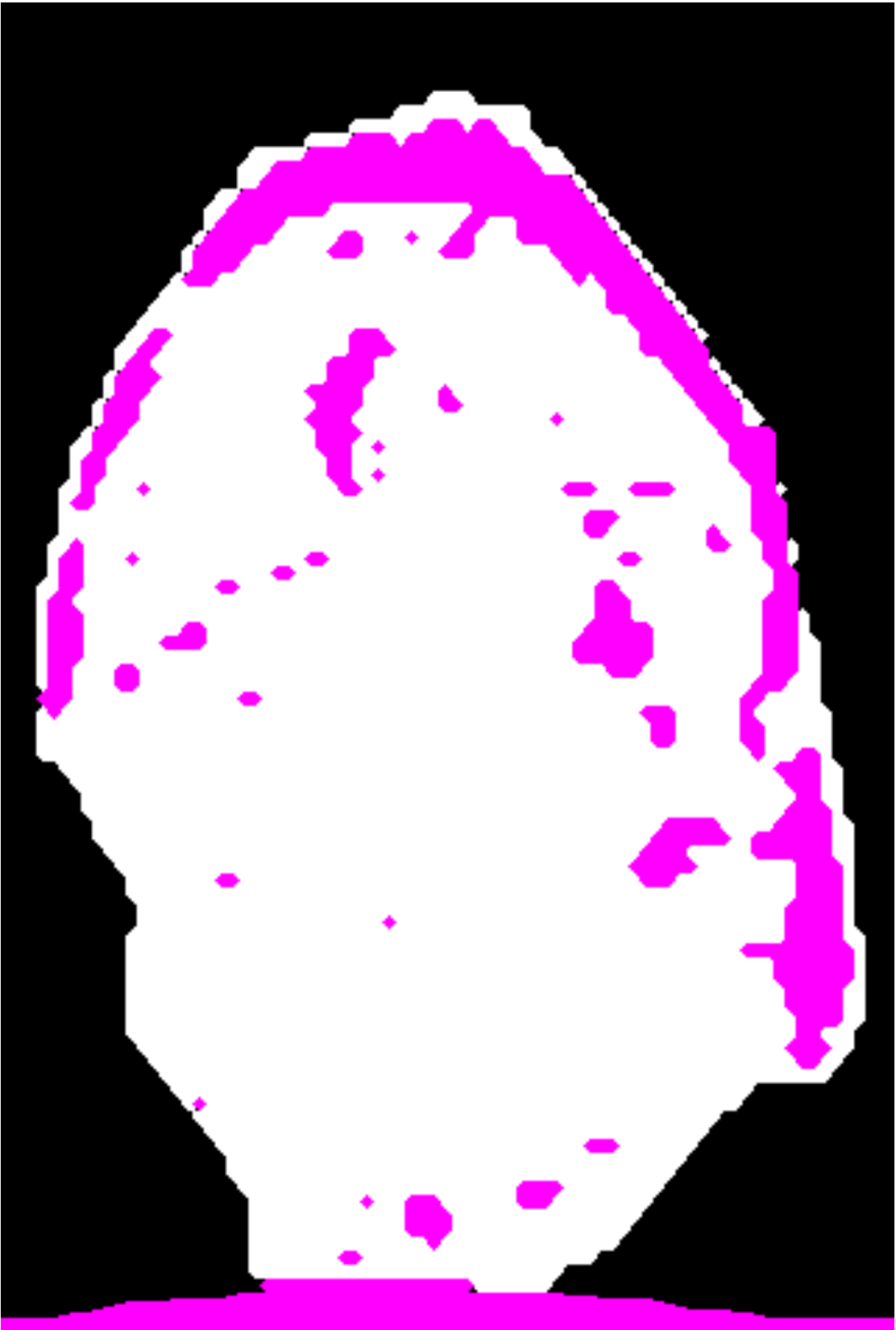}
                 \centerline{(b)$k=2$}
	\end{minipage} 
	\begin{minipage}[t]{.24\linewidth}
		\includegraphics[width=\linewidth, height=\linewidth]{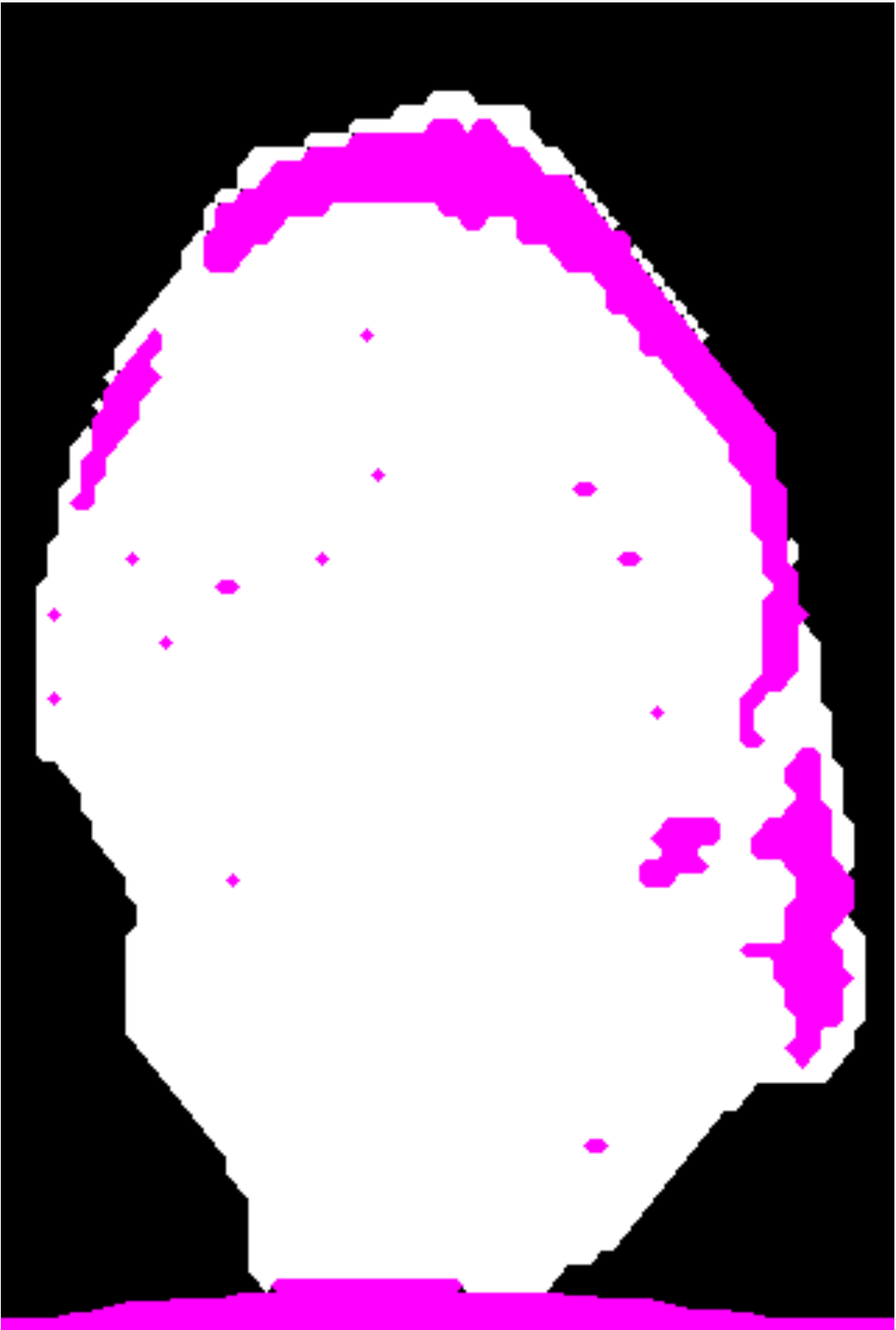}
                 \centerline{(c)$k=2.5$}
	\end{minipage} 
	\begin{minipage}[t]{.24\linewidth}
		\includegraphics[width=\linewidth, height=\linewidth]{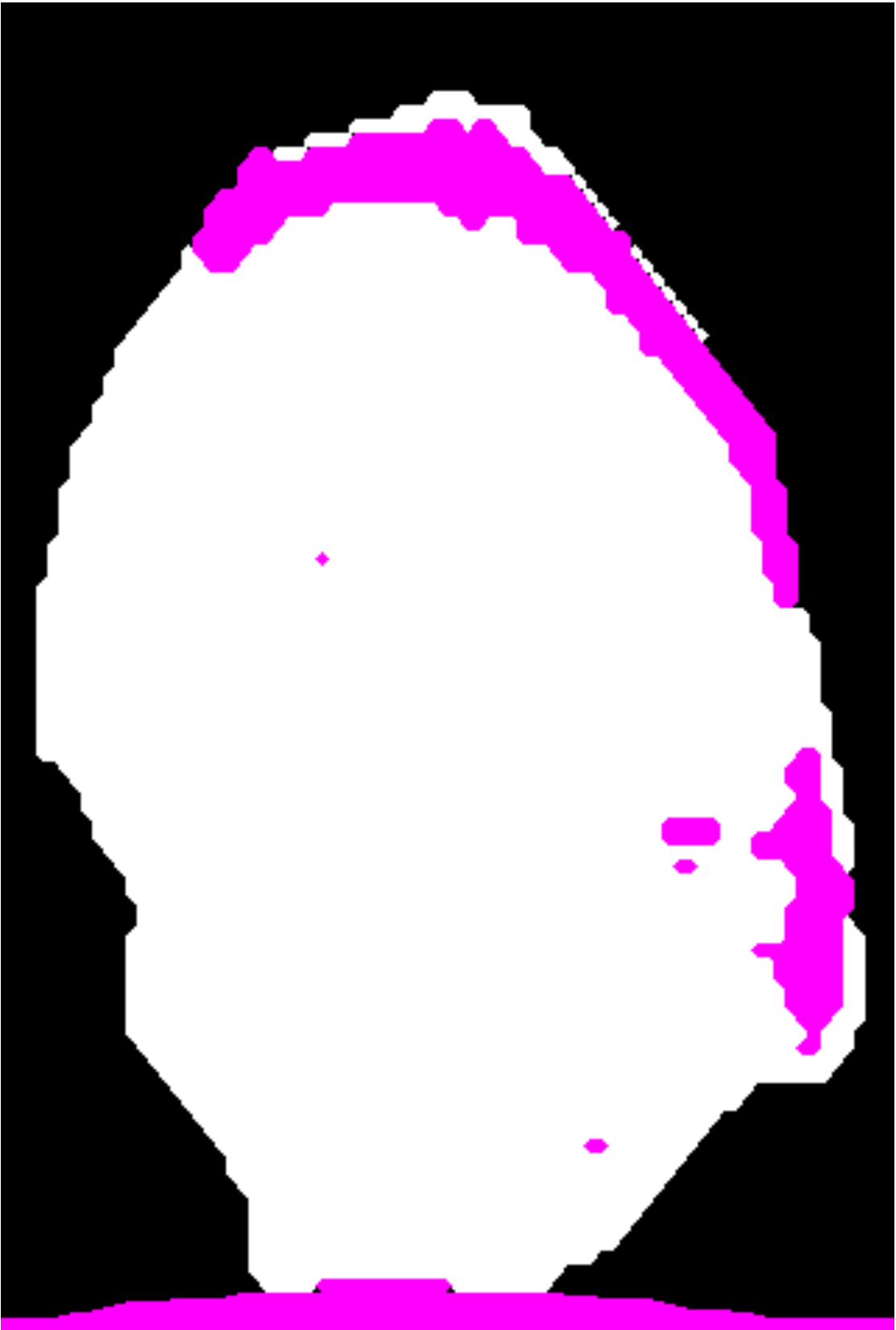}
                 \centerline{(d)$k=3$}
	\end{minipage} 
	\caption{(a) The HFU image and the confident regions obtained by fusing results in the top rows of (b),(c), and (d). (b)(c)(d) Top row shows the segmentation results obtained by GC-LAE with different values of $k$ in the first round (no global energy). Bottom row shows the segmentation results obtained by GC-LAE in the second round with different k values (with global energy).}
\label{fig:compare_GCLAE}
\end{figure}

\subsubsection{Segment LNP and fat by GC-LAE}

In the proposed framework, we apply GC-LAE in two rounds. In the first round, GC-LAE is applied with the local adaptive energy only. The top row in Figure \ref{fig:compare_GCLAE} shows the segmentation results of GC-LAE obtained by several different $k$ in the first round. Because the contrast between LNP and the fat at left hand side (green arrow) is low, GC-LAE may segment some LNP voxels as the fat region when $k$ is small or segment all voxels as LNP when $k$ is high. 

As demonstrated, there is not a single value for $k$ that will work well in all regions. Therefore we apply GC-LAE with several different $k$ values, and then fuse the results. The automatic selection of up to three k values is described in Sec. \ref{subsec:determine k}. We use the intersection of the LNP regions obtained with different $k$ as the confident region for LNP (white region in Figure \ref{fig:compare_GCLAE}(a) bottom), and we use the intersection of the fat regions as the confident region for fat (pink region in Figure \ref{fig:compare_GCLAE}(a)). From these confident regions, we determine the mean and STD of LNP and the fat at different depths. To suppress the noise in the resulting mean and STD values, we further fit the mean and STD profiles along the depth into two separate spline functions using the spline-based approach with RANSAC \cite{fischler1981random}.

In the second round, GC-LAE is applied with local adaptive-energy term and the global energy term (Eq. (\ref{eq:threshold})). The global energy is defined by the depth-dependent profiles obtained from the confident regions in the first round. The bottom row of Figure \ref{fig:compare_GCLAE} shows the segmentation results of GC-LAE with global energy which have more accurate boundary on the left. Similar to the first round of GC-LAE, we can fuse these results to obtain new confident regions for both LNP and fat, and derive the new depth-dependent mean and STD profiles. 

\subsubsection{Update PBS threshold}
\label{sec:update PBS}

Before applying NGC for the final refinement, the depth-dependent PBS threshold $I^T_p$ in (\ref{eq:threshold2}) is defined as $\mu_{LNP,p}-\sigma_{LNP,p}$ and $\sigma_p$ in (\ref{eq:threshold2}) is defined as $\sigma_{LNP,p}/2$.

\begin{figure}[t]
	\centering
	\begin{minipage}[t]{.24\linewidth}
		\includegraphics[width=\linewidth, height=\linewidth]{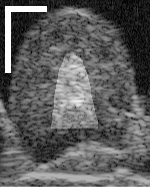}
	\end{minipage} 
	\begin{minipage}[t]{.24\linewidth}
		\includegraphics[width=\linewidth, height=\linewidth]{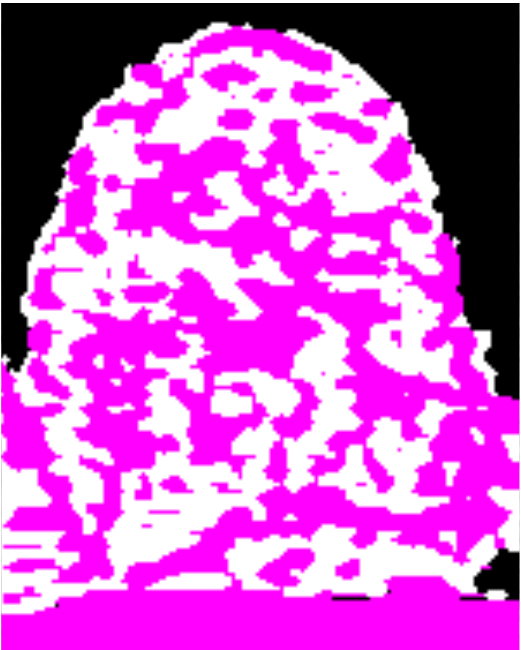}
                 \centerline{$ratio=0.57$}
	\end{minipage} 
	\begin{minipage}[t]{.24\linewidth}
		\includegraphics[width=\linewidth, height=\linewidth]{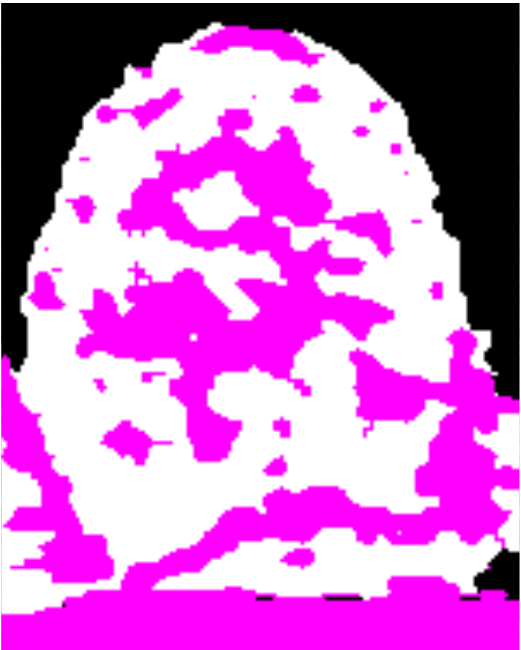}
                 \centerline{$ratio=0.53$}
	\end{minipage} 
	\begin{minipage}[t]{.24\linewidth}
		\includegraphics[width=\linewidth, height=\linewidth]{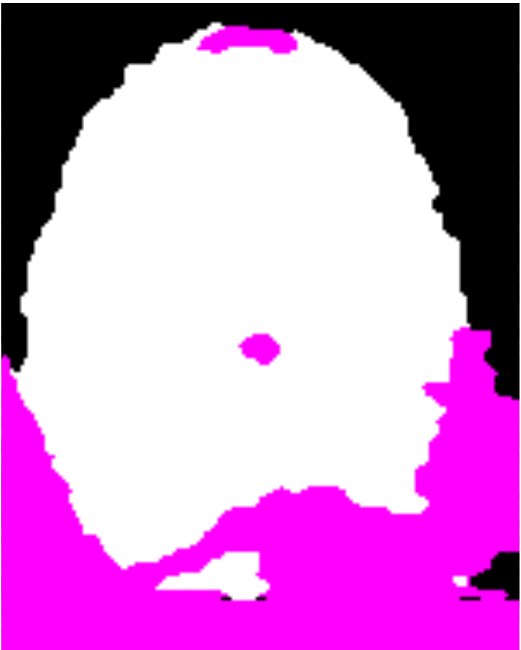}
                 \centerline{$ratio = 0.02$}
	\end{minipage} 
	\begin{minipage}[t]{.24\linewidth}
		\includegraphics[width=\linewidth, height=\linewidth]{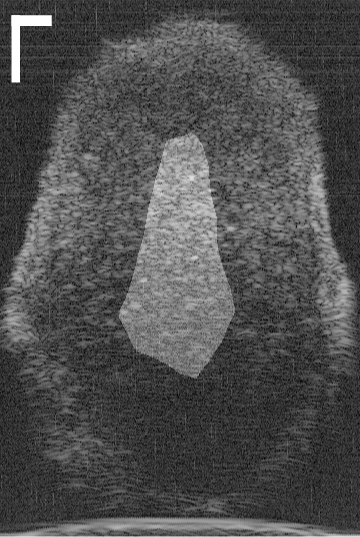}
	\end{minipage} 
	\begin{minipage}[t]{.24\linewidth}
		\includegraphics[width=\linewidth, height=\linewidth]{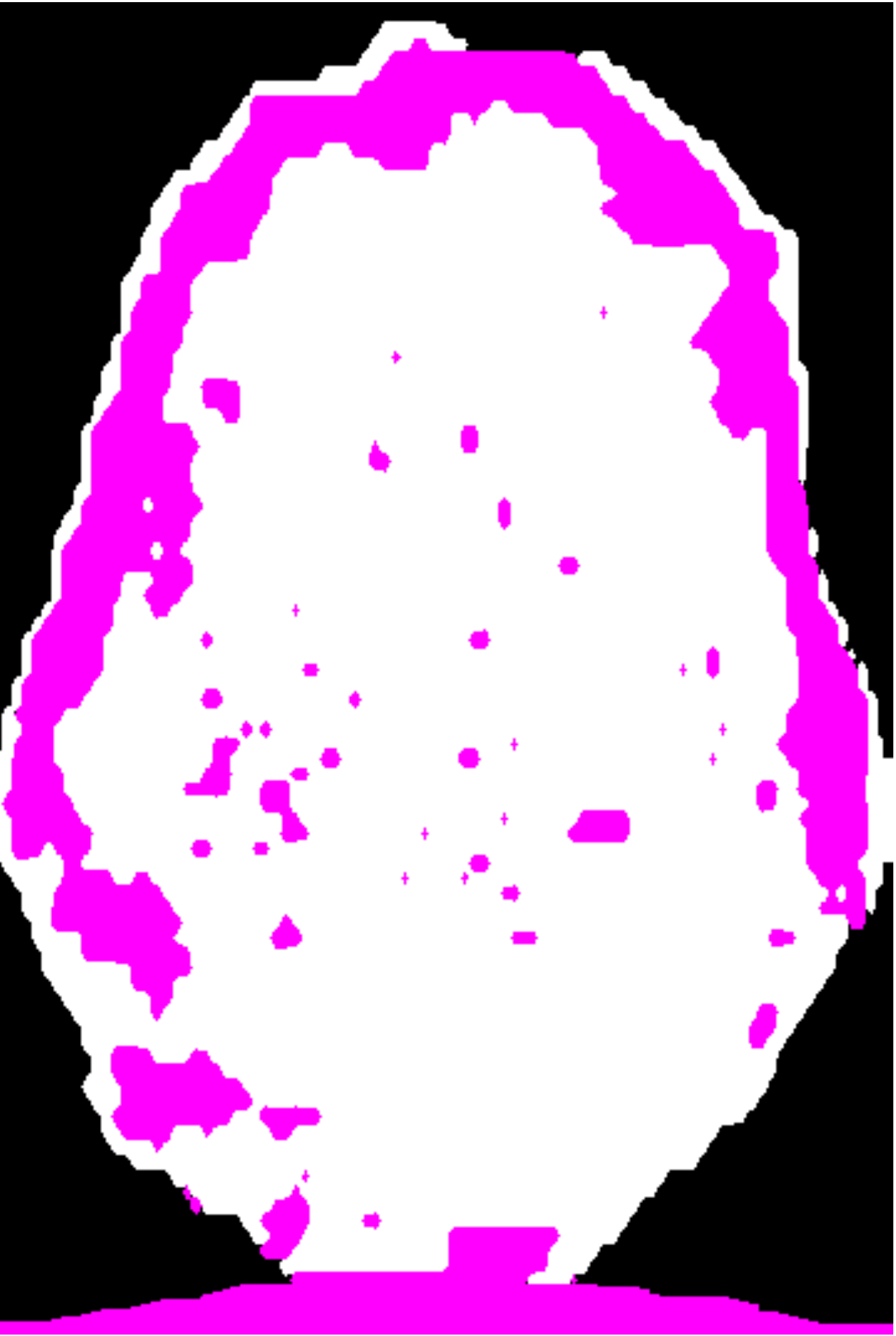}
		\centerline{$ratio=0.07$}
                 \centerline{$k=1.5$}
	\end{minipage} 
	\begin{minipage}[t]{.24\linewidth}
		\includegraphics[width=\linewidth, height=\linewidth]{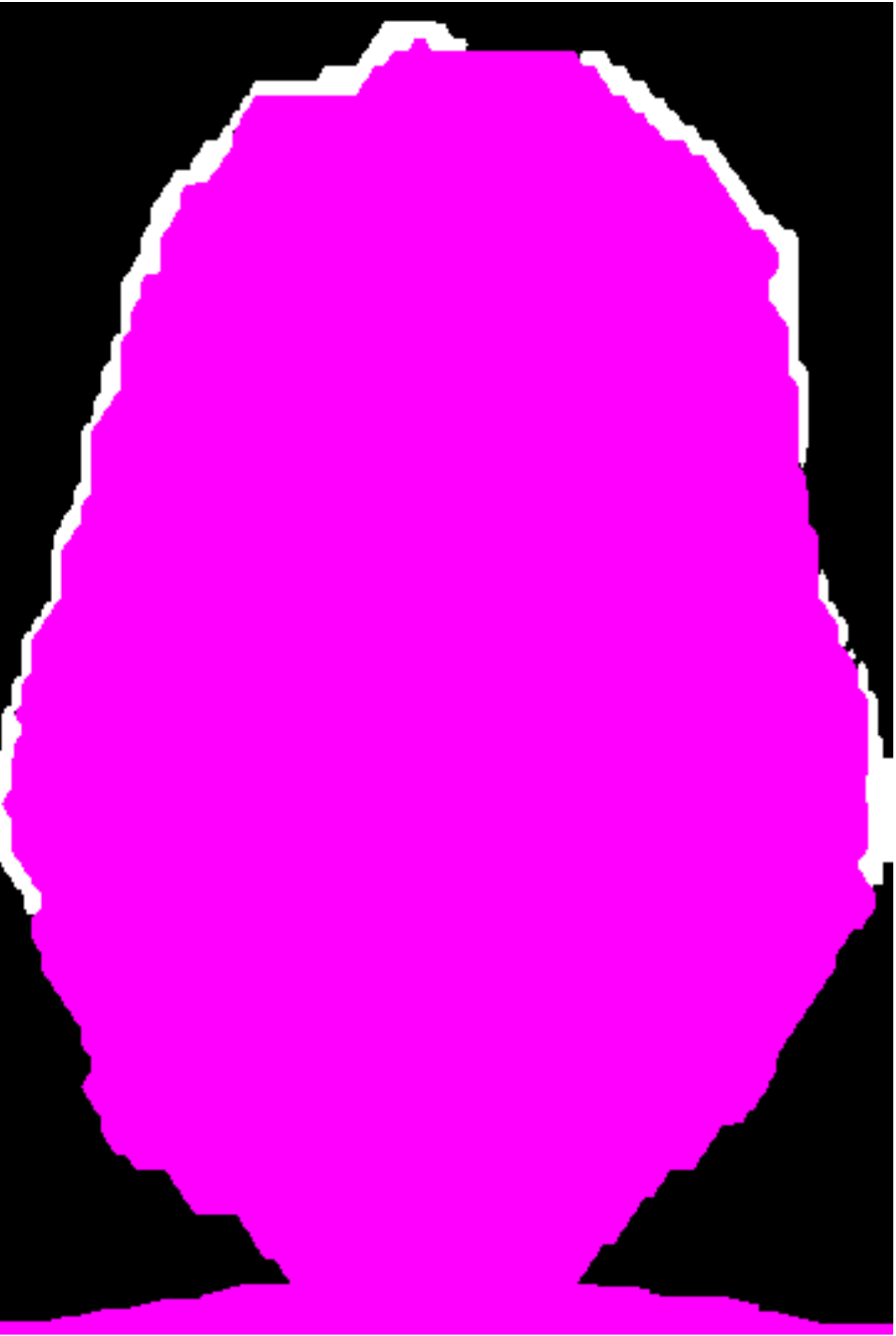}
		\centerline{$ratio=1$}
                 \centerline{$k=2.5$}
	\end{minipage} 
	\begin{minipage}[t]{.24\linewidth}
		\includegraphics[width=\linewidth, height=\linewidth]{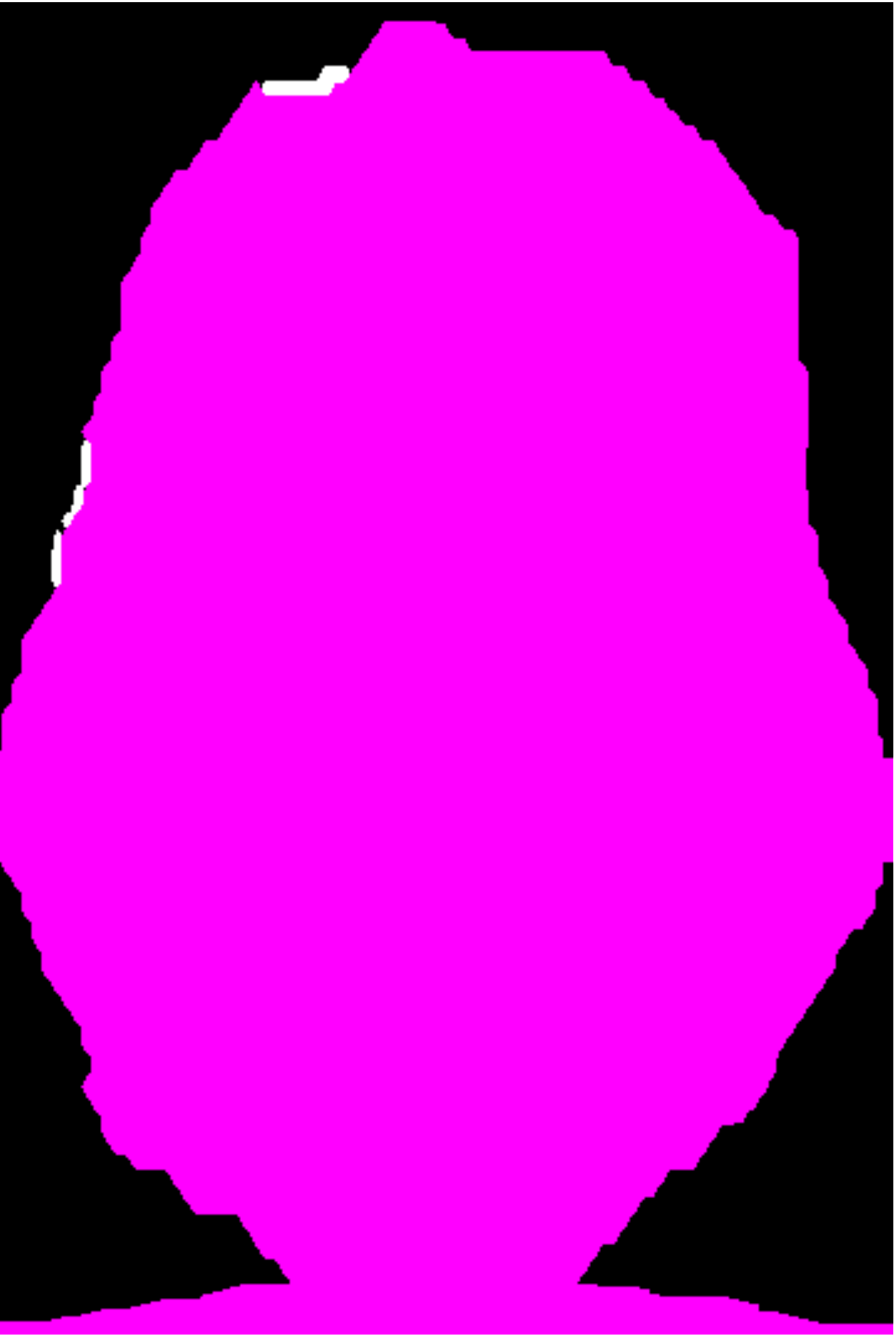}
		\centerline{$ratio=1$}
                 \centerline{$k=3.5$}
	\end{minipage} 
	\caption{The HFU image and its segmentation results obtained by GC-LAE with different $k$ values. The height of top LN is 2.7 mm, and the height of the bottom LN is 7.9 mm.}
\label{fig:over_GCLAE}
\end{figure}

\subsection{Automatic Selection of Several $k$ Values}
\label{subsec:determine k}

Small LNs ($<=$3.6 mm) have higher contrast between LNP and fat than large LNs ($>$3.6 mm) because they typically suffer less-severe attenuation effects. For small LNs, a larger $k$ value is preferred to get rid of unwanted noise. For large LNs, a smaller $k$ value is preferred to find all boundaries. Figure \ref{fig:over_GCLAE} shows the segmentation results of GC-LAE for three values of $k$. The top row shows a small LN (2.7 mm), and the bottom row shows a large LN (7.9 mm). As shown in Figure \ref{fig:over_GCLAE}, an improper $k$ value may lead to extremely poor results. In the top LN, the segmentation results of GC-LAE when $k=1.5$ and $k=2.5$ are poor because GC-LAE may be too sensitive to noise when the value of $k$ is too small. In the bottom LN, GC-LAE may be unable to differentiate objects when the value of $k$ is too large. 

In our experiment, we apply GC-LAE with $k$ values from 1 to 4 with interval 0.5, and then we use the fat ratio inside the inner mask to determine proper $k$ automatically. The inner mask is defined as the area with distance over 70\% of the largest distance to the boundary of LN-mask. The bright region in Figure \ref{fig:over_GCLAE} shows the inner masks of corresponding LNs. After getting the inner mask, we compute the fat ratio inside the inner mask. As indicated in Figure \ref{fig:over_GCLAE}, poor results tend to have larger fat ratio. In this experiment, we choose up to 3 $k$ values having fat ratio under 0.32. The automatically selected $k$ values of the top LN are 3, 3.5, and 4 and those of the bottom LN are 1 and 1.5.

\begin{figure}[t]
	\centering
	\begin{minipage}[t]{.24\linewidth}
		\includegraphics[width=\linewidth, height=\linewidth]{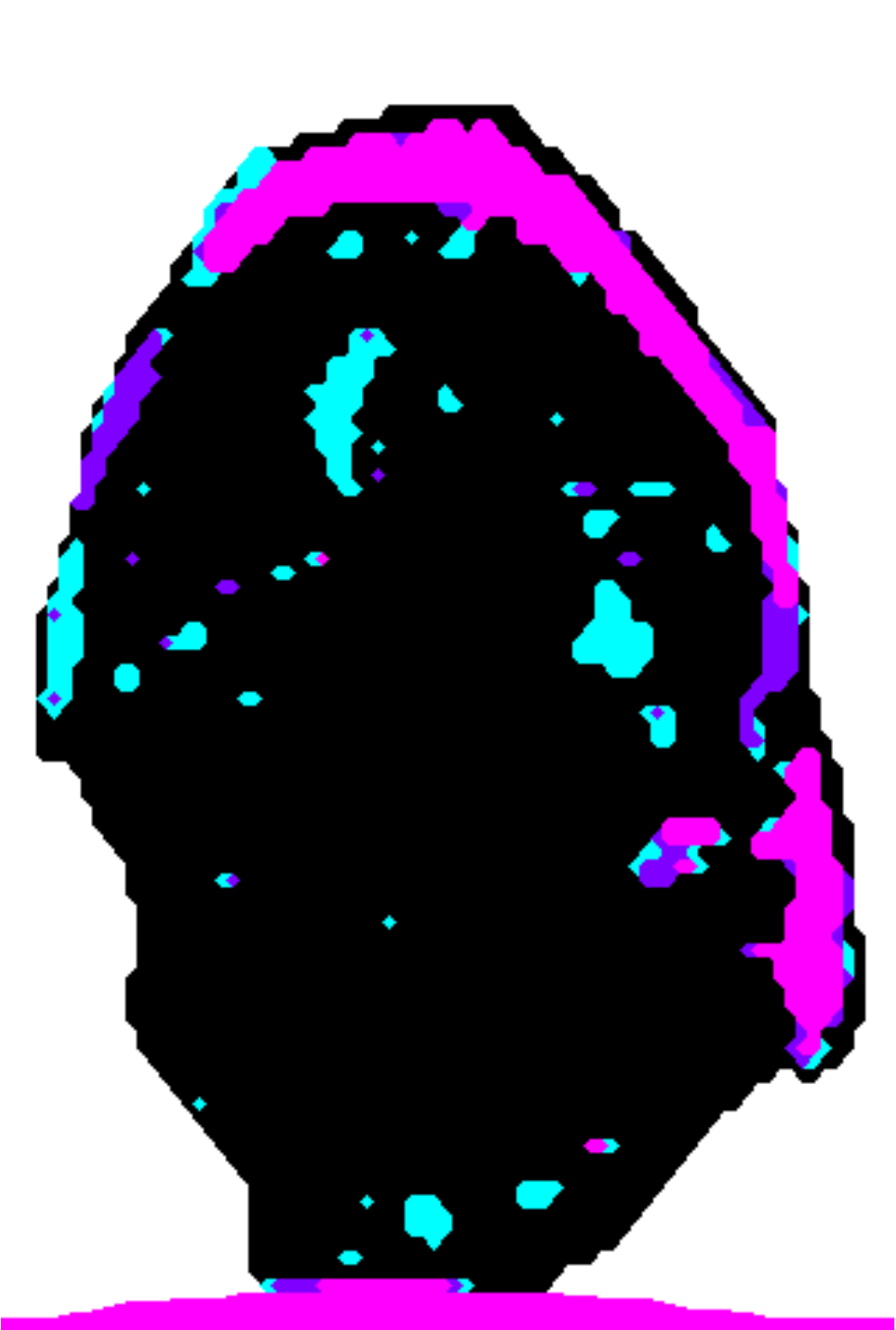}
                 \centerline{(a)}
	\end{minipage} 
	\begin{minipage}[t]{.24\linewidth}
		\includegraphics[width=\linewidth, height=\linewidth]{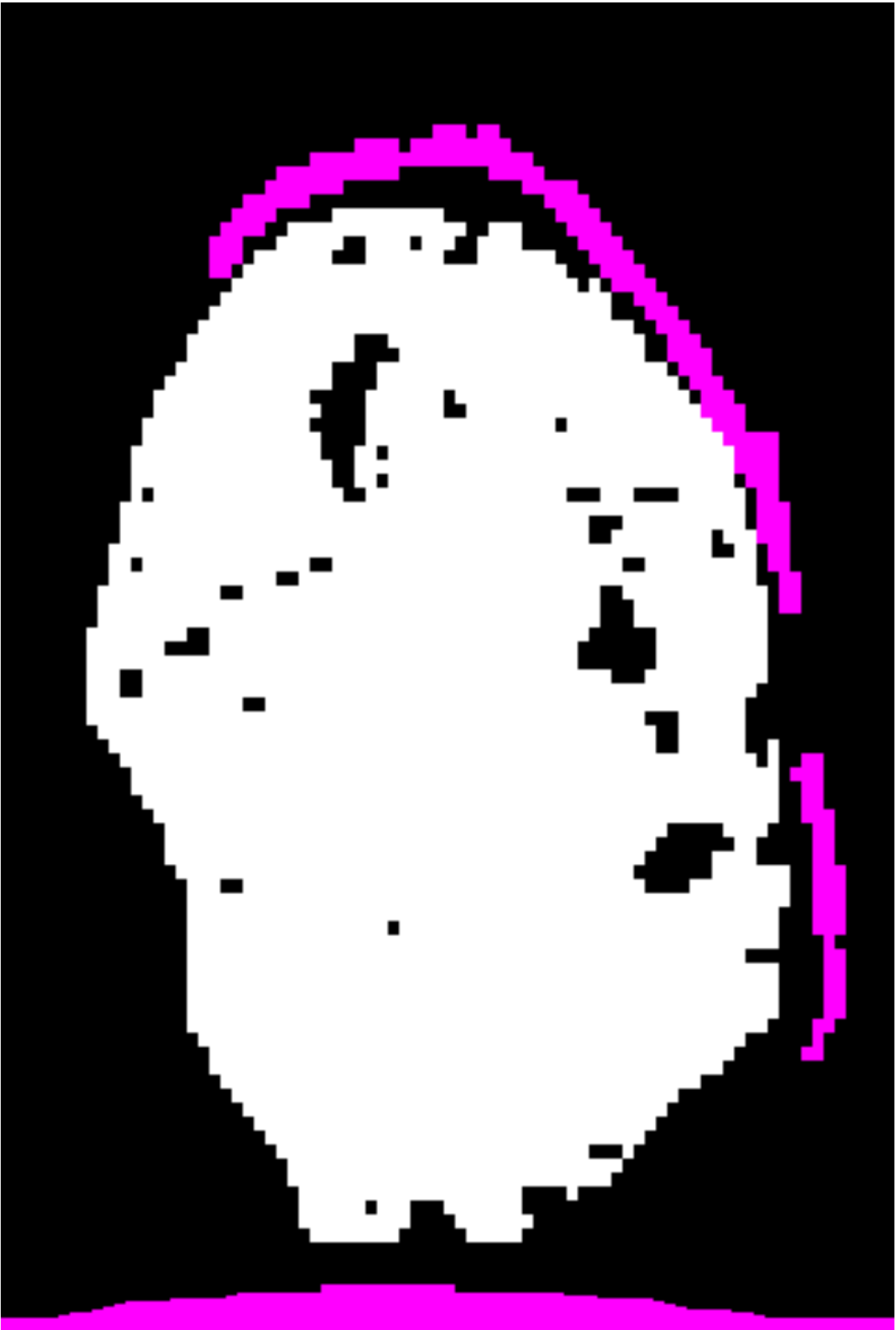}
                 \centerline{(b)}
	\end{minipage} 
	\begin{minipage}[t]{.24\linewidth}
		\includegraphics[width=\linewidth, height=\linewidth]{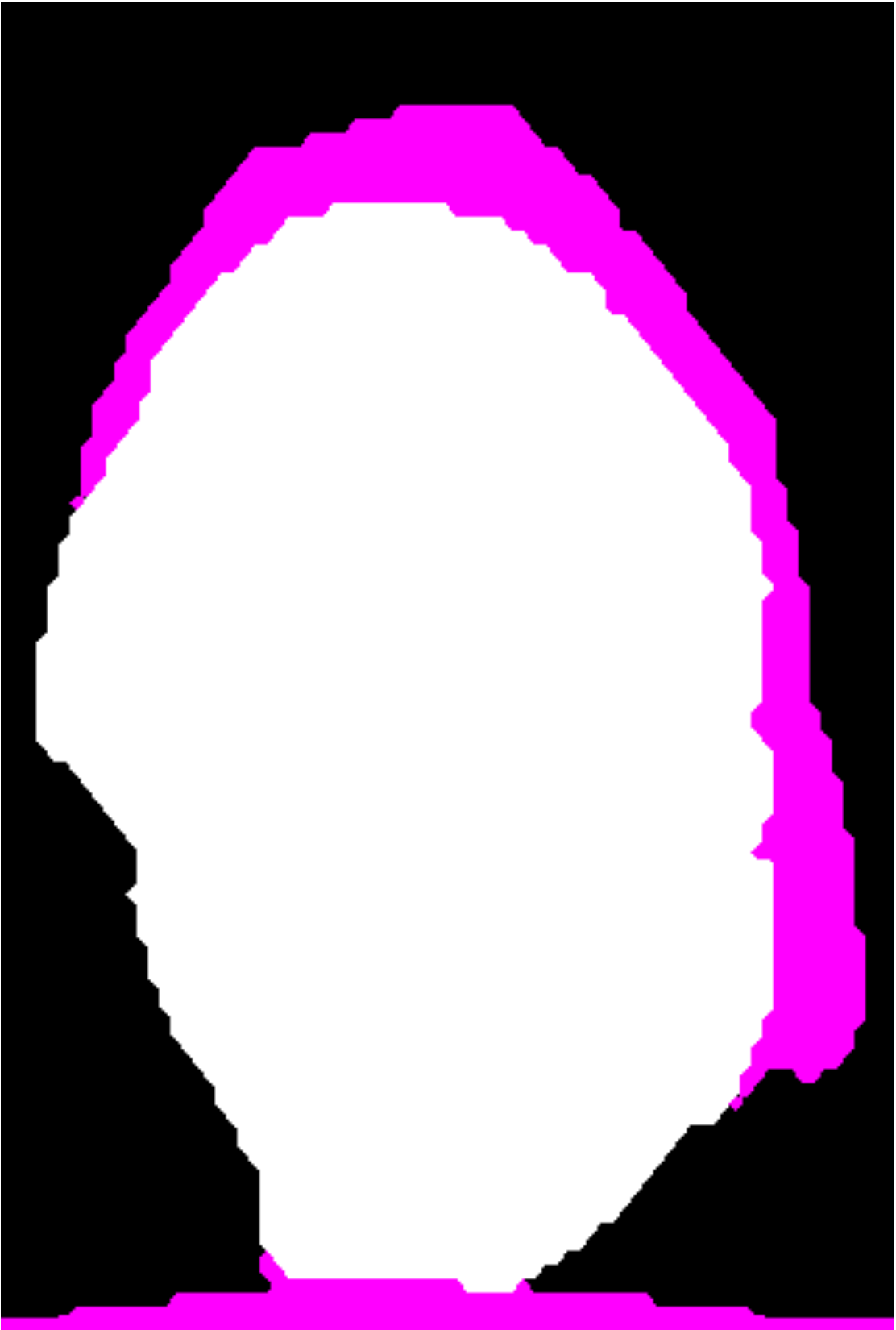}
                 \centerline{(c)}
	\end{minipage} 
	\begin{minipage}[t]{.24\linewidth}
		\includegraphics[width=\linewidth, height=\linewidth]{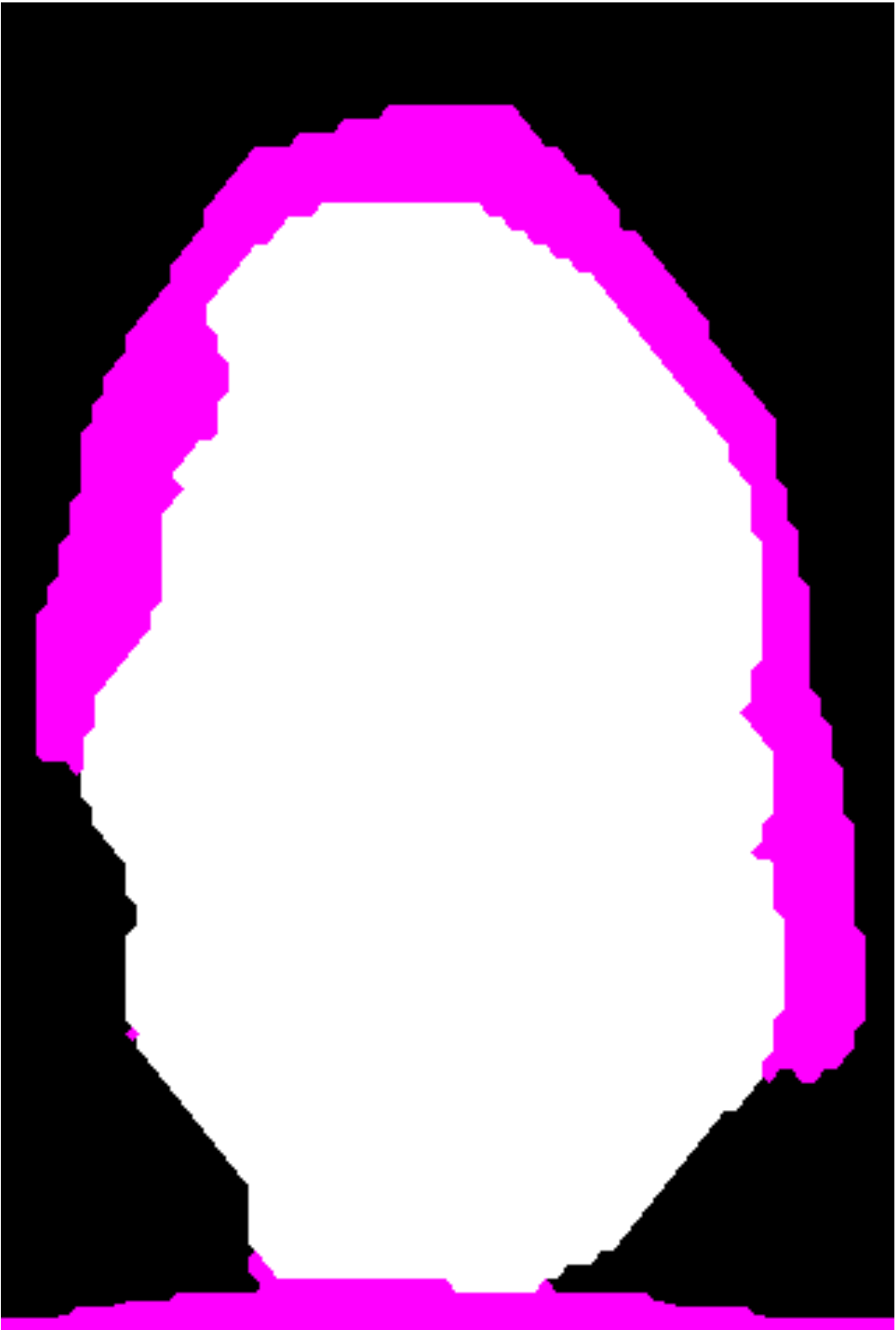}
                 \centerline{(d)}
	\end{minipage} 
	\caption{(a) The fusion of the binary segmentation maps by GC-LAE using different $k$ values. The region outside the LN-mask is in white. The blue, purple, and pink regions are detected as fat in one, two, and three segmentation results, respectively. (b) The LNP seeds are in white and the fat seeds are in pink. (c) The final NGC result based on the results of GC-LAE with global energy (i.e. using two rounds of GC-LAE). (d) The final NGC result based on the results of GC-LAE without global energy  (i.e. using only one round of GC-LAE).}
\label{fig:183_GCLAE}
\end{figure}

\subsection{Refinement Process}

\begin{figure*}[t]
	\centering
	\begin{minipage}[t]{.16\linewidth}
		\includegraphics[width=\linewidth, height=\linewidth]{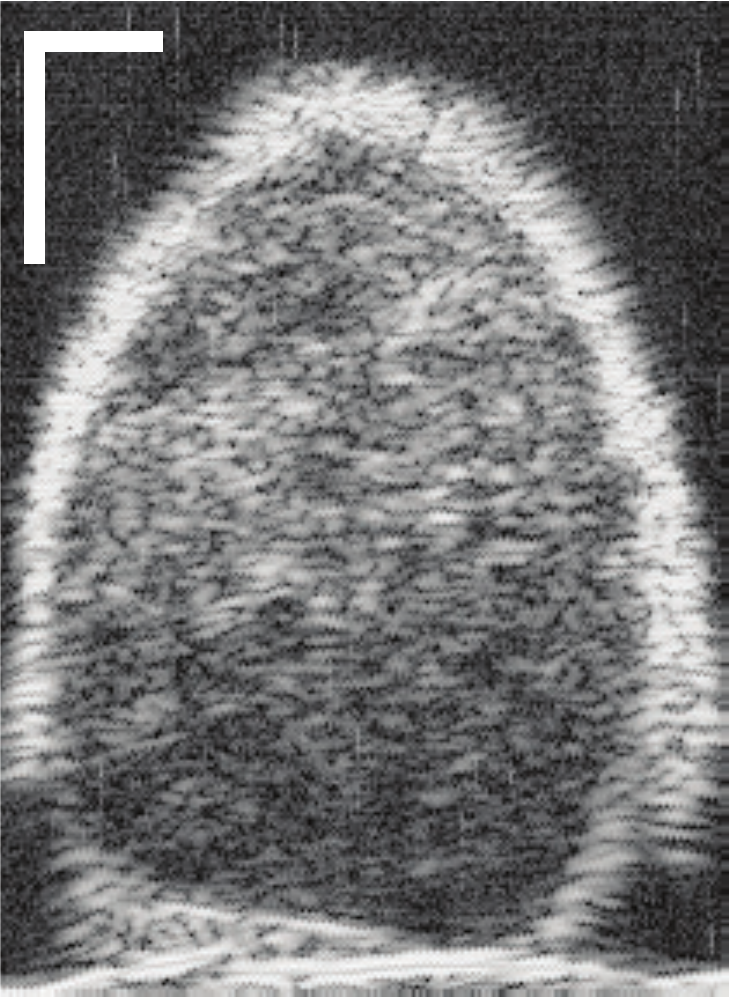}
	\end{minipage} 
	\vspace*{4pt}
	\begin{minipage}[t]{.16\linewidth}
		\includegraphics[width=\linewidth, height=\linewidth]{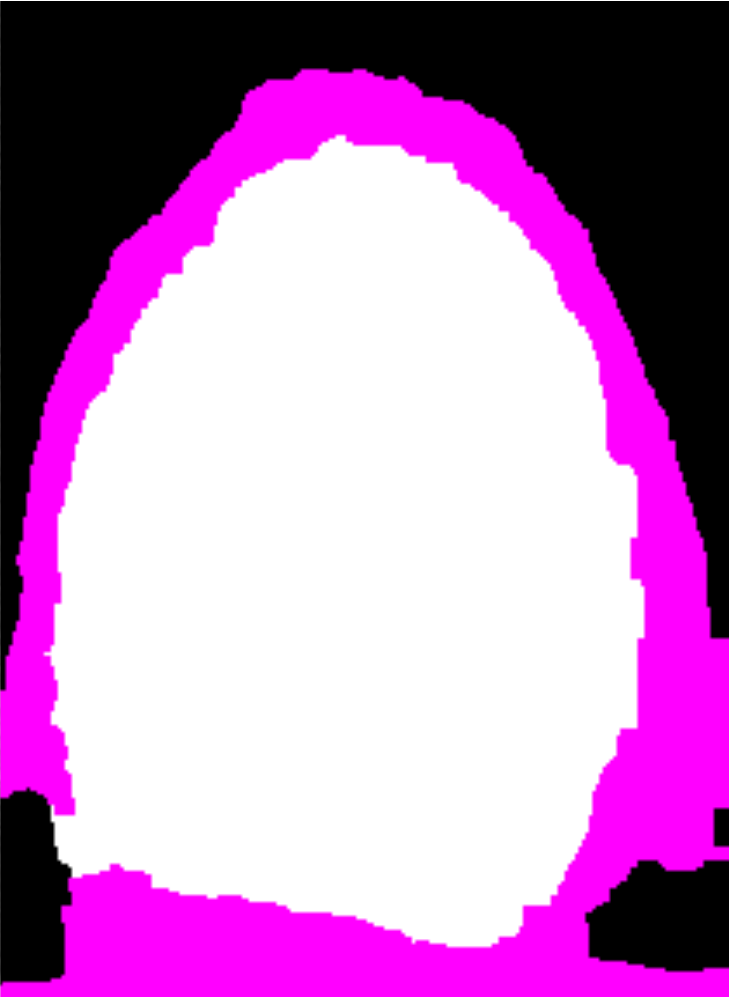}
	\end{minipage} 
	\begin{minipage}[t]{.16\linewidth}
		\includegraphics[width=\linewidth, height=\linewidth]{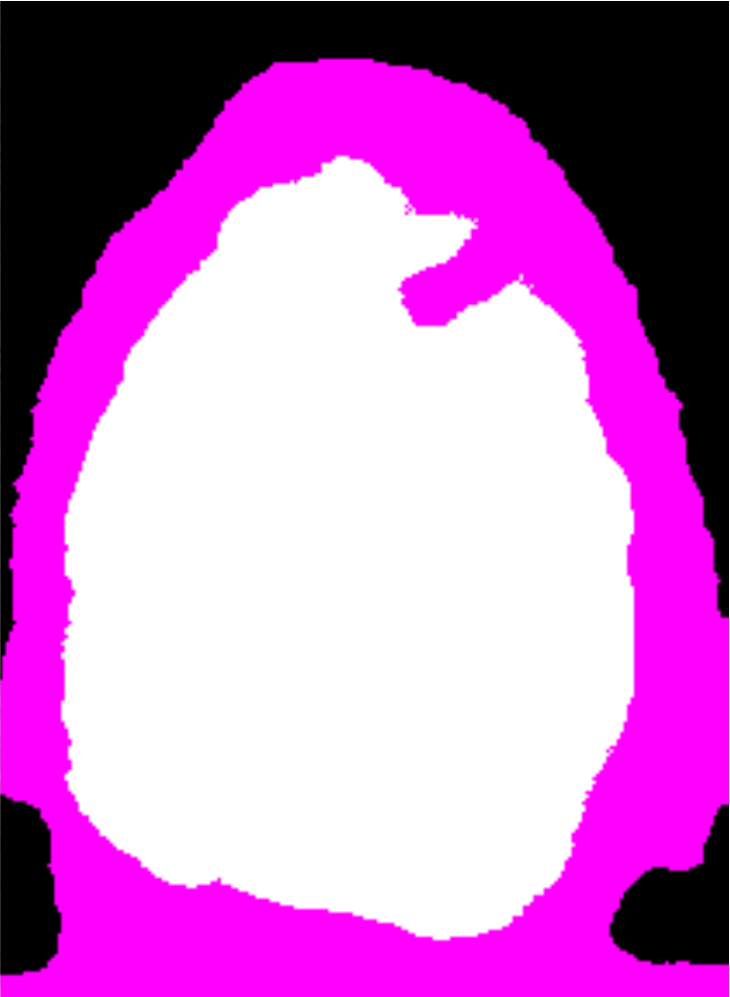}
                 \centerline{DSC = 0.9493}
	\end{minipage} 
	\begin{minipage}[t]{.16\linewidth}
		\includegraphics[width=\linewidth, height=\linewidth]{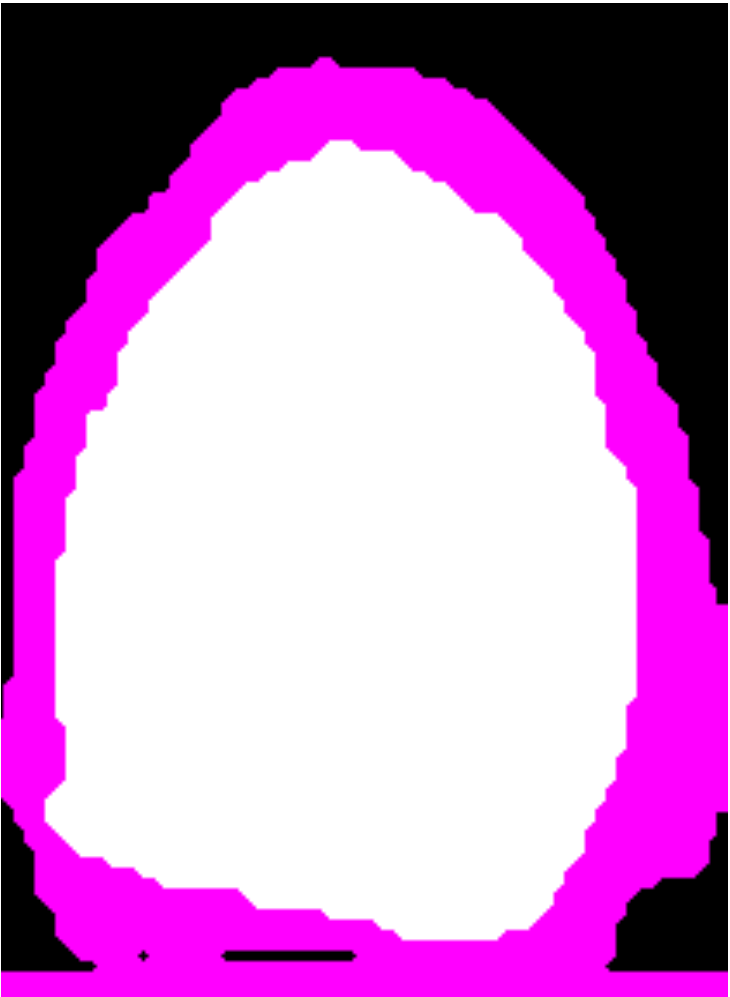}
                 \centerline{DSC = 0.9729}
	\end{minipage} 
	\begin{minipage}[t]{.16\linewidth}
		\includegraphics[width=\linewidth, height=\linewidth]{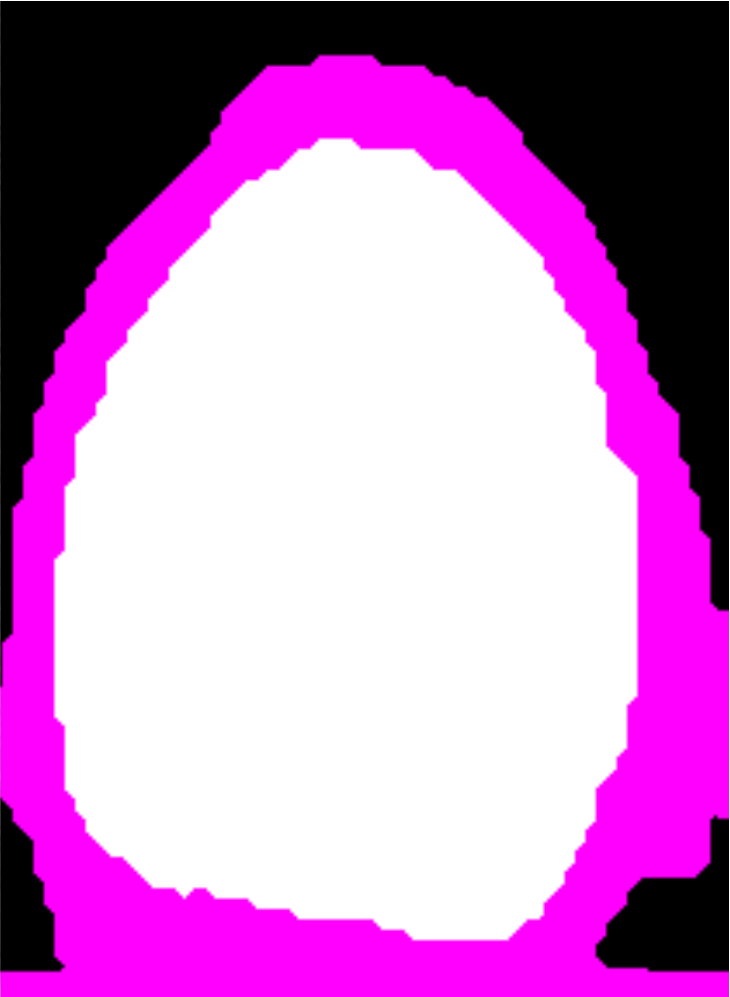}
                 \centerline{DSC = 0.9809}
	\end{minipage} 
	\begin{minipage}[t]{.16\linewidth}
		\includegraphics[width=\linewidth, height=\linewidth]{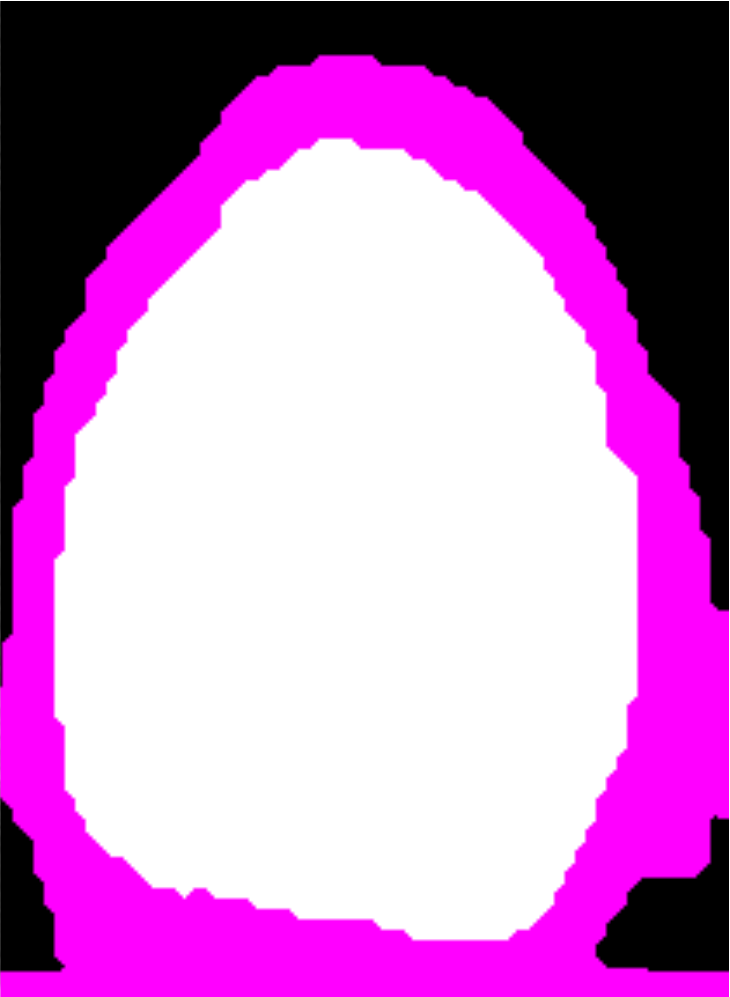}
                 \centerline{DSC = 0.9805}
	\end{minipage} 
	\begin{minipage}[t]{.16\linewidth}
		\includegraphics[width=\linewidth, height=\linewidth]{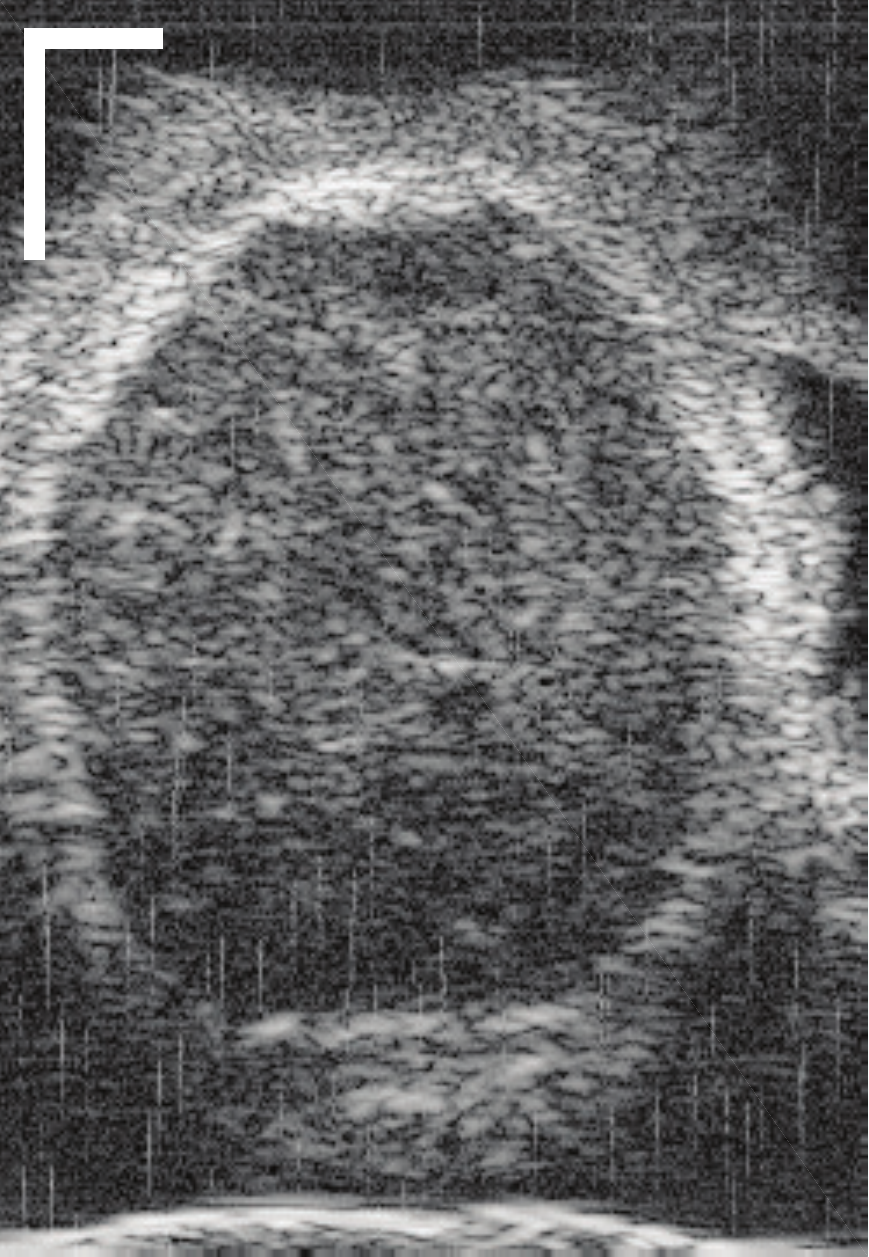}
                 \centerline{(a)}
	\end{minipage} 
	\begin{minipage}[t]{.16\linewidth}
		\includegraphics[width=\linewidth, height=\linewidth]{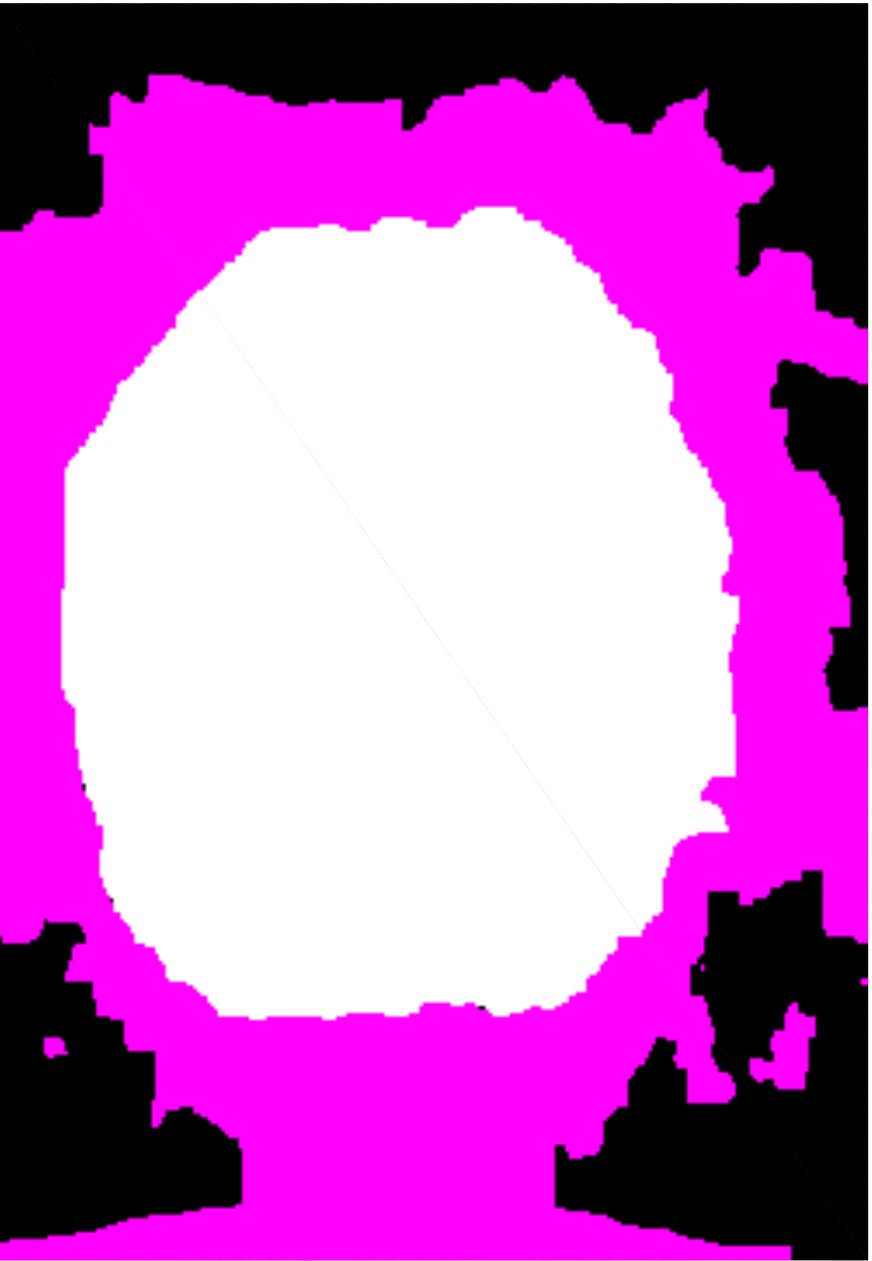}
                 \centerline{(b)}
	\end{minipage} 
	\begin{minipage}[t]{.16\linewidth}
		\includegraphics[width=\linewidth, height=\linewidth]{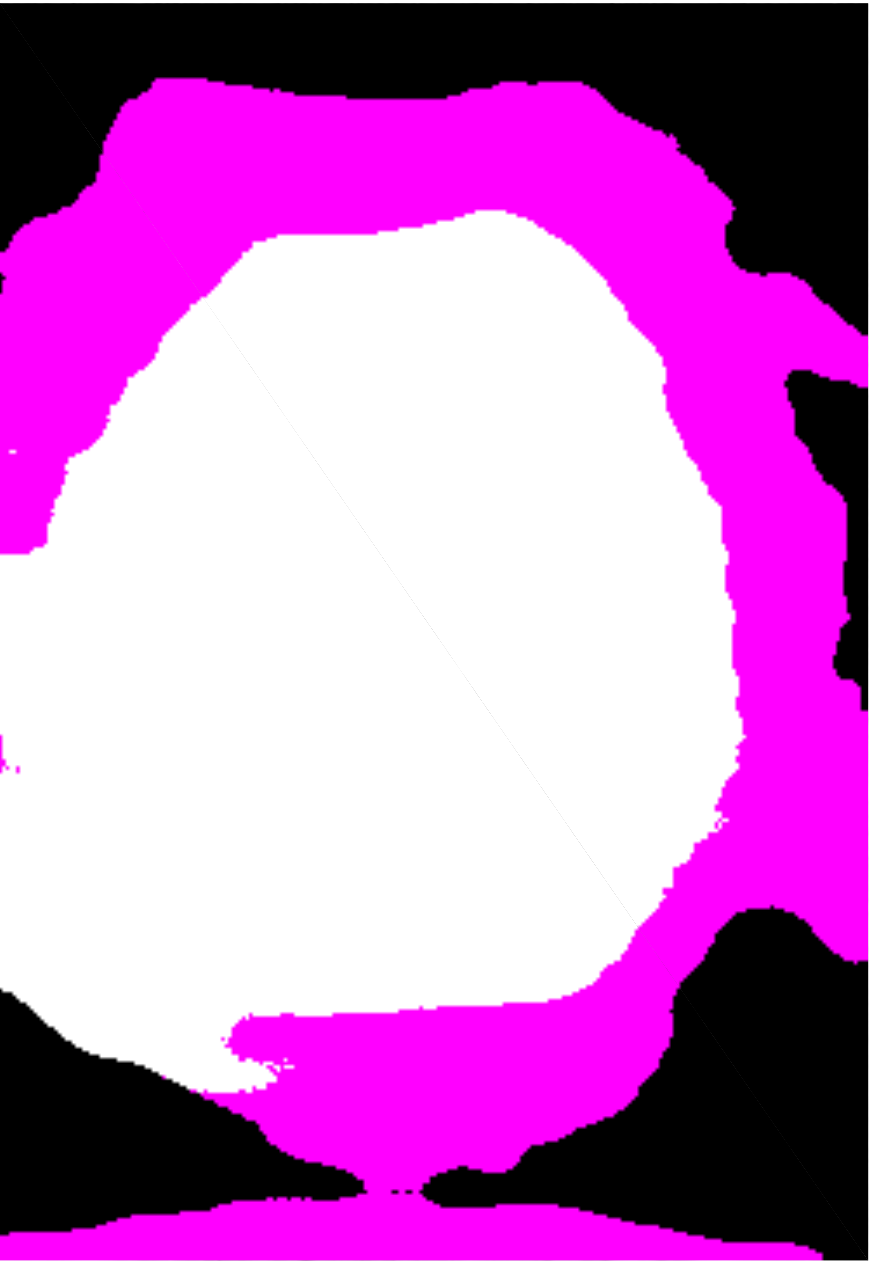}
                 \centerline{(c) DSC=0.9075}
	\end{minipage} 
	\begin{minipage}[t]{.16\linewidth}
		\includegraphics[width=\linewidth, height=\linewidth]{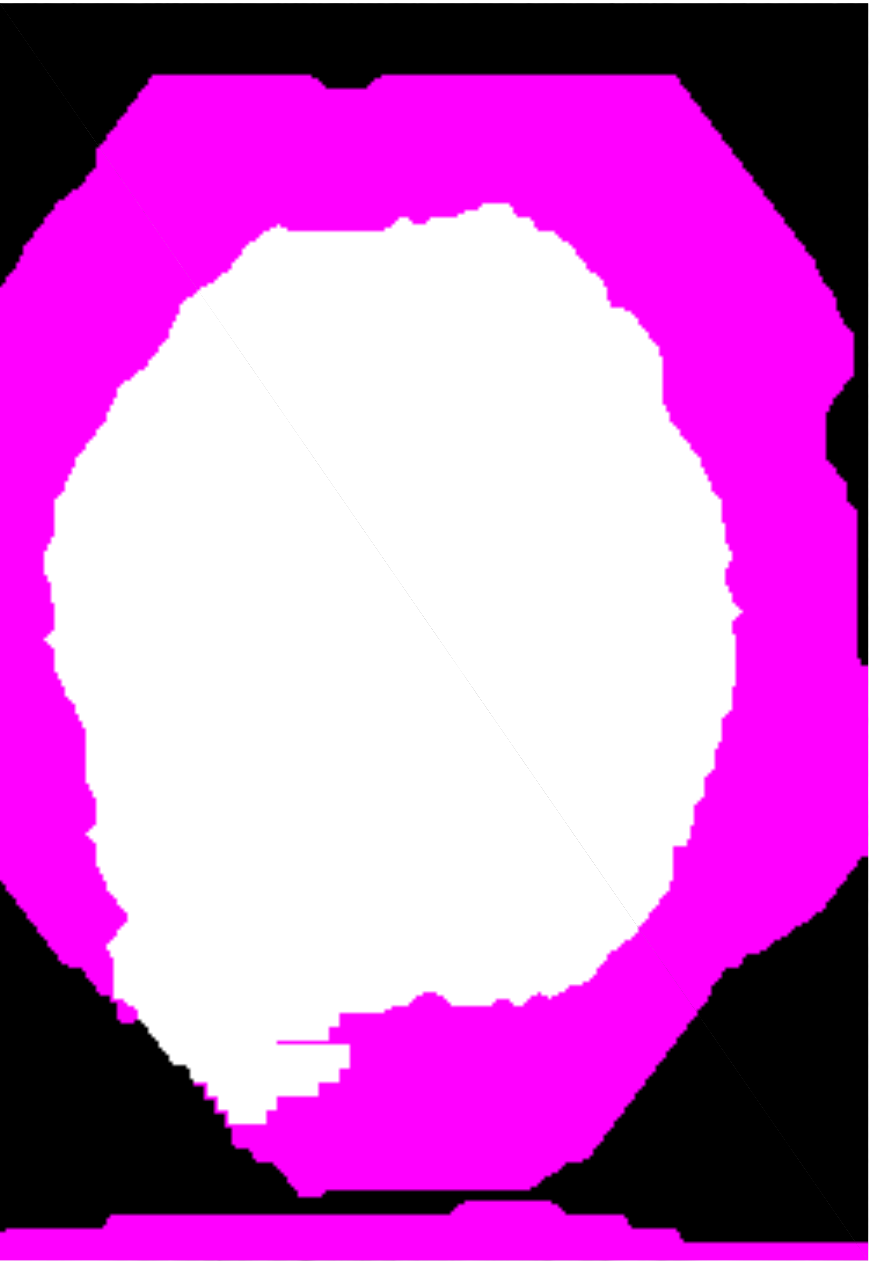}
                 \centerline{(d) DSC=0.9502}
	\end{minipage} 
	\begin{minipage}[t]{.16\linewidth}
		\includegraphics[width=\linewidth, height=\linewidth]{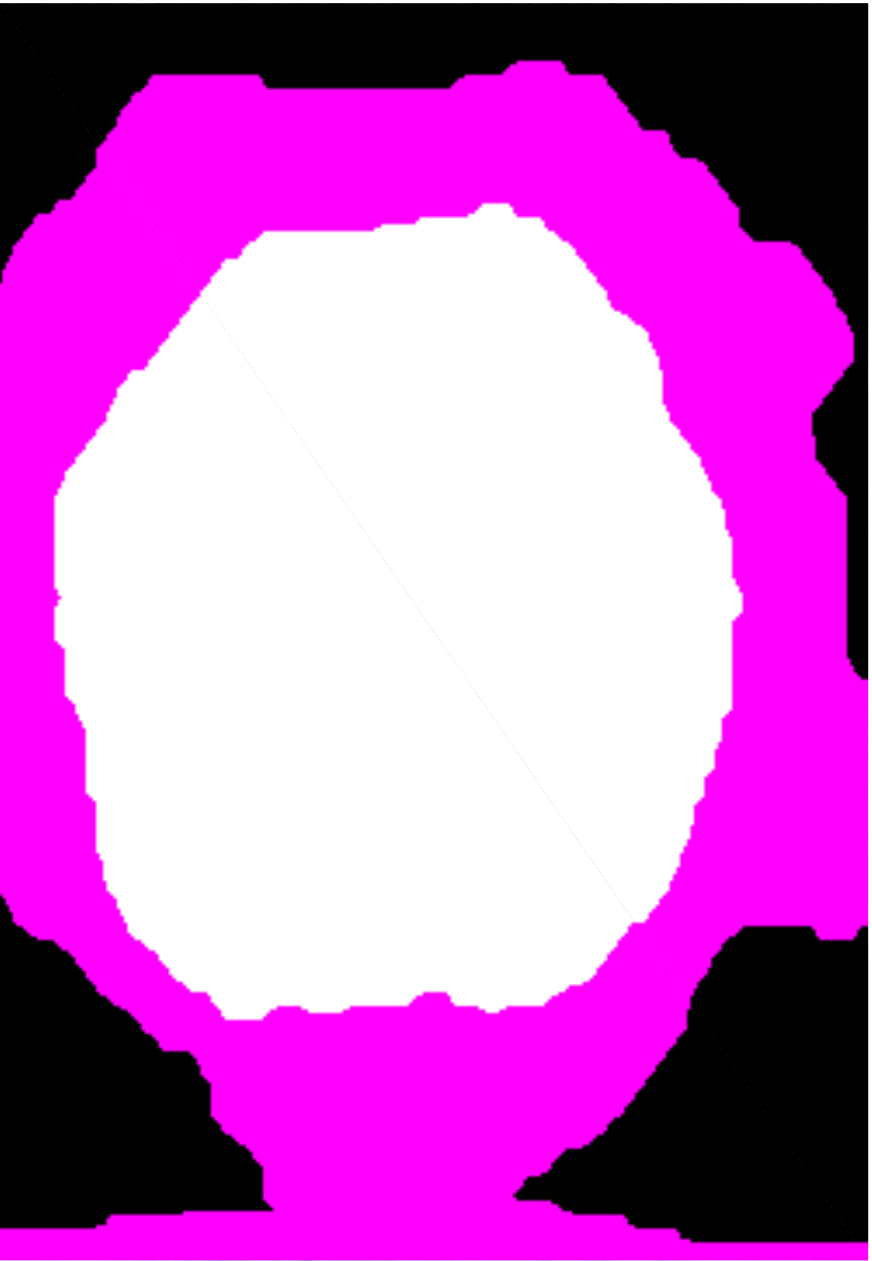}
                 \centerline{(e) DSC=0.9656}
	\end{minipage} 
	\begin{minipage}[t]{.16\linewidth}
		\includegraphics[width=\linewidth, height=\linewidth]{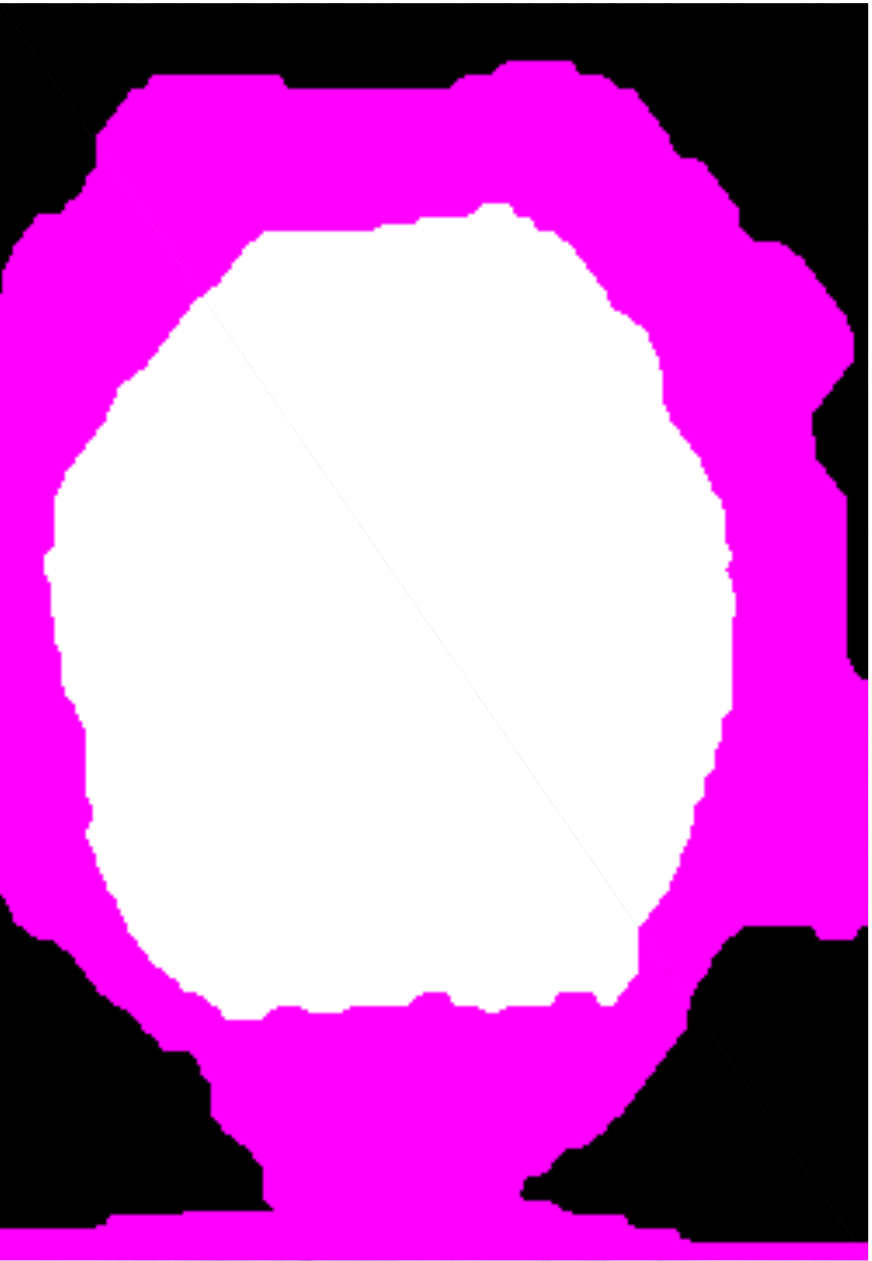}
                 \centerline{(f) DSC=0.9611}
	\end{minipage} 	
		\begin{minipage}[t]{.16\linewidth}
		\includegraphics[width=\linewidth, height=\linewidth]{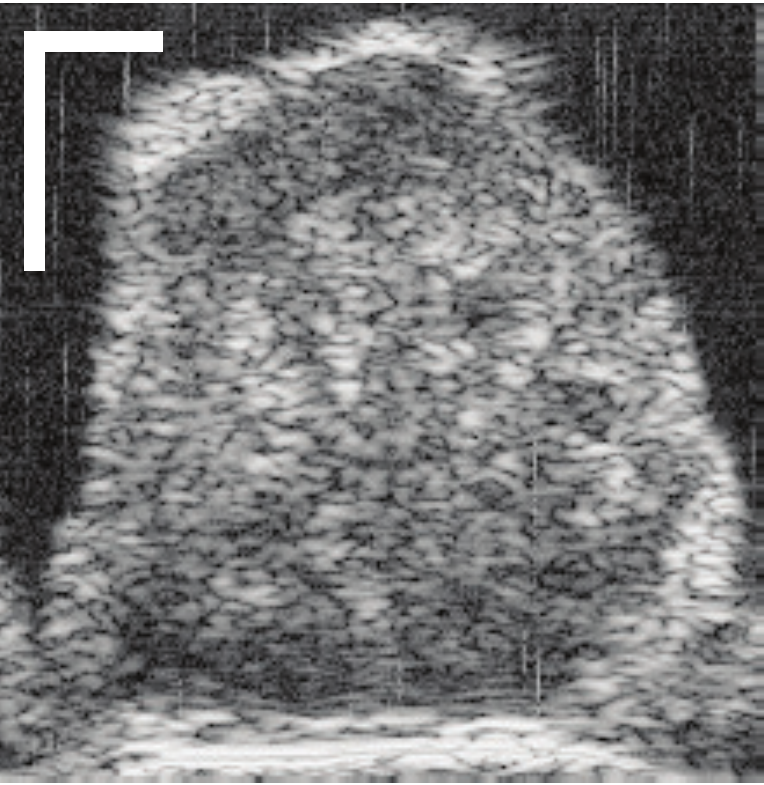}
	\end{minipage} 
	\begin{minipage}[t]{.16\linewidth}
		\includegraphics[width=\linewidth, height=\linewidth]{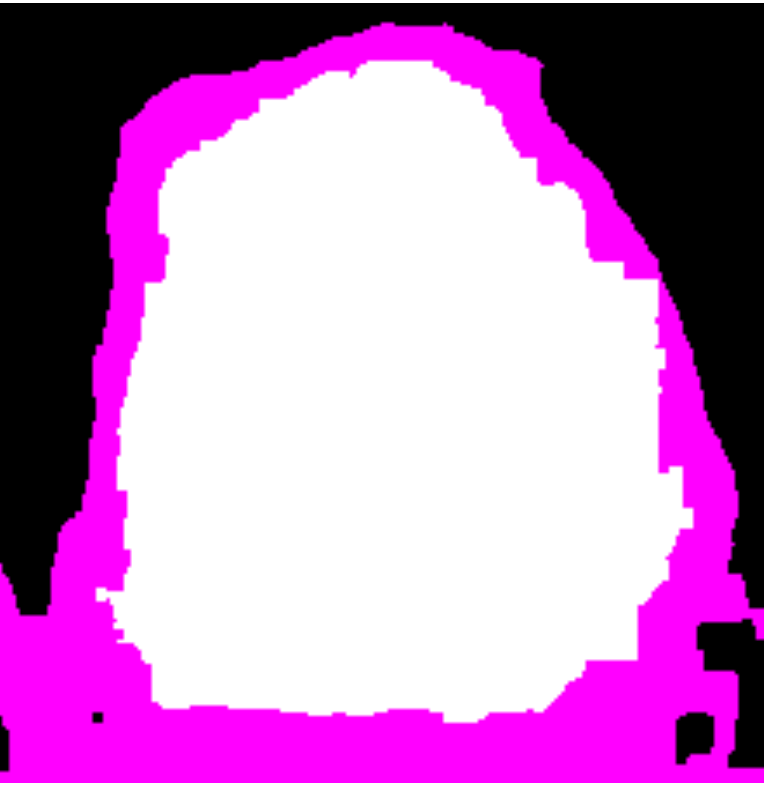}
	\end{minipage} 
	\begin{minipage}[t]{.16\linewidth}
		\includegraphics[width=\linewidth, height=\linewidth]{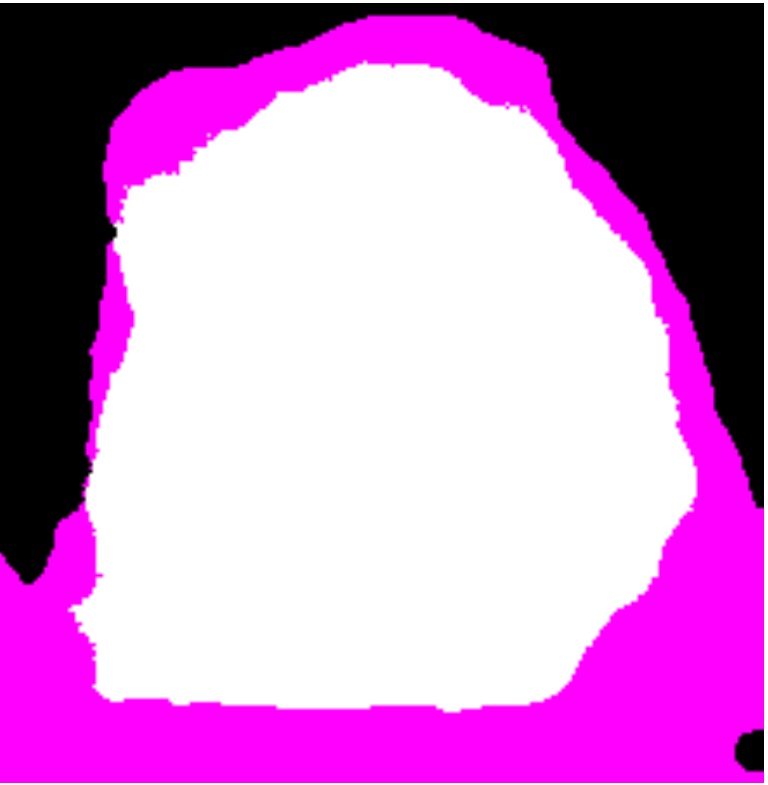}
                 \centerline{DSC = 0.9112}
	\end{minipage} 
	\begin{minipage}[t]{.16\linewidth}
		\includegraphics[width=\linewidth, height=\linewidth]{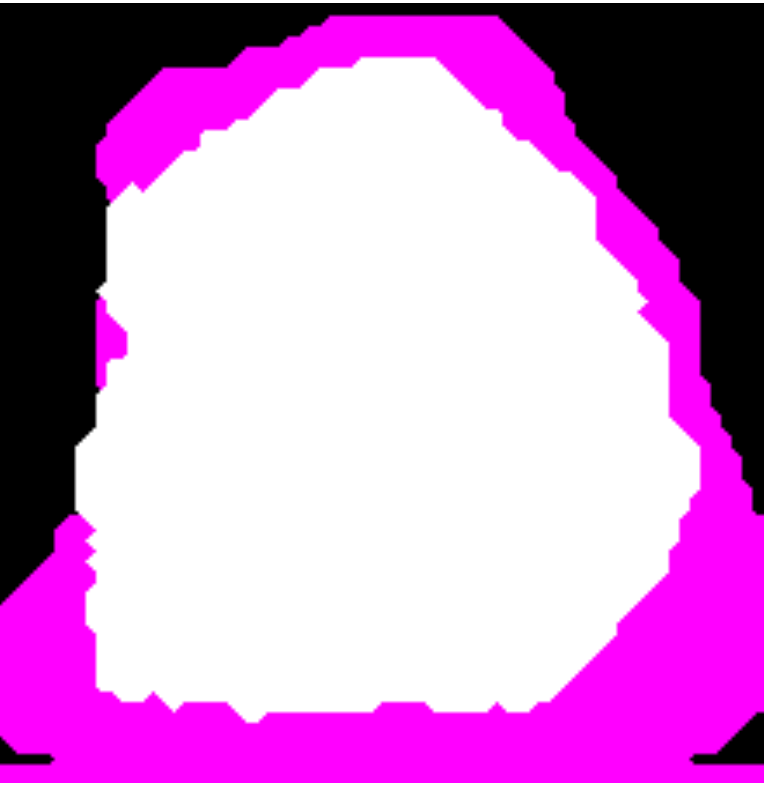}
                 \centerline{DSC = 0.8971}
	\end{minipage} 
	\begin{minipage}[t]{.16\linewidth}
		\includegraphics[width=\linewidth, height=\linewidth]{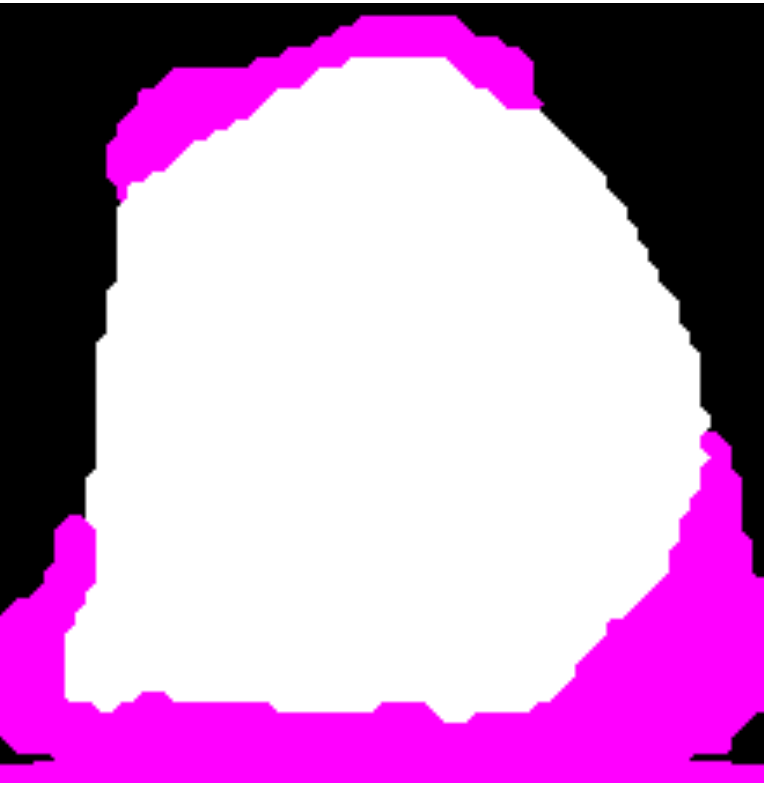}
                 \centerline{DSC = 0.8852}
	\end{minipage} 
	\begin{minipage}[t]{.16\linewidth}
		\includegraphics[width=\linewidth, height=\linewidth]{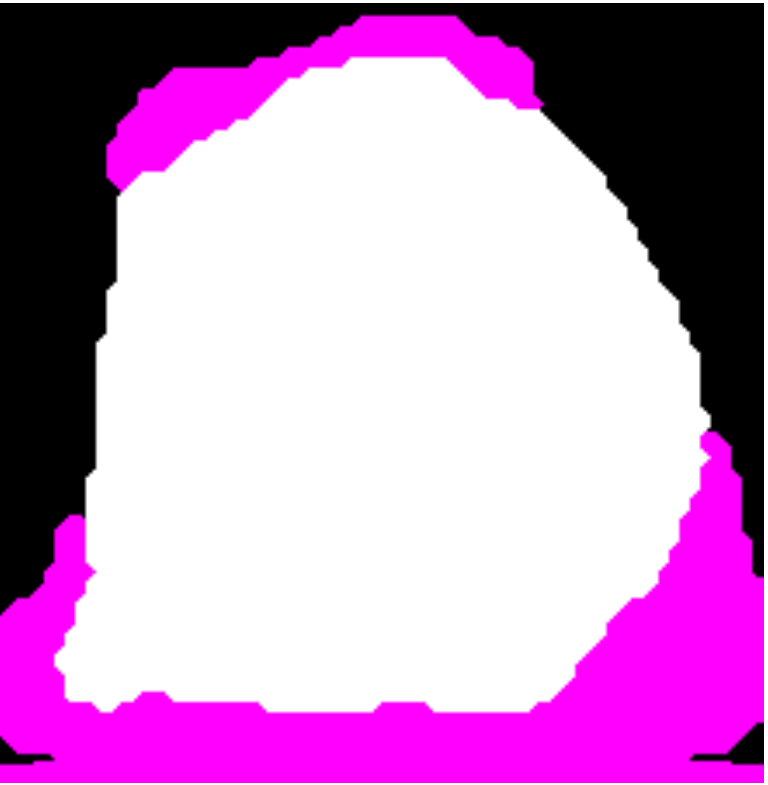}
                 \centerline{DSC = 0.8858}
	\end{minipage} 
	\begin{minipage}[t]{.16\linewidth}
		\includegraphics[width=\linewidth, height=\linewidth]{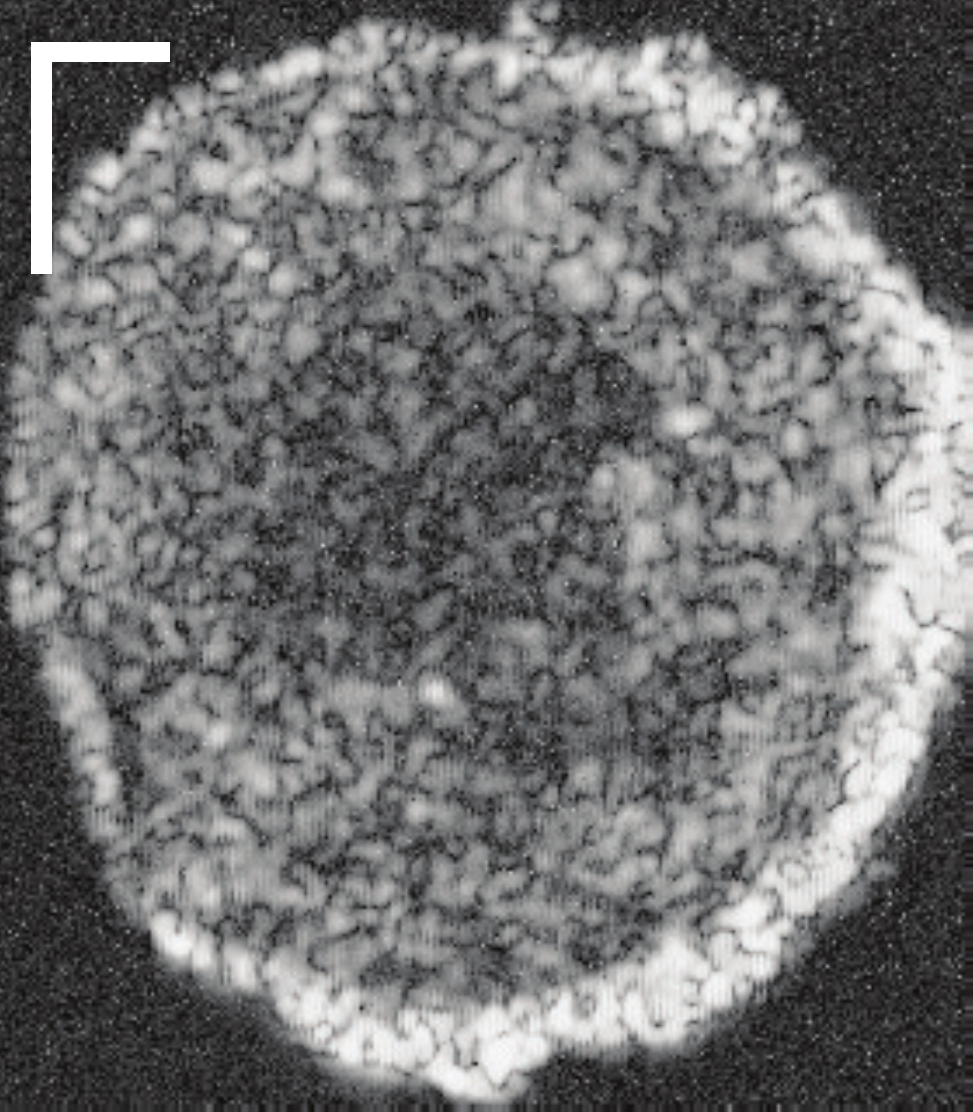}
                 \centerline{(a)}
	\end{minipage} 
	\begin{minipage}[t]{.16\linewidth}
		\includegraphics[width=\linewidth, height=\linewidth]{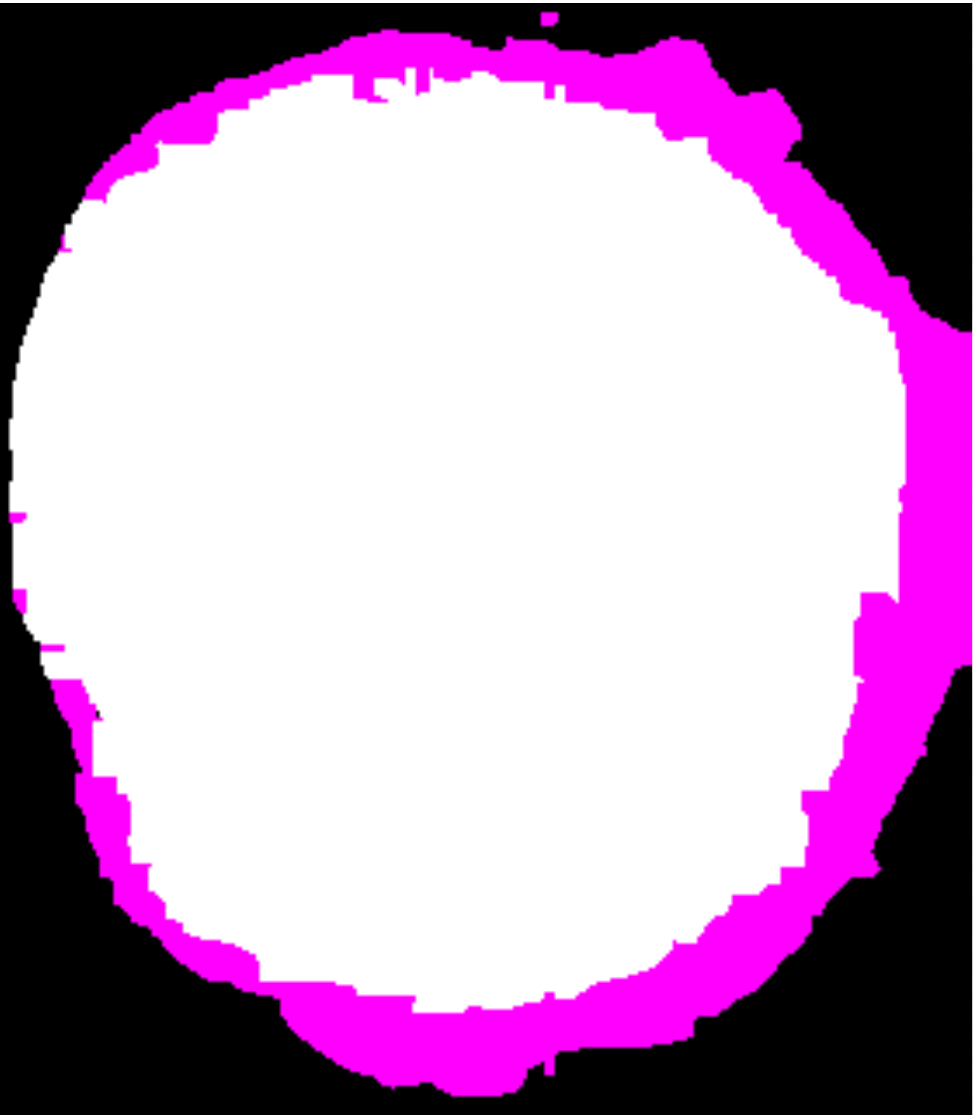}
                 \centerline{(b)}
	\end{minipage} 
	\begin{minipage}[t]{.16\linewidth}
		\includegraphics[width=\linewidth, height=\linewidth]{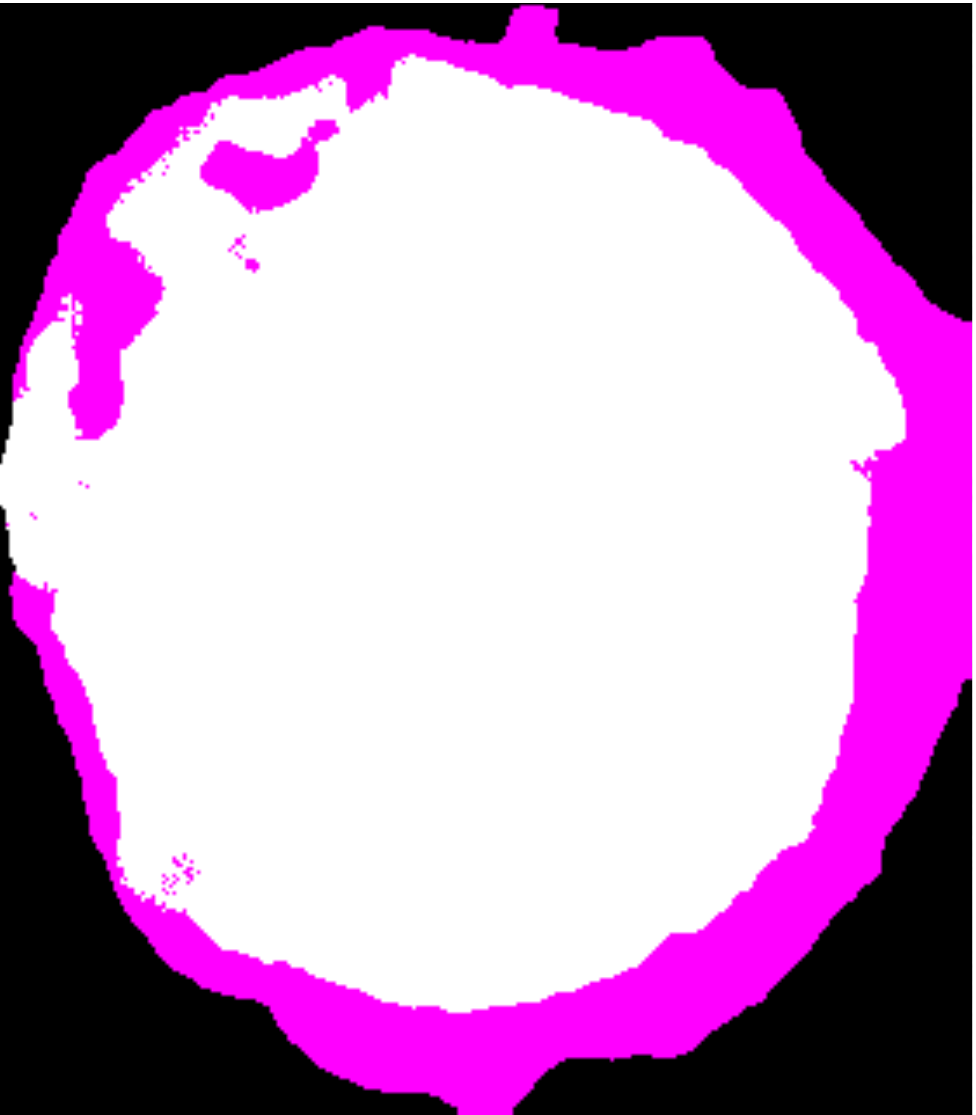}
                 \centerline{(c) DSC=0.9381}
	\end{minipage} 
	\begin{minipage}[t]{.16\linewidth}
		\includegraphics[width=\linewidth, height=\linewidth]{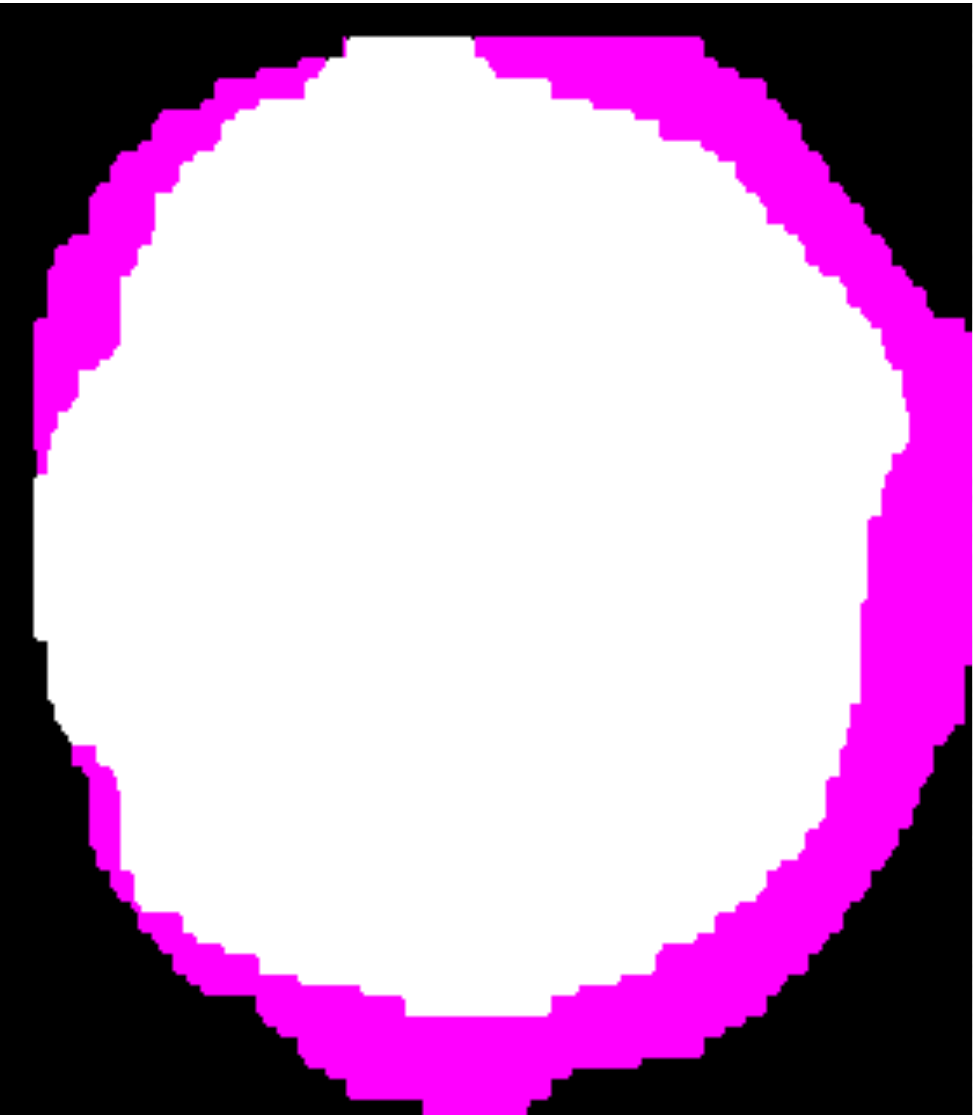}
                 \centerline{(d) DSC=0.9594}
	\end{minipage} 
	\begin{minipage}[t]{.16\linewidth}
		\includegraphics[width=\linewidth, height=\linewidth]{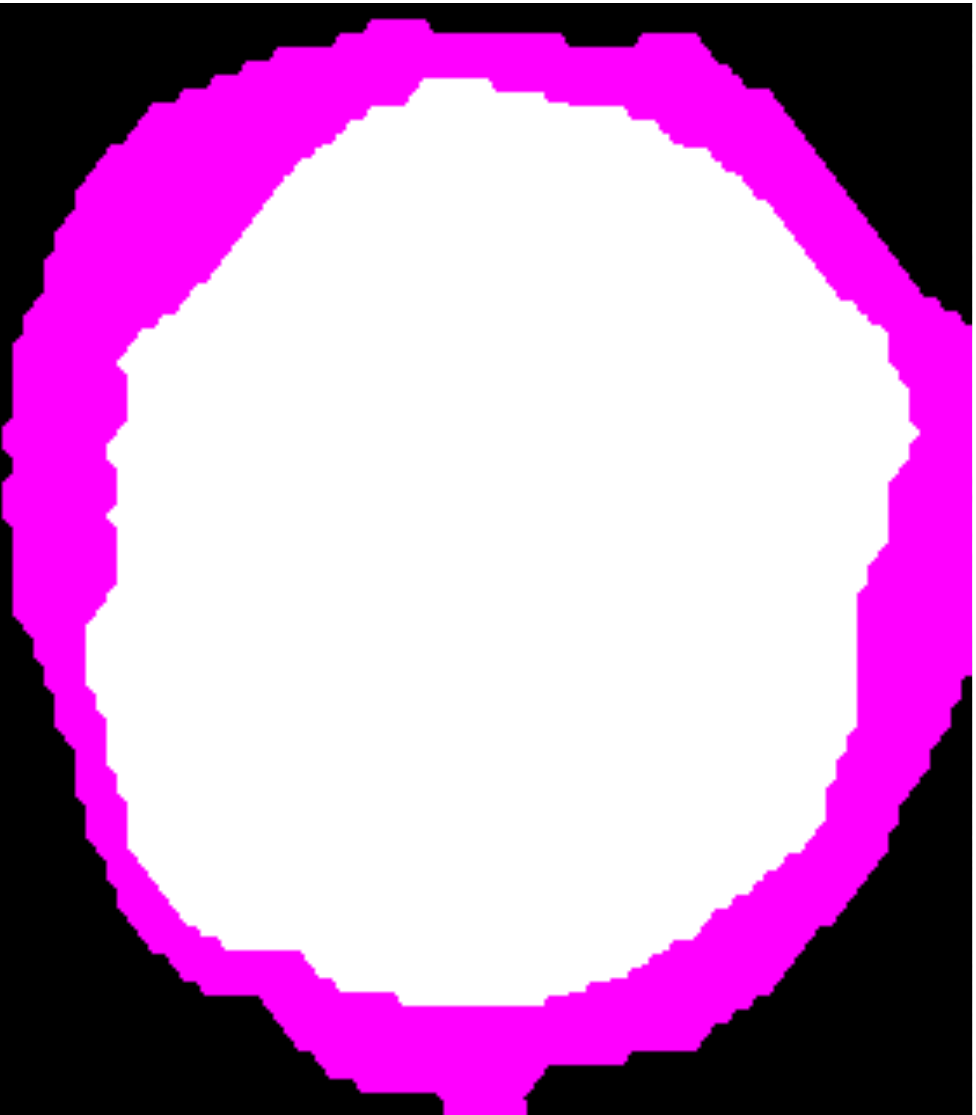}
                 \centerline{(e) DSC=0.9534}
	\end{minipage} 
	\begin{minipage}[t]{.16\linewidth}
		\includegraphics[width=\linewidth, height=\linewidth]{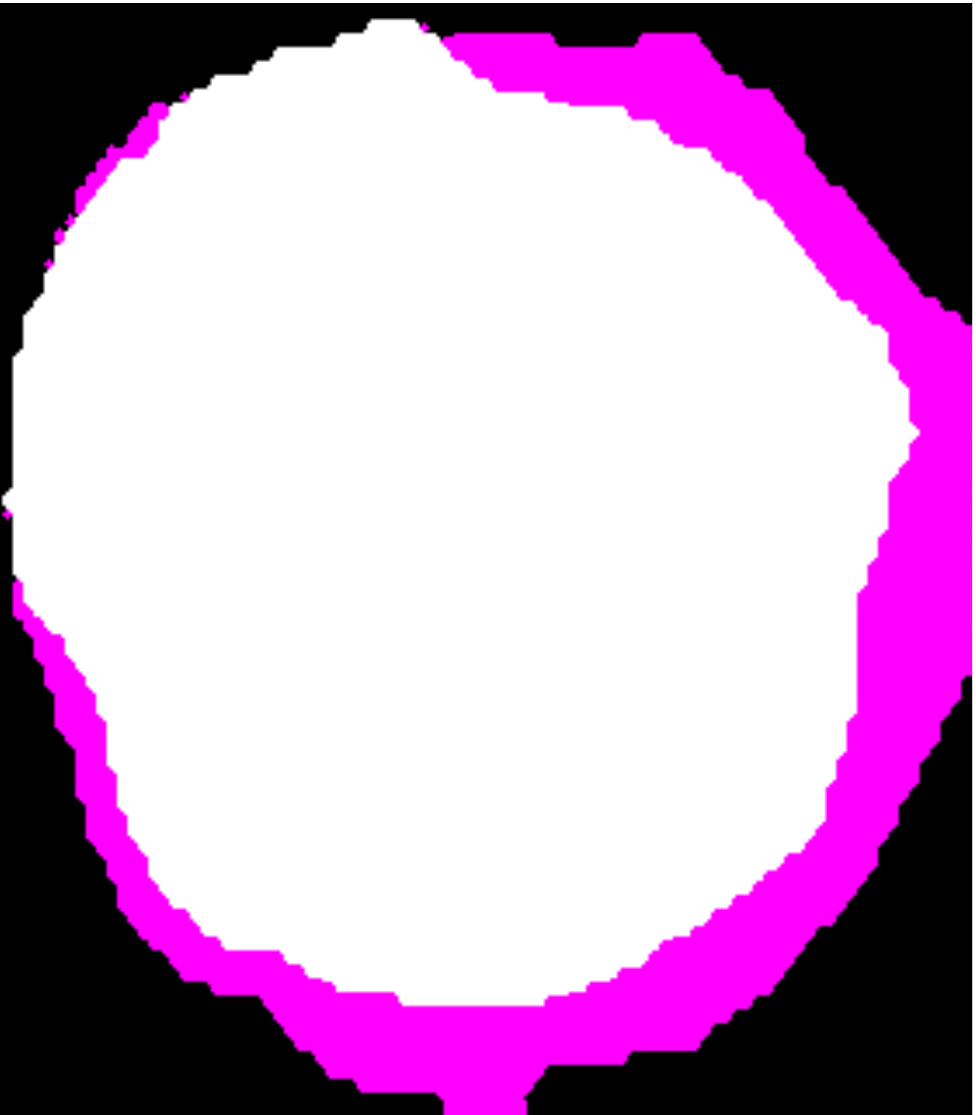}
                 \centerline{(f) DSC=0.9677}
	\end{minipage} 
	\caption{(a) Target HFU image. (b) Manual segmentation result. (c) The result of RFC+STS-LS. (d) The result of NGC with depth-dependent profiles. (e) The result of proposed method without global energy (one round of GC-LAE). (f) The result of proposed method with global energy (two rounds of GC-LAE).}
\label{fig:result_all}
\end{figure*}

NGC is used to refine the result of GC-LAE to guarantee the nested relationship among PBS, fat and LNP, and obtain the final segmentation result. As described in Sec. \ref{sec:update PBS}, we obtain depth-dependent PBS parameters $I^T_p$ and $\sigma_p$ based on the GC-LAE segmentation result. We also generate seeds for the fat and LNP regions based on the previous segmentation result. NGC can correct the errors of GC-LAE such as the mislabeled ``halo'' voxels outside the fat shown in Figure \ref{fig:result_flow}(d) (indicated by the green arrow in Figure \ref{fig:result_flow}(e)). The final segmentation result of NGC is shown in Figure \ref{fig:result_flow}(f). 

Initial seeds for the LNP and fat regions are obtained from the confident regions obtained by the second round GC-LAE (the bottom row of Figure \ref{fig:compare_GCLAE}). As shown in Figure \ref{fig:183_GCLAE}(a), we first obtained a fused segmentation map from the GC-LAE results. The voxel values $T_p$ in black, blue, purple, and pink regions are 0, 1, 2, and 3, which are the number of times that each voxel is considered to be fat by the three segmentation results. The pink region is considered the confident fat region, and the black region is the confident LNP region. As can be seen, there are some ``halo" pixels outside the pink region that are falsely considered as LNP. We apply a morphological opening to remove the halo region of confident LNP region and define the remaining region as the seeds of LNP. We also set confident fat region as seeds for the fat region. Furthermore, we remove any fat (pink) regions that are completely contained by LNP seeds through connectivity analysis, so that the fat seeds only lie on the outer boundary of the LN-mask. As shown in Figure \ref{fig:183_GCLAE}(b), pink pixels are the final seeds for fat, and the white pixels are the seeds for LNP. 

To make NGC more likely to cut along the boundary between LNP and fat regions obtained by GC-LAE, we define the edge cost between voxels by the sum of the three binary segmentation results obtained using GC-LAE with different values of $k$ as shown in Figure \ref{fig:183_GCLAE}(a). Making use of the fact that the fat region is outside the LNP and should have intensity higher than the LNP , we define the edge cost, $V_{p,q}$, as:

\begin{equation*}
	V_{p,q} =
  	\begin{dcases}
    		1  & T_p>T_q \\
    		4 & T_p=T_q \\
		\infty & T_p<T_q
		\end{dcases} 
\end{equation*}
Note that $T_p$ is set as $T_q+1$ when $p$ is outside the LN-mask and $q$ is inside the LN-mask. By this setting, NGC will tend to cut the LNP-fat boundary along the boundary of GC-LAE segmentations. 

For fat-PBS boundary, we do not change the edge cost, so NGC cuts the fat-PBS boundary by the intensity and the new depth-dependent PBS threshold obtained in Sec. \ref{sec:update PBS}. Figure \ref{fig:183_GCLAE}(c) and (d) compares the result of NGC based on the segmentation results of GC-LAE with and without global energy. The result based on GC-LAE with global energy obtained more accurate LNP-fat boundary on the left.

\section{Results}
\label{sec:result}

In this study, we compared the result of the current, described method with two other methods: our own prior work, called NGC with depth-dependent profile \cite{kuo2015novel} and the method developed by Bui et al, called statistical transverse slice level-set (STS-LS) with random forest classification (RFC) \cite{bui2015random}.

The NGC method is based on the EM framework \cite{kuo2015novel}. In the expectation phase, NGC is applied to segment LNP, fat and PBS simultaneously. In the maximization phase, the mean and STD of the intensity in each depth of each object is obtained by the previous segmentation result of NGC. A spline-based fitting method is applied to fit the mean intensity (resp. STD) of each object as a function of depth to establish the depth-dependent profiles of each objects. These depth-dependent profiles are used in the NGC during the expectation phase of next iteration. The iteration stops when the segmentation result of NGC is converged. 

The STS-LS with RFC algorithm contains an initialization step and a refinement step \cite{bui2015random}. In the initialization step, 15 features are extracted for each voxel, and RFC is employed to obtain an initial segmentation result. In the refinement step, the initial result is refined by a 3-phase, local-region-based, level-set segmentation method. The Gamma distribution is used to model the envelope data of HFU images. During the contour evolution, Gamma distribution parameters are estimated locally in transverse slices (i.e., depth-dependent).

\subsection{Illustrative Results}

Figure \ref{fig:result_all} displays the segmentation results of four typical LNs to illustrate the advantages and disadvantages of the three different segmentation methods. RTS-LS with RFC and NGC with depth-dependent profiles segment LNP mainly by intensity information, and GC-LAE-based methods segment LNP mainly by the intensity similarity of local regions.

\subsubsection{LNs with Clear Boundary and Consistent Intensity Distributions}

The top LN has clear boundary and consistent intensity distributions. Therefore, all four segmentation methods can obtain satisfactory results. However, RFC+STS-LS may converge to a false local minimum during deformation as shown in the top figure of Figure \ref{fig:result_all}(c). In contrast, the GC-based approach can avoid this kind of error. 

\subsubsection{LNs with Clear Boundary and Non-homogeneous Acoustic Attenuation}

As discussed in introduction, depth-dependent profiles may not perform adequately in the presence of inconsistent profiles, where the intensity of voxels within the same depth in one object may not be consistent. To illustrate this point, the second row of Figures \ref{fig:result_all}(c) and \ref{fig:result_all}(d) show the segmentation result of RFC+STS-LS and NGC with depth-dependent profiles, respectively. Comparing these results to the manual-segmentation result in Figure \ref{fig:result_all}(b), we can identify a fat region on the left mislabeled as LNP because the fat in the right region of the image at the same depth is brighter than the fat in left. In this case, NGC with depth-dependent profiles and RFC+STS-LS were unable to correctly segment the fatty region on the left. Because the contrast between LNP and fat is strong enough on the left side of the LN, GC-LAE was able to find the boundary correctly (Figure \ref{fig:result_all}(e) and \ref{fig:result_all}(f)) independent of whether the global-energy term is used.

\subsubsection{LNs with Blurry Boundary and Low Intensity Difference between LNP and the Fat}

GC-LAE determines a boundary based on the similarity of local intensity distributions. Therefore, it may not be able to find the boundary correctly when the boundary is blurry. In the third LN of Figure \ref{fig:result_all}, the boundary between LNP and the fat in middle is extremely blurry and the intensity difference between them is low. In this case, GC-LAE cannot find the boundary correctly even with global energy because of insufficient contrast, but RFC+STS-LS and NGC with depth-dependent profiles were able to provide better segmentation results according to intensity distributions. RFC+STS-LS obtain the best results in this case by updating intensity distributions gradually with deformation.

\subsubsection{LNs with Blurry Boundary and High Intensity Difference between LNP and the Fat}

The bottom row in Figure \ref{fig:result_all} shows a transverse slice of an LN with a ``blurry'' boundary at the upper left corner, and the intensity of the fat on right is much brighter than LNP. In this case, RFC+STS-LS remained at a false local minimum during the segmentation of this LN. NGC with depth-dependent profiles performed better, but it was adversely influenced by the blurry boundary. GC-LAE without global information also incorrectly segments some bright voxels of LNP as fat. In this example, GC-LAE with a global-energy term has the best performance. Using the additional information provided by the depth-dependent profiles, the resulting boundary nearly matches the true boundary.

\subsection{Results in Our Database}
\label{sec:result in our database}

To evaluate the segmentation result, we use Dice similarity coefficient (DSC), which is defined as:

\[
	DSC=2|X\cap Y|/(|X|+|Y|),
\]
where $X$ and $Y$ are the results of the segmentation algorithm and the ground truth (i.e., manual expert segmentation), respectively. DSC evaluates the similarity between two segmentation results $X$ and $Y$.

Table \ref{tab:42} shows the segmentation results for the 42 Colorrectal LNs used in \cite{bui2015random}, which is a subset of the images in the complete dataset. We compare the performance of the RFC+STS-LS, NGC with depth-dependent profiles, and the current, described GC-LAE-based method with and without depth-dependent profiles. Table \ref{tab:42} demonstrates that the GC-LAE outperforms two other methods by obtaining highest average DSC in all three objects.

Table \ref{tab:115} shows the DSC values for 115 LNs of our whole database, which includes the 42 LNs used in Table \ref{tab:42}. It compares the performance of NGC with depth-dependent profiles, the described method without global energy (one round of GC-LAE), and the proposed method with global energy (two rounds of GC-LAE). As shown in Table \ref{tab:115}, GC-LAE obtained higher average DSC than NGC with depth-dependent profile method in all three objects. With the information of depth-dependent profiles, GC-LAE can obtain more accurate results in LNP and fat segmentation and lead to better accuracy. In general, the total pixel number of the fat is less than the total number of LNP region. Therefore, with the same number of errors, the DSC of the fat will be lower than that of the LNP. 

\bgroup
\def\arraystretch{1.2}
\begin{table}[t]
\centering
\caption{The mean and standard deviation of DSC for the segmented LNP regions by different methods in 42 Colorrectal LNs}
\resizebox{0.99\linewidth}{!}{
\begin{tabular}{>{\centering}m{0.15\linewidth}>{\centering}m{0.25\linewidth}>{\centering\arraybackslash}m{0.25\linewidth}>{\centering\arraybackslash}m{0.25\linewidth}}
	\hline \hline
	 & RFC+STS-LS & NGC with Depth-dependent profile & GC-LAE with global energy \\ 
	\hline
	LNP & 0.937$\pm$0.020 & 0.939$\pm$0.025 & 0.949$\pm$0.023 \\
	Fat & 0.828$\pm$0.094 & 0.813$\pm$0.082 & 0.837$\pm$0.076 \\
	PBS & 0.961$\pm$0.009 & 0.957$\pm$0.012 & 0.961$\pm$0.013 \\
	\hline \hline
	\end{tabular}
	}
\label{tab:42}
\end{table}
\egroup

\bgroup
\def\arraystretch{1.2}
\begin{table}[t]
\centering
\caption{The mean and standard deviation of DSC for the segmented LNP regions by different methods in 115 LNs}
\resizebox{0.99\linewidth}{!}{
\begin{tabular}{>{\centering}m{0.15\linewidth}>{\centering}m{0.25\linewidth}>{\centering\arraybackslash}m{0.25\linewidth}>{\centering\arraybackslash}m{0.25\linewidth}}
	\hline \hline
	 & NGC with Depth-dependent profile & GC-LAE without global energy & GC-LAE with global energy \\ 
	\hline
	LNP & 0.929$\pm$0.037 & 0.934$\pm$0.041 & 0.937$\pm$0.035 \\
	Fat & 0.793$\pm$0.098 & 0.818$\pm$0.090 & 0.822$\pm$0.085 \\
	PBS & 0.956$\pm$0.016 & 0.959$\pm$0.016 & 0.959$\pm$0.016 \\
	\hline \hline
	\end{tabular}
	}
\label{tab:115}
\end{table}
\egroup

\bgroup
\def\arraystretch{1.2}
\begin{table}[t]
\centering
\caption{The average execution time of the proposed method in minutes.}
\resizebox{0.99\linewidth}{!}{
\begin{tabular}{>{\centering}m{0.15\linewidth}>{\centering}m{0.25\linewidth}>{\centering\arraybackslash}m{0.25\linewidth}>{\centering\arraybackslash}m{0.25\linewidth}}
	\hline \hline
	 & STS-LS & NGC with Depth-dependent profile & GC-LAE \\ 
	\hline
	54 LNs  &19.4 $\pm$ 17.4 &  2.4$\pm$1.6 & 4.4$\pm$1.8 \\
	115 LNs &  & & 4.4$\pm$3.0 \\
	\hline \hline
	\end{tabular}
	}
\label{tab:time}
\end{table}
\egroup

\subsection{Computation Time}

In NGC with depth-dependent profiles and GC-LAE, we down sample all testing HFU images by an integer factor between 2 or 4 to yield an image whose dimensions are always smaller than 100 by 100 by 100 to reduce the execution time. Table \ref{tab:time} lists the average execution time in minutes per LN over 54 Colorrectal LNs used in \cite{bui2016local} and 115 LNs including breast LNs. (These times were obtained using a PC with a 2.6-GHz i7 processor using MatLab 2016b without use of GPUs or any parallel computing.)


\section{Discussion and Conclusions}
\label{sec:discussion and conclusion}

\begin{figure}[t]
	\centering
	\begin{minipage}[t]{.90\linewidth}
		\includegraphics[width=\linewidth]{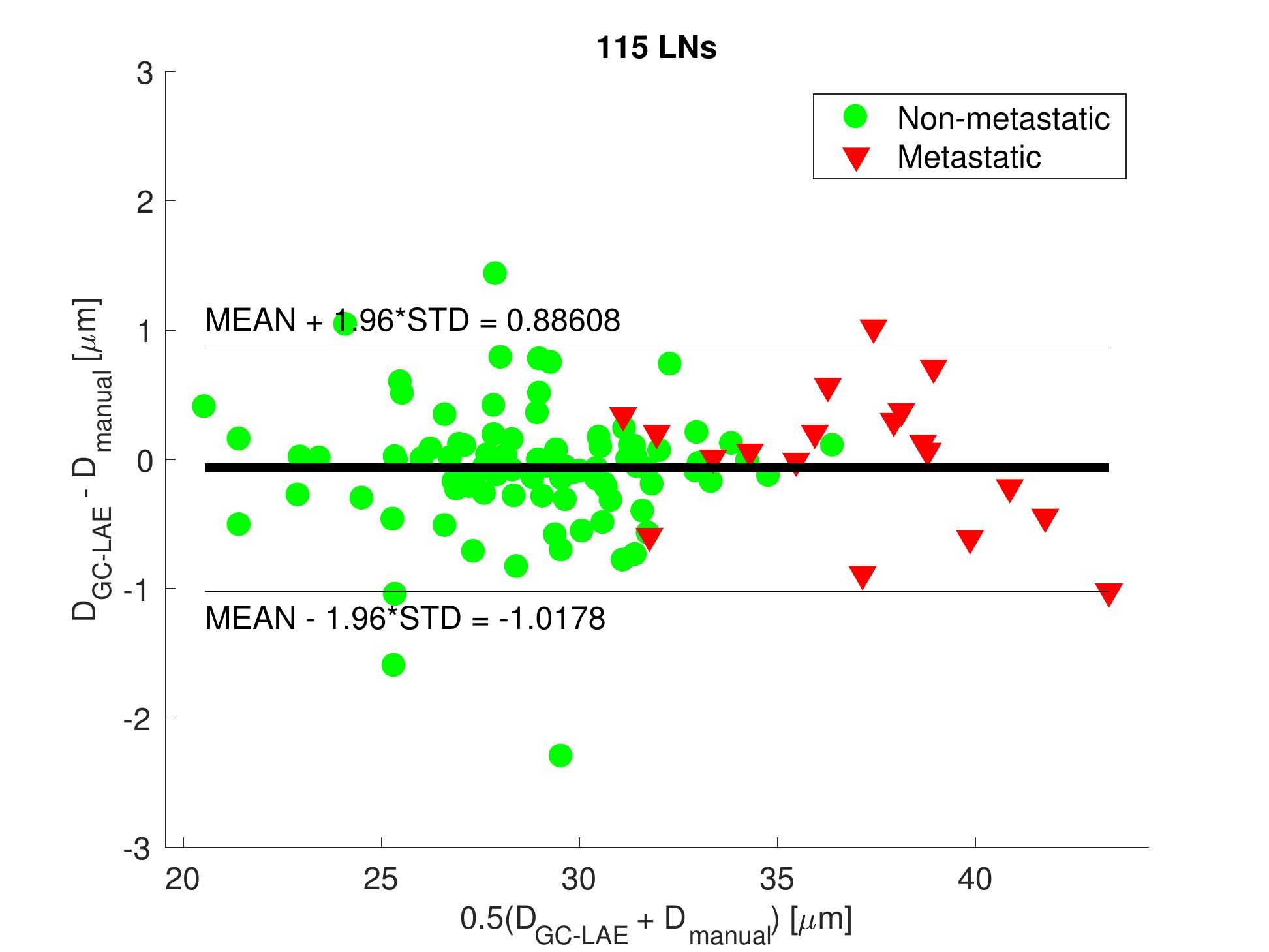}
                 \centerline{(a)}
	\end{minipage} 
	\caption{Difference vs. the mean of the average effective scatterer size estimated by the result of manual segmentation and proposed method on 115 LNs. }
\label{fig:chart}
\end{figure}

\subsection{Clinical Significance}

The goal of our QUS studies is to enable the pathologist to target suspicious regions within the LN for histologic evaluation. This new segmentation algorithm makes the complete QUS analysis process automatic from 3D HFU data acquisition to using QUS to identify suspicious regions. Section \ref{sec:result in our database} clearly establishes that the new segmentation method performs successfully, but it does not provide information about how QUS estimates may be affected by superior segmentation. As shown in Figure \ref{fig:chart}, the average effective scatterer size (D) estimates obtained using manual segmentation and the proposed automatic segmentation agreed well. The mean difference was -0.066 $\mu$m, and the 95$\%$ limits of agreement (i.e., mean $\pm$ 1.96 standard deviations) were -1.02 $\mu$m and +0.89 $\mu$m. As illustrated in our previous studies \cite{mamou2011three}\cite{mamou2010three} and also visible in Figure \ref{fig:chart}, D is a very important QUS parameter for detection of metastatic LNs. Therefore, our segmentation method does not degrade metastases detection using QUS, while providing a completely automatic framework. 

\subsection{Future Work}

\subsubsection{Inter- and Intra-observer variabilities}

The gold standard used for evaluating our segmentation algorithm came from a single expert and because there exists no other means to carefully segment the 3D ultrasound data, no absolute means exist to evaluate the quality of the expert segmentation Therefore, it is reasonable to expect that different experts would give slightly different results. Future studies will be performed to quantify inter- and intra-observer variabilities. Nevertheless, when blindly presented with automatic segmentation and manual segmentation images, our expert was unable to differentiate them.

\subsubsection{Breast LNs}

Our database also includes more than 100 axillary LNs acquired from breast-cancer patients; axillary LNs have a different topology than the LNs acquired from colorectal cancers. In particular, they often contain fatty deposits inside the LNP creating new challenging aspects for automatic segmentation. In future studies, we will meet this challenge by modifying the GC-LAE framework to segment these internal fatty deposits. Now, a 2-layer NGC is used to separate LNP, fat, and PBS. Because fatty deposits are inside the LNP, they also satisfy a nested relationship, and we can extend the 2-layer NGC to 3-layer NGC to segment these fatty deposits as well.

\subsubsection{Other HFU segmentation problems}

The new segmentation method presented here was successfully applied to HFU data acquired from freshly excised human LNs, but the ability to segment objects with spatially varying intensity distributions is essential and applicable to a wide range of imaging applications where HFU is used. HFU images usually suffer from inhomogeneous acoustic attenuation, and GC-LAE could help segment objects in such images. 

\subsection{Concluding Remarks}

This paper describes a novel, fully automatic framework to segment LNP, fat, and PBS simultaneously in LN images. The algorithm combines CG-LAE and NGC. NGC generates a LN mask, which contains only LNP and fat, while GC-LAE further separates LNP and fat. GC-LAE is insensitive to the direction of attenuation and performs well in the presence of inhomogeneous acoustic attenuation in HFU images. Compared to our previous work using NGC with depth-dependent profiles in a EM-based framework, the new framework is superior in handling inhomogeneous intensity distributions and can find the LNP-fat boundary even when the intensity contrast between the two are spatially varying greatly.

The proposed framework further refines the results of GC-LAE by NGC, which can exploit the nested relationship between PBS, fat, and LNP to obtain anatomically correct segmentation. With NGC, the missing boundary between LNP and PBS is well defined by the convex hull of fat. NGC always achieves the global minimum of the defined energy functional. Compared to level-set-based approaches for LN segmentation, the described framework is less vulnerable to converging to a false local minimum during deformation and therefore is not dependent on initialization values.

\ifCLASSOPTIONcaptionsoff
  \newpage
\fi



\bibliographystyle{IEEEtran}
\bibliography{IEEEabrv,refs}

\begin{thebibliography}{10}
\providecommand{\url}[1]{#1}
\csname url@samestyle\endcsname
\providecommand{\newblock}{\relax}
\providecommand{\bibinfo}[2]{#2}
\providecommand{\BIBentrySTDinterwordspacing}{\spaceskip=0pt\relax}
\providecommand{\BIBentryALTinterwordstretchfactor}{4}
\providecommand{\BIBentryALTinterwordspacing}{\spaceskip=\fontdimen2\font plus
\BIBentryALTinterwordstretchfactor\fontdimen3\font minus
  \fontdimen4\font\relax}
\providecommand{\BIBforeignlanguage}[2]{{%
\expandafter\ifx\csname l@#1\endcsname\relax
\typeout{** WARNING: IEEEtran.bst: No hyphenation pattern has been}%
\typeout{** loaded for the language `#1'. Using the pattern for}%
\typeout{** the default language instead.}%
\else
\language=\csname l@#1\endcsname
\fi
#2}}
\providecommand{\BIBdecl}{\relax}
\BIBdecl

\bibitem{saegusa2013three}
E.~Saegusa-Beecroft, J.~Machi, J.~Mamou, M.~Hata, A.~Coron, E.~T. Yanagihara,
  T.~Yamaguchi, M.~L. Oelze, P.~Laugier, and E.~J. Feleppa, ``Three-dimensional
  quantitative ultrasound for detecting lymph node metastases,'' \emph{Journal
  of surgical research}, vol. 183, no.~1, pp. 258--269, 2013.

\bibitem{mamou2011three}
J.~Mamou, A.~Coron, M.~L. Oelze, E.~Saegusa-Beecroft, M.~Hata, P.~Lee,
  J.~Machi, E.~Yanagihara, P.~Laugier, and E.~J. Feleppa, ``Three-dimensional
  high-frequency backscatter and envelope quantification of cancerous human
  lymph nodes,'' \emph{Ultrasound in medicine \& biology}, vol.~37, no.~3, pp.
  345--357, 2011.

\bibitem{mamou2010three}
J.~Mamou, A.~Coron, M.~Hata, J.~Machi, E.~Yanagihara, P.~Laugier, and E.~J.
  Feleppa, ``Three-dimensional high-frequency characterization of cancerous
  lymph nodes,'' \emph{Ultrasound in medicine \& biology}, vol.~36, no.~3, pp.
  361--375, 2010.

\bibitem{coron2012quantitative}
A.~Coron, J.~Mamou, E.~Saegusa-Beecroft, M.~L. Oelze, T.~Yamaguchi, M.~Hata,
  J.~Machi, E.~Yanagihara, P.~Laugier, and E.~J. Feleppa, ``A quantitative
  ultrasound-based method and device for reliably guiding pathologists to
  metastatic regions of dissected lymph nodes,'' in \emph{Biomedical Imaging
  (ISBI), 2012 9th IEEE International Symposium on}.\hskip 1em plus 0.5em minus
  0.4em\relax IEEE, 2012, pp. 1064--1067.

\bibitem{coron2008three}
A.~Coron, J.~Mamou, M.~Hata, J.~Machi, E.~Yanagihara, P.~Laugier, and E.~J.
  Feleppa, ``Three-dimensional segmentation of high-frequency ultrasound echo
  signals from dissected lymph nodes,'' in \emph{Ultrasonics Symposium, 2008.
  IUS 2008. IEEE}.\hskip 1em plus 0.5em minus 0.4em\relax IEEE, 2008, pp.
  1370--1373.

\bibitem{resch2013lymph}
A.~Resch and C.~Langner, ``Lymph node staging in colorectal cancer: old
  controversies and recent advances,'' \emph{World J Gastroenterol}, vol.~19,
  no.~46, pp. 8515--8526, 2013.

\bibitem{Gonzalez:2006:DIP:1076432}
R.~C. Gonzalez and R.~E. Woods, \emph{Digital Image Processing (3rd
  Edition)}.\hskip 1em plus 0.5em minus 0.4em\relax Upper Saddle River, NJ,
  USA: Prentice-Hall, Inc., 2006.

\bibitem{ishikawa2003exact}
H.~Ishikawa, ``Exact optimization for markov random fields with convex
  priors,'' \emph{Pattern Analysis and Machine Intelligence, IEEE Transactions
  on}, vol.~25, no.~10, pp. 1333--1336, 2003.

\bibitem{delong2009globally}
A.~Delong and Y.~Boykov, ``Globally optimal segmentation of multi-region
  objects,'' in \emph{Computer Vision, 2009 IEEE 12th International Conference
  on}.\hskip 1em plus 0.5em minus 0.4em\relax IEEE, 2009, pp. 285--292.

\bibitem{kuo2016nested}
J.-w. Kuo, J.~Mamou, O.~Aristiz{\'a}bal, X.~Zhao, J.~A. Ketterling, and
  Y.~Wang, ``Nested graph cut for automatic segmentation of high-frequency
  ultrasound images of the mouse embryo,'' \emph{IEEE transactions on medical
  imaging}, vol.~35, no.~2, pp. 427--441, 2016.

\bibitem{kuo2015novel}
J.-w. Kuo, J.~Mamou, Y.~Wang, E.~Saegusa-Beecroft, J.~Machi, and E.~J. Feleppa,
  ``A novel nested graph cuts method for segmenting human lymph nodes in 3d
  high frequency ultrasound images,'' in \emph{Biomedical Imaging (ISBI), 2015
  IEEE 12th International Symposium on}.\hskip 1em plus 0.5em minus 0.4em\relax
  IEEE, 2015, pp. 372--375.

\bibitem{6814274}
M.~Nosrati and G.~Hamarneh, ``Local optimization based segmentation of
  spatially-recurring, multi-region objects with part configuration
  constraints,'' \emph{Medical Imaging, IEEE Transactions on}, vol.~33, no.~9,
  pp. 1845--1859, Sept 2014.

\bibitem{bui2015level}
T.~M. Bui, A.~Coron, J.~Mamou, E.~Saegusa-Beecroft, J.~Machi, A.~Dizeux, S.~L.
  Bridal, and E.~J. Feleppa, ``Level-set segmentation of 2d and 3d ultrasound
  data using local gamma distribution fitting energy,'' in \emph{Biomedical
  Imaging (ISBI), 2015 IEEE 12th International Symposium on}.\hskip 1em plus
  0.5em minus 0.4em\relax IEEE, 2015, pp. 1110--1113.

\bibitem{bui2015random}
T.~M. Bui, A.~Coron, L.~Bridal, J.~Mamou, E.~J. Feleppa, E.~Saegusa-Beecroft,
  and J.~Machi, ``Random forest classification and local region-based,
  level-set segmentation for quantitative ultrasound of human lymph nodes,'' in
  \emph{Ultrasonics Symposium (IUS), 2015 IEEE International}.\hskip 1em plus
  0.5em minus 0.4em\relax IEEE, 2015, pp. 1--4.

\bibitem{bui2016local}
T.~M. Bui, A.~Coron, J.~Mamou, E.~Saegusa-Beecroft, T.~Yamaguchi,
  E.~Yanagihara, J.~Machi, S.~L. Bridal, and E.~J. Feleppa, ``Local
  transverse-slice-based level-set method for segmentation of 3d,
  high-frequency ultrasonic backscatter from dissected human lymph nodes,''
  \emph{IEEE Transactions on Biomedical Engineering}, 2016.

\bibitem{felzenszwalb2004efficient}
P.~F. Felzenszwalb and D.~P. Huttenlocher, ``Efficient graph-based image
  segmentation,'' \emph{International journal of computer vision}, vol.~59,
  no.~2, pp. 167--181, 2004.

\bibitem{huang2012robust}
Q.-H. Huang, S.-Y. Lee, L.-Z. Liu, M.-H. Lu, L.-W. Jin, and A.-H. Li, ``A
  robust graph-based segmentation method for breast tumors in ultrasound
  images,'' \emph{Ultrasonics}, vol.~52, no.~2, pp. 266--275, 2012.

\bibitem{zheng2012graph}
L.~Zheng and Q.~Huang, ``A graph-based segmentation method for 3d ultrasound
  images,'' in \emph{Control Conference (CCC), 2012 31st Chinese}.\hskip 1em
  plus 0.5em minus 0.4em\relax IEEE, 2012, pp. 4001--4005.

\bibitem{chang2015graph}
H.~Chang, Z.~Chen, Q.~Huang, J.~Shi, and X.~Li, ``Graph-based learning for
  segmentation of 3d ultrasound images,'' \emph{Neurocomputing}, vol. 151, pp.
  632--644, 2015.

\bibitem{boykov2001fast}
Y.~Boykov, O.~Veksler, and R.~Zabih, ``Fast approximate energy minimization via
  graph cuts,'' \emph{IEEE Transactions on pattern analysis and machine
  intelligence}, vol.~23, no.~11, pp. 1222--1239, 2001.

\bibitem{kolmogorov2004energy}
V.~Kolmogorov and R.~Zabin, ``What energy functions can be minimized via graph
  cuts?'' \emph{Pattern Analysis and Machine Intelligence, IEEE Transactions
  on}, vol.~26, no.~2, pp. 147--159, 2004.

\bibitem{fischler1981random}
M.~A. Fischler and R.~C. Bolles, ``Random sample consensus: a paradigm for
  model fitting with applications to image analysis and automated
  cartography,'' \emph{Communications of the ACM}, vol.~24, no.~6, pp.
  381--395, 1981.

\end{thebibliography}
\end{document}